\documentclass[review,1p,times]{amsart}

\def\Onera{ONERA}

\usepackage{amsmath,amsfonts,amssymb,mathtools, amsthm, dsfont}
\usepackage{stmaryrd}
\usepackage{gensymb}
\usepackage{bm, bbm}

\usepackage{subfigure}

\usepackage{indentfirst}

\usepackage{float}
\usepackage[table]{xcolor}
\usepackage{xcolor}
\usepackage{lscape}

\usepackage{relsize}
\usepackage{enumitem}
\usepackage{hyphenat}
\usepackage[final]{microtype}
\usepackage[lined, ruled, linesnumbered]{algorithm2e}
\usepackage[T1]{fontenc}
\usepackage[utf8]{inputenc}
\usepackage[english]{babel}
\usepackage[left=1.5in,right=1.3in,top=1.1in,bottom=1.1in,includefoot,includehead,headheight=13.6pt]{geometry}

\usepackage{silence}

\usepackage{aecompl}
\usepackage{url}

\usepackage[printonlyused,withpage]{acronym}

\usepackage{ifpdf}

\ifpdf
  \usepackage[pdftex]{graphicx}
\else
  \usepackage{graphicx}
\fi

\graphicspath{{.}{images/}}

\usepackage{color}
\definecolor{linkcol}{rgb}{0,0,0.4}
\definecolor{citecol}{rgb}{0.5,0,0}

\def\argmin{\operatornamewithlimits{arg\,min}}

\renewcommand{\epsilon}{\varepsilon}

\ifpdf
  \usepackage[pagebackref,hyperindex=true]{hyperref}
\else
  \usepackage[dvipdfm,pagebackref,hyperindex=true]{hyperref}
\fi

\usepackage{cleveref}
\usepackage{crossreftools}

\renewcommand*{\backref}[1]{}
\renewcommand*{\backrefalt}[4]{%
\ifcase #1 %
(Not cited.)%
\or
(Cited on page~#2.)%
\else
(Cited on pages~#2.)%
\fi}

\usepackage{pdfpages}

\newcommand{\itr}{\textsf{T}}

\newcommand{\Nset}{\mathbb{N}}
\newcommand{\Rset}{\mathbb{R}}
\newcommand{\esp}{\mathbb{E}}

\newcommand{\pref}[1]{(\ref{#1})}

\definecolor{orange}{rgb}{0.99,0.69,0.07}
\definecolor{pythonblue}{RGB}{0,51,153}
\definecolor{pythongreen}{RGB}{51,153,51}
\definecolor{gris}{gray}{0.80}

\newcommand{\Pytorch}{\texttt{PyTorch}}
\newcommand{\GPytorch}{\texttt{GPyTorch}}
\newcommand{\smt}{\texttt{smt}}
\newcommand{\python}{\texttt{python}}
\newcommand{\x}{\mathbf{x}}
\newcommand{\y}{\mathbf{y}}

\newcommand{\F}{F}

\newcommand{\Nobs}{I}
\newcommand{\Nstep}{N}

\newcommand{\indic}{\mathbbm{1}}

\newcommand{\Xobs}{X}
\newcommand{\Yobs}{Y}
\newcommand{\Xobsb}{\mathbf{\Xobs}}
\newcommand{\Yobsb}{\mathbf{\Yobs}}
\newcommand{\Xobsbi}{\mathbf{\Xobs}_i}
\newcommand{\Xobsbj}{\mathbf{\Xobs}_j}
\newcommand{\Id}{\mathbf{I}}
\newcommand{\InputSpace}{\mathcal{X}}
\newcommand{\OutputSpace}{\mathcal{Y}}

\newcommand{\Dim}{d}

\newcommand{\NT}{\Nobs_{\mathrm{T}}}
\newcommand{\NS}{\Nobs_{\mathrm{s}}}
\newcommand{\NV}{\Nobs_{\mathrm{V}}}
\newcommand{\NTOT}{\Nobs_{\mathrm{Tot}}}

\newcommand{\weight}{\omega}

\newcommand{\Ker}{K}
\newcommand{\Kerb}{\mathbf{K}}
\newcommand{\Kerl}{K+\lambda}
\newcommand{\Kers}{\Ker^{\star}}
\newcommand{\Kerbs}{\Kerb^{\star}}

\newcommand{\norm}[2]{\left\| #1 \right\|_{#2}}
\newcommand{\absolute}[1]{\left| #1 \right|}
\newcommand{\lrangle}[1]{\left\langle #1 \right\rangle}
\newcommand{\Ltwo}{L^2}
\newcommand{\eigenV}{e}
\newcommand{\eigenv}{\sigma}

\newcommand{\HS}{\mathcal{H}}
\newcommand{\FS}{\mathfrak{F}}
\newcommand{\RKHS}{\HS_{\Ker}}

\newcommand{\RKHSl}{\HS_{\Kerl}}
\newcommand{\RKHSn}{\HS_{n}}

\newcommand{\TK}{T_{\Ker}}

\newcommand{\G}{G}
\newcommand{\Gapprox}{\G}

\newcommand{\GapproxNLamb}[1]{\Gapprox_{{#1}}}

\newcommand{\FapproxN}{\Gapprox}
\newcommand{\FapproxNLamb}[1]{\Gapprox_{{#1}}}
\newcommand{\FapproxNs}[1]{\Gapprox^{\star}_{{#1}}}

\newcommand{\fapprox}{g}

\newcommand{\FapproxPCE}{\Gapprox_\basisnumb}

\newcommand{\epsimax}{\eta}
\newcommand{\floor}[1]{\left\lfloor\smash{#1}\right\rfloor}

\newcommand{\DKL}{D_{\mathrm{KL}}}

\newcommand{\var}{\mathrm{Var}}

\newcommand{\lambdamin}{\lambda_{\mathrm{min}}}
\newcommand{\cstarb}{\mathbf{c}^{\star}}
\newcommand{\cstar}{c^{\star}}
\newcommand{\Basis}{\mathcal{B}}
\newcommand{\sparsity}{S}

\newcommand{\basisnumb}{R}
\newcommand{\basisset}{\mathcal{K}}
\newcommand{\node}{\boldsymbol{\vartheta}}

\newcommand{\NF}{\Nobs_{\mathrm{f}}}
\newcommand{\Pif}{\pi_{\mathrm{f}}}
\newcommand{\NC}{\Nobs_{\mathrm{c}}}
\newcommand{\Pic}{\pi_{\mathrm{c}}}

\newcommand{\Fn}{\Gapprox_{\mathrm{f}}}
\newcommand{\Fnd}{\Gapprox_{\mathrm{c}}}

\newcommand{\NRMSE}{e_{\mathrm{NRMSE}}}
\newcommand{\RMSE}{e_{\mathrm{RMSE}}}

\newcommand{\ERRrel}{e_{\mathrm{MRE}}}
\newcommand{\Qtwo}{\mathrm{Q}^{2}}

\newcommand{\Mach}{M}
\newcommand{\Aoa}{\alpha}

\newtheorem{definition}{Definition}
\newtheorem{proposition}{Proposition}
\newtheorem{problem}{Problem}
\newtheorem{theorem}{Theorem}
\newtheorem{remark}{Remark}
\newtheorem{assumption}{Assumption}

\newcounter{example}[section]

\newcommand{\amend}[1]{\textcolor{black}{#1}}

\begin{document}

\title[Learning kernels from data]{Learning "best" kernels from data in Gaussian process regression. With application to aerodynamics}

\author[J.-L. Akian]{Jean-Luc Akian}
\address[J.-L. Akian]{Materials and Structures, \Onera--The French Aerospace Lab, France}
\email{jean-luc.akian@onera.fr}

\author[L. Bonnet]{Luc Bonnet}
\address[L. Bonnet]{Computational Fluid Mechanics, \Onera--The French Aerospace Lab, France}
\email{luc.bonnet@ens-paris-saclay.fr}

\author[H. Owhadi]{Houman Owhadi}
\address[H. Owhadi]{Applied and Computational Mathematics, California Institute of Technology, USA}
\email{owhadi@caltech.edu}

\author[\'E. Savin]{\'Eric Savin}
\address[\'E. Savin]{Information Processing and Systems, \Onera--The French Aerospace Lab, France}
\thanks{Corresponding author: \'E. Savin, ONERA--The French Aerospace Lab, 6 chemin de la Vauve aux Granges, FR-91123 Palaiseau cedex, France (Eric.Savin@onera.fr).}
\email{eric.savin@onera.fr}

\begin{abstract}
This paper introduces algorithms to select/design kernels in Gaussian process regression/kriging surrogate modeling techniques. We adopt the setting of kernel method solutions in \emph{ad hoc} functional spaces, namely Reproducing Kernel Hilbert Spaces (RKHS), to solve the problem of approximating a regular target function given observations of it, \emph{i.e.} supervised learning. A first class of algorithms is kernel flow, which was introduced in the context of classification in machine learning. It can be seen as a cross-validation procedure whereby a "best" kernel is selected such that the loss of accuracy incurred by removing some part of the dataset (typically half of it) is minimized. A second class of algorithms is called spectral kernel ridge regression, and aims at selecting a "best" kernel such that the norm of the function to be approximated is minimal in the associated RKHS. Within Mercer's theorem framework, we obtain an explicit construction of that "best" kernel in terms of the main features of the target function. Both approaches of learning kernels from data are illustrated by numerical examples on synthetic test functions, and on a classical test case in turbulence modeling validation for transonic flows about a two-dimensional airfoil.
\end{abstract}


\date{\today}

\maketitle

%-----------% Introduction %-----------%
\section{Introduction}\label{sec:chap3_introduction}

Cruise flight conditions of commercial aircraft are mostly transonic, such that the flow is locally supersonic due to the geometry of actual wing profiles. Acceleration of the flow on the profile upper surface induces a depression yielding the lift force. A discontinuity, or shock wave, arises if this depression is too sharp in order to balance the pressure gradient at the trailing edge between the upper and lower surfaces. Both the location and strength of the shock wave are responsible for a significant part of the drag force. These features are thus critical in view of optimizing a wing profile, for instance minimizing the drag force considering constant lift force. Mild alterations of profiles can alleviate these issues by smoothing out the discontinuity, hence increasing the lift force while decreasing the drag force.

Such complex aerodynamic design and analysis typically use high-fidelity computational fluid dynamics (CFD) tools for optimization or uncertainty quantification, considering in addition some uncertain operational, environmental, or manufacturing parameters. High-fidelity simulations are needed to detail the flow structures, while non-intrusive methods are further required when variable parameters have to be taken into account for sensitivity and robustness analyses. Due to their complexity, flow solvers are indeed preferably treated as black boxes computing output quantities of interest as functions of input parameters. In aerodynamic applications with CFD software, one single function evaluation can take up to several hours. It is then not conceivable to use this type of complex models for all function evaluations that may be needed to estimate an optimum or an average output, say. A middle ground has thus to be found. One way to solve this issue is to accept to evaluate the complex model at some sample points while using an approximation at the remaining points to mimic the behavior of that complex model. This approximation is called a metamodel, or a surrogate model. Obviously, its quality will strongly influence the trustworthiness of the obtained optimum or average. In that respect, one may wish the surrogate model to fulfill the following two properties\string:
\begin{itemize}
	\item It has to be cheap to evaluate in order to be able to possibly obtain tens of thousands of function evaluations in a reasonable time; typically, an evaluation should be less than 1 second;
	\item It has to be as accurate as possible (given some metric) in order to be confident in these multiple evaluations.
\end{itemize}
This problem can be summed up as follows, in the setting of supervised learning:
\begin{problem}\label{pb:01}
Let $\F : \InputSpace\to \OutputSpace$ be a smooth function mapping an input set $\InputSpace\subset\Rset^{\Dim}$ to an output set $\OutputSpace\subseteq\Rset$, where $\Dim\in\Nset^*$ is the dimension of the input set. Given $\Nobs$ observations of that function denoted by $(\Xobsb, \Yobsb) = \smash{(\Xobsbi, \Yobs_i=\F(\Xobsbi))_{i=1}^\Nobs}$, approximate $\F$.
\end{problem}
\noindent Here $\F$ can be for instance a very complex CFD  computation requiring several hours to obtain one observation $(\Xobsbi, \Yobs_i)$. There exist many different methods to construct a surrogate model. These methods depend on the available information and to cite a few among others\string: Polynomial Chaos Expansion (PCE) and polynomial regression \cite{Ernst12, Ghanem91, LeMaitre10, Nouy10, Queipo05, Soize04, Xiu02}, Proper Orthogonal Decomposition (POD) \cite{Berkooz93, Bui04, Chatterjee2000, Kosambi43, Mathelin12b}, Kriging \cite{Forrester09, Kleijnen09, Sacks89, Rasmussen06, Santner03}, Artificial Neural Networks (ANN) \cite{Berke93, Goodfellow16, Nguyen99, Simpson01, Sun19, Thirumalainambi03, Wallach06, Zhang21}, more recently Physics Informed Neural Networks (PINN) for PDEs \cite{Karniadakis21, Sun20}, \emph{etc}. In this paper, we will focus on so-called kernel methods \cite{Rasmussen06, Scholkopf01} within the framework of Reproducing Kernel Hilbert Spaces (RKHS) \cite{Paulsen2016} to construct a surrogate model, or metamodel, or approximation of $\F$ solving \Cref{pb:01}. This setting is chosen because it has a solid theoretical background and it is quite flexible, in terms of numerical implementation, regularization, convergence, automatization, and interpretability. Also its connections with Kriging and Gaussian process regression \cite{Rasmussen06, Scholkopf01} are well established. These methods have been applied in machine learning, pattern recognition, signal analysis, and more recently to approximate numerical solutions of linear and nonlinear partial differential equations \cite{Chen21b,Raissi18}. Besides, the link between ANN and kernel methods is outlined in \emph{e.g.} \cite{Owhadi20}.

A decisive aspect of these methods, though, is the choice of hyperparameters in hierarchical models of kernel functions and, possibly more importantly, the choice of a relevant kernel base function in this hierarchical modeling. We basically address this issue in the present work, and the remainder of this paper is structured as follows. \Cref{sec:chap3_reg} presents classical kernel regression methods together with their connections. These methods raise the issue of finding a ``best'' kernel function, which is addressed here in two ways. First, the Kernel Flow algorithm initially implemented in a classification context in \cite{Chen21, Owhadi19} is applied to the regression context in \Cref{sec:chap3_KF}. Second, a spectral version of the classical Kernel Ridge Regression algorithm within the framework of Mercer's theorem is developed in \Cref{sec:SKKR_algo}. Mercer's framework is reminded in the appendix \ref{subsec:mercer}, and subsequently used in \Cref{sec:SSKRR} and \Cref{sec:NSKRR} to introduce two algorithms based on regression and projection approaches. The Polynomial Chaos Expansion (PCE) method is briefly reviewed in \Cref{subsec:PCE} for completeness. The proposed algorithms are finally tested on synthetic examples in \Cref{sec:numerical_ex} and on a more complex aerodynamic example in \Cref{subsec:example_UMRIDA_3D}. Also a summary of the theory of RKHS is provided in the appendix \ref{subsec:chap3_RKHS}.

%-----------% Section Regression %-----------%
\section{Regression setting}\label{sec:chap3_reg}

We adopt the setting of kernel method solutions to construct an approximation to \Cref{pb:01} in a functional Hilbert space; see \emph{e.g.} \cite{Kadri16,Scovel19}.

\subsection{Optimal recovery solution}\label{subsec:chap3_ors}

Let $\Ker: \InputSpace\times\InputSpace \to \mathbb{R}$ be a positive definite kernel function and let $\RKHS$ be the Reproducing Kernel Hilbert Space (RKHS) associated with that kernel; see the definitions \ref{def:RKHS} and \ref{def:Kernel} in the appendix \ref{subsec:chap3_RKHS}. Using the norm $\norm{\cdot}{\RKHS}$ in that functional space as the loss, the Optimal Recovery Solution (ORS) of \Cref{pb:01} is the minimizer of (see \cite[Theorems 12.4 and 12.5]{Scovel19}):
\begin{equation}\label{eq:opt_rec_sol}
    \left\{\begin{matrix}
        \min\limits_{\F\in{\RKHS}} \norm{\F}{\RKHS}^{2}\,, \\
        \text{subjected to } \F(\Xobsbi) = \Yobs_i\,, \;i = 1,\dots\Nobs\,.
        \end{matrix}\right.
\end{equation}
This regression provides a minimax optimal approximation of $\F$ in $\RKHS$ \cite{Micchelli77}. From the representer theorem \cite{Micchelli04}, the solution of \Cref{eq:opt_rec_sol} reads:
\begin{equation}
    \FapproxN(\x) = \sum\limits_{i=1}^{\Nobs} \alpha_i \Kerb(\x, \Xobsbi)\,.
\end{equation}
The expansion coefficients $\bm{\alpha} = \{\alpha_i\}_{i=1}^{\Nobs}$ are obtained by solving:
\begin{equation}
    \Kerb(\Xobsb, \Xobsb)\bm{\alpha} = \Yobsb,
\end{equation}
where $\Yobsb = \left(\Yobs_1,\dots\Yobs_{\Nobs}\right)^\itr$ and $\Kerb(\Xobsb, \Xobsb)$ is the Gram matrix defined by:
\begin{equation}\label{eq:kernelb_blockmatrix}
    \Kerb(\Xobsb, \Xobsb) = \begin{bmatrix} \Ker(\Xobsb_1, \Xobsb_1) & \cdots & \Ker(\Xobsb_1, \Xobsb_{\Nobs}) \\ \vdots  & \ddots  & \vdots  \\ \Ker(\Xobsb_{\Nobs}, \Xobsb_1) & \cdots & \Ker(\Xobsb_{\Nobs}, \Xobsb_{\Nobs}) \end{bmatrix}\,.
\end{equation}
Thus one has:
\begin{equation}\label{eq:solution_ORS}
    \F(\x) \simeq\FapproxN(\x) = \Kerb(\x, \Xobsb) \Kerb(\Xobsb, \Xobsb)^{-1} \Yobsb\,,
\end{equation}
where:
\begin{equation}\label{eq:kernelb_linematrix}
    \Kerb(\x, \Xobsb) = \begin{pmatrix}\Ker(\x, \Xobsb_1) & \dots & \Ker(\x, \Xobsb_{\Nobs})\end{pmatrix}.
\end{equation}
\Cref{eq:solution_ORS} implies that the ORS is interpolant, that is, $\FapproxN(\Xobsbi) = \Yobs_i$, $\forall i=1,\dots\Nobs$. In some cases, depending on the position of the data points $\Xobsb$ and their number, the kernel matrix $\Kerb(\Xobsb, \Xobsb)$ might be ill-conditioned and thus numerically non invertible. Therefore, kernel ridge regression is often preferred because it ensures that the kernel matrix is indeed invertible by adding a smoothing term.

\subsection{Kernel ridge regression solution}\label{subsec:chap3_krr}

Let $\lambda > 0 $. The Kernel Ridge Regression (KRR) solution of \Cref{pb:01} is \cite{Owhadi20}:
\begin{equation}\label{eq:rid_reg_sol}
\begin{matrix}
        \min\limits_{\F\in{\RKHS}} \sum\limits_{i=1}^{\Nobs}\left(\Yobs_i - \F(\Xobsbi)\right)^{2} + \lambda\norm{\F}{\RKHS}^{2}\,. \\
 \end{matrix}
\end{equation}
The parameter $\lambda$ adds a penalization term that controls the smoothness of the KRR solution. It is useful to avoid overfitting and is often called nugget. From the representer theorem \cite{Micchelli04}, the solution of \Cref{eq:rid_reg_sol} reads:
\begin{equation}
    \FapproxNLamb{\lambda}(\x) = \sum\limits_{i=1}^{\Nobs} \alpha_i \Kerb(\x, \Xobsbi)\,.
\end{equation}
The expansion oefficients  $\bm{\alpha} = \{\alpha_i\}_{i=1}^{\Nobs}$ are obtained by solving:
\begin{equation}\label{eq:alpha_rrs}
    \left(\Kerb(\Xobsb, \Xobsb) + \lambda\Id_{\Nobs}\right)\bm{\alpha} = \Yobsb\,,
\end{equation}
where $\Kerb(\Xobsb, \Xobsb)$ is the Gram matrix defined by \Cref{eq:kernelb_blockmatrix} and $\Id_{\Nobs}$ is the $\Nobs\times\Nobs$ identity matrix. The matrix $\Kerb(\Xobsb, \Xobsb) + \lambda\Id_\Nobs$ is invertible if $\lambda\geq 0 $. Thus the prediction at an unobserved point $\x$ reads:
\begin{equation}\label{eq:solution_KRR}
    \F(\x) \simeq \FapproxNLamb{\lambda}(\x) = \Kerb(\x, \Xobsb) \left(\Kerb(\Xobsb, \Xobsb) + \lambda\Id_\Nobs\right)^{-1} \Yobsb\,.
\end{equation}
The main difference with \Cref{eq:opt_rec_sol} is that the KRR solution is not interpolant because of the addition of the parameter $\lambda$, that controls its possible overfitting. In practical cases, this parameter is usually chosen as $\lambda \ll 1$. It may also be interpreted as the variance of some measurement noise. For $\lambda = 0$, $\FapproxNLamb{0} \equiv \FapproxN$.

\subsection{Deterministic error estimation of the KRR solution}\label{subsec:chap3_det_error}

Let $\F\in\RKHS$ be the ground truth function and let $\FapproxNLamb{\lambda}$ be its KRR approximation \pref{eq:solution_KRR} with $\lambda \geq 0$. From \cite[Theorem 8.4]{Owhadi20}, one has for any $\x \in \InputSpace$:
\begin{equation}\label{eq:bound_error_RKHS}
   \absolute{\F(\x) - \FapproxNLamb{\lambda}(\x)} \leq \sigma(\x) \norm{\F}{\RKHS}
\end{equation}
with $\norm{\F}{\RKHS}<+\infty$, and:
\begin{equation}\label{eq:bound_error_RKHSl}
    \absolute{\F(\x) - \FapproxNLamb{\lambda}(\x)} \leq \sqrt{\sigma^2(\x)+\lambda} \norm{\F}{\RKHSl}
\end{equation}
with $\norm{\F}{\RKHSl} < +\infty$, $\RKHSl$ being the RKHS associated with the kernel $\Ker + \lambda$, and:
\begin{equation}\label{eq:sigma_GPR}
\sigma^2(\x) = \Ker(\x,\x) - \Kerb(\x, \Xobsb)\left(\Kerb(\Xobsb, \Xobsb) + \lambda\Id_\Nobs\right)^{-1}\Kerb(\Xobsb, \x)
\end{equation}
with $\Kerb(\Xobsb, \x) = \Kerb(\x, \Xobsb)^\itr$. Thus \Cref{eq:bound_error_RKHS} and \Cref{eq:bound_error_RKHSl} provide with bounds on the deterministic error $\absolute{\F(\x) - \FapproxNLamb{\lambda}(\x)}$, which depend on the norms $\norm{\F}{\RKHS}$, $\norm{\F}{\RKHSl}$, and the variance $\sigma^2(\x)$ which is independent of $\F$. Therefore, reducing the discrepancy between the ground truth function $\F$ and its approximation $\FapproxNLamb{\lambda}$ at some point $\x\in\InputSpace$ amounts to lowering $\sigma^2(\x)$ independently of $\F$, and/or finding a ``best'' kernel $\Ker$ in a sense that is elaborated further on in \Cref{sec:chap3_KF} and \Cref{sec:SKKR_algo} below.

%-----------% Section KF %-----------%
%!TEX root = main.tex

\section{Kernel Flow algorithm}\label{sec:chap3_KF}

 A challenging aspect of kernel methods is to determine which kernel $\Ker$ to select in order to address \Cref{pb:01}. A kernel may be either a parametric or a non parametric function. In the former case, a certain number of parameters have to be determined for each type of kernel, for instance the kernels presented in the appendix \ref{subsec:examples_kernel}. They are often called the hyperparameters of the kernel. Multiple methods exist in order to determine them, including maximum likelihood \cite{Williams96} (choosing the parameters which maximize the probability of observing the data), Bayesian inference \cite{Schwai04} (placing a prior on the kernel and conditioning with respect to the data), cross-validation \cite{Chen21} (splitting $(\Xobsb, \Yobsb)$ into training data and validation data in a controlled or uncontrolled way), \emph{etc}. Here we follow another approach to find a ``best'' kernel $\Ker$ in a sense that is clarified below in \Cref{subsec:chap3_best_kernel}, the Kernel Flow (KF) iterative algorithm of \cite{Owhadi19}. It was first used in a machine learning context for classification \cite{Owhadi19,Yoo21} and more recently in geophysical forecasting \cite{Hamzi21_forecast} and with dynamical systems \cite{Darcy21,Hamzi21_dyn}. %KF algorithms have been applied to various problems in image classification, geoscience, or physics in \emph{e.g.} \cite{Darcy21,Hamzi21_dyn,Hamzi21_forecast,Owhadi19,Yoo21}.
 Actually it can be seen as an equivalent of cross-validation in a regression context, performing double, or nested cross-validation \cite{Chen21,Stone74}. \amend{Early attempts to learn kernels from data can be found in geostatistics, for example, where the spatial correlation structures of data are described in terms of so-called variograms introduced by Matheron \cite{Matheron63} and their estimates; see \emph{e.g.} \cite{Cressie1993,Davis2002,Stein99}}.
 
 The main objective of the KF algorithm is to learn kernels $\Ker_n$ of the following form:
\begin{equation}\label{eq:Ker_nonp}
	\Ker_n(\Xobsb_1, \Xobsb_2) = \Ker(f_n(\Xobsb_1), f_n(\Xobsb_2);\bm{\theta})\,,
\end{equation}
where $\Ker$ is a base kernel, for instance the Gaussian kernel defined by \Cref{eq:Gaussian_kernel}, and $f_n: \InputSpace \to \InputSpace $ is called the flow in the input space at the $n$--th iteration step. This can be understood as a non-parametric approach to iteratively find a ``best'' kernel $\Ker_n$, where instead of searching for hyperparameters $\bm{\theta}$, a whole flow function $f_n$ is sought for. The KF algorithm can also be used in a parametric way when one rather seeks to iteratively learn the hyperparameters $\bm{\theta}$ of a base kernel $\Ker$, for example the length scales $\bm{\theta}\equiv(\gamma_i)_{i=1}^\Dim$ of a Gaussian kernel \pref{eq:Gaussian_kernel_ARD}. That is,
\begin{equation}\label{eq:Ker_p}
    \Ker_n(\Xobsb_1, \Xobsb_2) = \Ker(\Xobsb_1, \Xobsb_2;\bm{\theta}_n)\,,
\end{equation}
where $\bm{\theta}_n$ are the hyperparameters at the $n$--th iteration step of the parametric KF algorithm. This approach is the one retained in the remainder of the paper.

\subsection{What is the ``best'' kernel?}\label{subsec:chap3_best_kernel}

We still have to define when a kernel is considered as the ``best'' one. Here, a kernel $\Ker$ is selected as the ``best'' one if the number of regression points can be halved without losing too much accuracy, where the latter is measured with the RKHS norm associated with that kernel \cite{Chen21,Owhadi19}.
\begin{figure}
    \centering
    \subfigure[$\NF$ observations.]{\includegraphics[width=.4\linewidth]{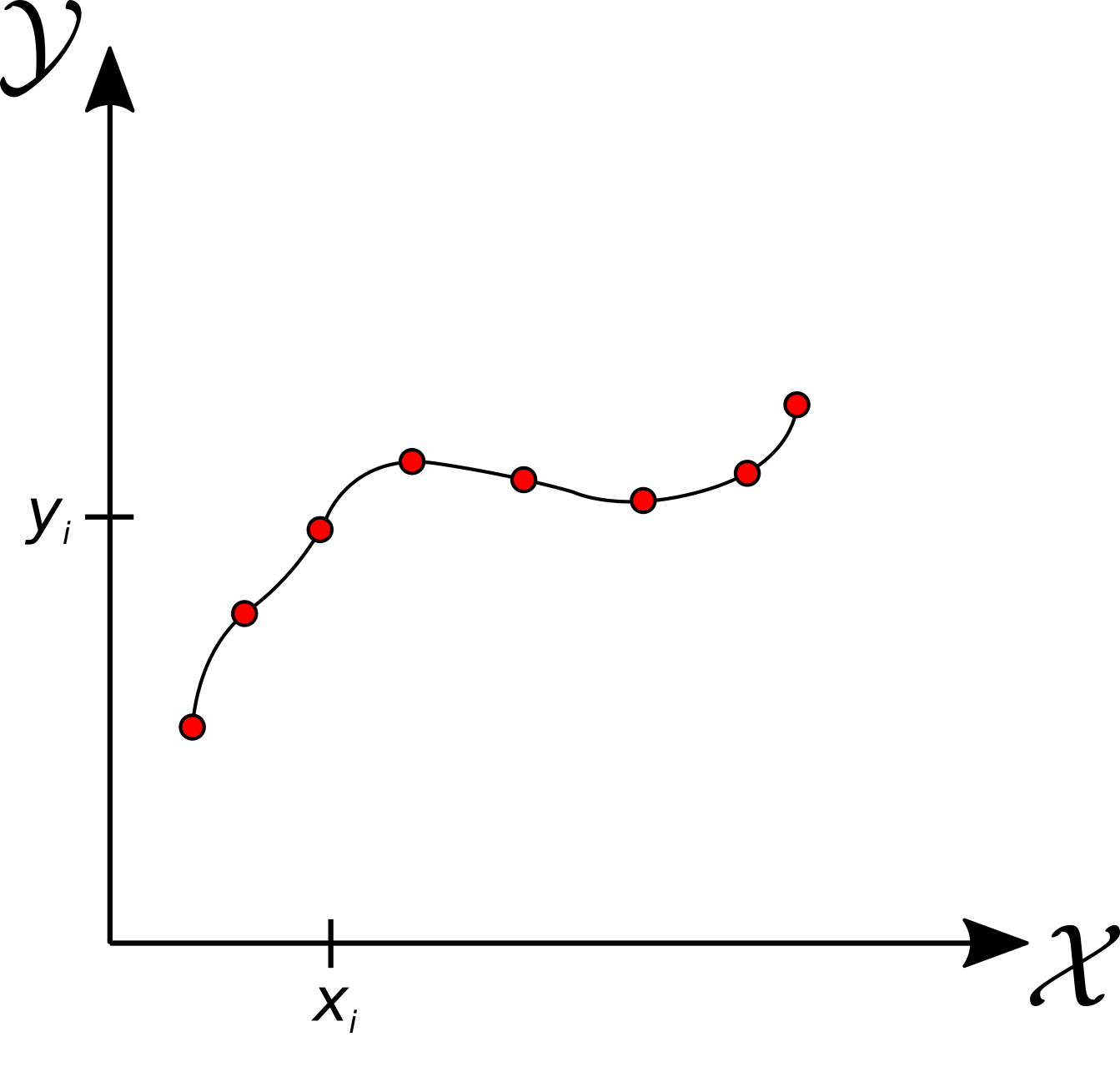}}
    \subfigure[$\NC$ observations.]{\includegraphics[width=.4\linewidth]{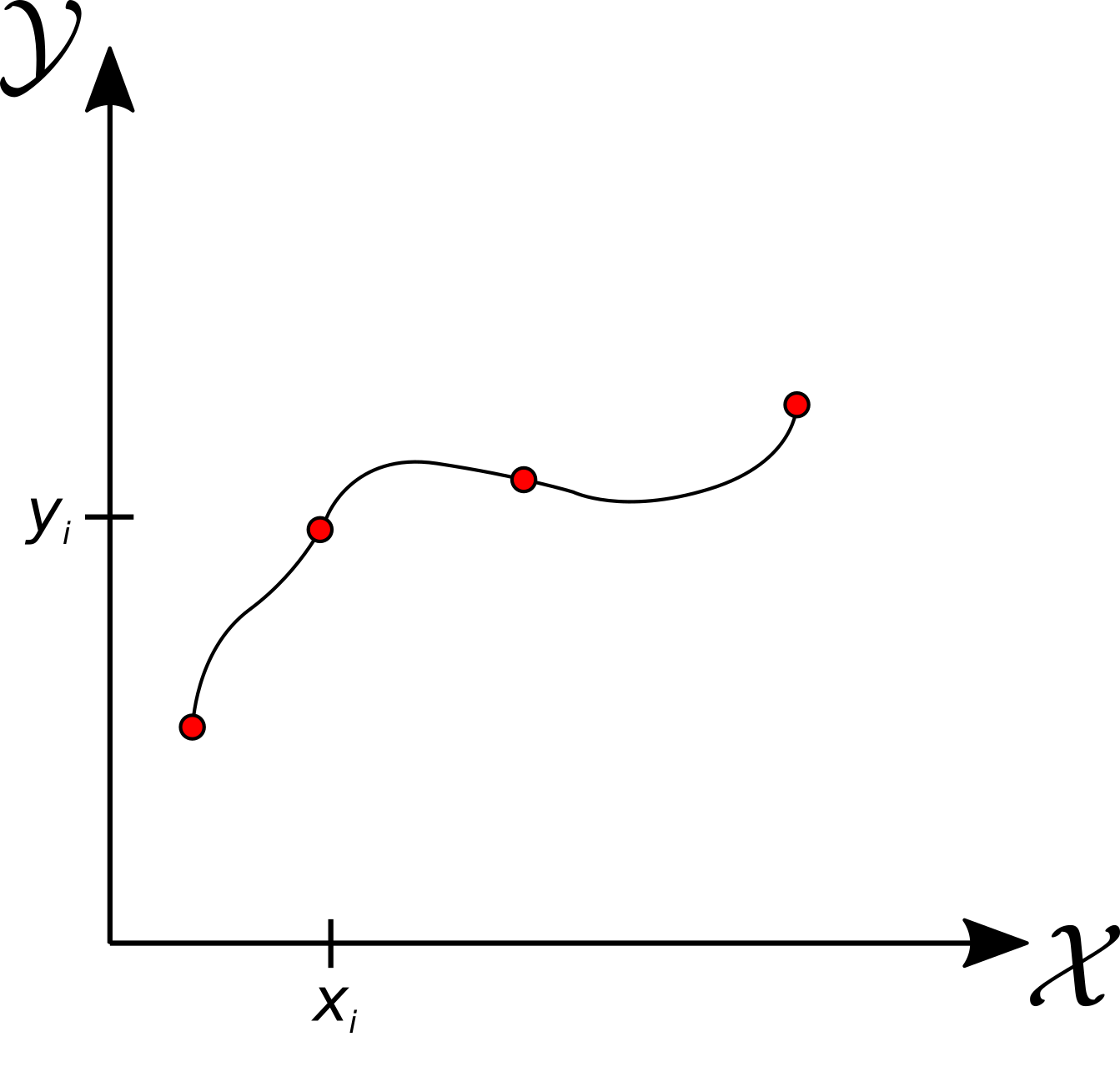}}
    \caption{(a) $\Fn$: Interpolating $\F$ with kernel $\Ker$ and $\NF$ observations. (b) $\Fnd$: Interpolating $\F$ with kernel $\Ker$ and $\NC = \floor{\frac{\NF}{2}}$ observations.}
    \label{fig:KF_halved}
\end{figure}

We start by selecting $\NF \leq \Nobs$ observations of $\F$ with which we construct an approximation $\Fn$ using \Cref{eq:solution_ORS}, or \Cref{eq:solution_KRR} and a yet to be selected nugget $\lambda$, with kernel $\Ker_n$ at the $n$--th iteration of the form \pref{eq:Ker_p}. We also select $\NC = \smash{\floor{\frac{\NF}{2}}}$ observations among these $\NF$ ones and construct an approximation $\Fnd$ of $\F$ with the same kernel $\Ker_n$ using the same foregoing methods; see \Cref{fig:KF_halved} where $\floor{\cdot}$ is the floor function. Then we introduce the following function $\rho_n\in[0,1]$ to quantify the accuracy of the surrogate model at iteration $n$ as:
\begin{equation} \label{eq:rho_n}
	\rho_n(\Xobsb_{\Pif^{n}}, \Xobsb_{\Pic^{n}};\bm{\theta}_n) = \frac{\left\| \Fn - \Fnd \right\|^{2}_{\RKHSn}}{\left\| \Fn \right\|^{2}_{\RKHSn}}\,,
\end{equation}
where the functional space $\RKHSn$ is the unique RKHS associated with the kernel $\Ker_n$; $\Pif^{n}$ and $\Pic^{n}$ are the indices corresponding to the $\NF$ and $\NC$ observations, respectively, at iteration $n$; and $\Xobsb_{\Pif^{n}}$ and $\Xobsb_{\Pic^{n}}$ are the corresponding inputs. In that way, the smaller $\rho_n$ is, the better the kernel $\Ker_n$ is. That is, if $\rho_n$ gets close to zero, the kernel $\Ker_n$ is the ``best'' one according to the definition stated above. It is shown in \cite{Owhadi19} that $\rho_n$ in \Cref{eq:rho_n} also reads: 
\begin{equation}\label{eq:rho_p}
	\rho_n(\Xobsb_{\Pif^{n}}, \Xobsb_{\Pic^{n}};\bm{\theta}_n) = 1 - \frac{\Yobsb_{\Pic^{n}}^\itr \Kerb(\Xobsb_{\Pic^{n}}, \Xobsb_{\Pic^{n}};\bm{\theta}_n)^{-1}\Yobsb_{\Pic^{n}}}{\Yobsb_{\Pif^{n}}^\itr \Kerb(\Xobsb_{\Pif^{n}}, \Xobsb_{\Pif^{n}};\bm{\theta}_n)^{-1}\Yobsb_{\Pif^{n}}}\,,
\end{equation}
where $\Yobsb_{\Pif^{n}}$ and $\Yobsb_{\Pic^{n}}$ are the observed outputs corresponding to the indices $\Pif^{n}$ and $\Pic^{n}$, respectively; %$\smash{\Kerb(f_n(\Xobsb_{\Pif^{n}}),f_n(\Xobsb_{\Pif^{n}}))}$ and $\smash{\Kerb(f_n(\Xobsb_{\Pic^{n}}), f_n(\Xobsb_{\Pic^{n}}))}$ are the matrices \pref{eq:kernelb_blockmatrix} constructed  with the kernel $\smash{\Ker_n}$ and the inputs $\smash{\Xobsb_{\Pif^{n}}}$ and $\smash{\Xobsb_{\Pic^{n}}}$, respectively, for the non-parametric case, 
and $ \Kerb(\Xobsb_{\Pif^{n}}, \Xobsb_{\Pif^{n}}; \bm{\theta}_n)$ and $ \Kerb(\Xobsb_{\Pic^{n}}, \Xobsb_{\Pic^{n}}; \bm{\theta}_n)$ are the matrices \pref{eq:kernelb_blockmatrix} constructed  with the kernel $\smash{\Ker_n}$ of \Cref{eq:Ker_p} and the inputs $\smash{\Xobsb_{\Pif^{n}}}$ and $\smash{\Xobsb_{\Pic^{n}}}$, respectively. The formula \pref{eq:rho_p} enables the numerical computation of the accuracy $\rho_n$, and stems from the identity $\smash{\norm{\FapproxN}{\RKHS}^2}=\smash{\Yobsb^\itr\Kerb(\Xobsb,\Xobsb)^{-1}\Yobsb}$ whenever $\FapproxN(\x)=\smash{\Kerb(\x,\Xobsb)\Kerb(\Xobsb,\Xobsb)^{-1}\Yobsb}$. The parametric KF algorithm is summarized below.

\subsection{Parametric KF algorithm}\label{subsec:chap3_P_KF_algo}

The parametric KF algorithm aims to determine one or more parameters of a chosen base kernel. Considering a family of kernels $\Ker_{\bm{\theta}}(\Xobsb_1, \Xobsb_2) = \Ker(\Xobsb_1, \Xobsb_2;\bm{\theta})$ parameterized by the parameters $\bm{\theta}$, the parametric version of the KF algorithm runs as follows from iteration $n$ to $n+1$:
\begin{enumerate}
	\item Select $\NF \leq \Nobs$ observations at random among the $\Nobs$ observations;
	\item Select $\NC = \floor{\frac{\NF}{2}}$ observations at random among these $\NF$ observations;
	\item Compute the accuracy $\rho_n(\mathbf{X}_{\Pif^{n}}, \mathbf{X}_{\Pic^{n}}; \bm{\theta}_n)$ given by \Cref{eq:rho_n}. The $\NC$ observations are used only in this step; 
	\item Compute the gradient $\bm{\nabla}_{\bm{\theta}}\rho_n$ of $\rho_n$ with respect to the parameters $\bm{\theta}_n$;
	\item Move $\bm{\theta}_n$ in the gradient descent direction $\bm{\nabla}_{\bm{\theta}}\rho_n$ of $\rho_n$; 
	% $\theta \leftarrow \theta - \eta{\nabla}_{\bm{\theta}}\rho_n$, where $\eta$ is the ``learning rate'';
	\item Return to step 1.
\end{enumerate}

This parametric version is applied to synthetic test functions in \Cref{sec:numerical_ex}, and to an aerodynamic test case using synthetic observations from a CFD software in \Cref{subsec:example_UMRIDA_3D}.

%-----------% Section SKRR %-----------%
\section{Spectral Kernel Ridge Regression algorithms}\label{sec:SKKR_algo}

In this section, we adopt a different perspective to determine what is a ``best'' kernel. New algorithms that we call Spectral Kernel Ridge Regression (SKRR) are introduced in this respect. Recalling \Cref{subsec:chap3_det_error} and the deterministic error bounds of \Cref{eq:bound_error_RKHS} and \Cref{eq:bound_error_RKHSl}, we observe that the pointwise error induced by the surrogate model is controlled by the norms $\norm{\F}{\RKHS}$ and $\smash{\norm{\F}{\RKHSl}}$ apart from the local variance $\sigma^2(\x)$ which is, again, independent of the ground truth $\F$. Therefore, we suggest to study the effect of the kernel $\Ker$ on the norm $\norm{\F}{\RKHS}$ and in this way to find a ``best'' kernel by minimizing the latter with respect to $\Ker$. That is, we aim to find the kernel $\Kers$ such that:
\begin{equation}\label{eq:best_min_Kers}
    \Kers = \argmin\limits_{\Ker} \norm{\F}{\RKHS}\,.
\end{equation}
For that purpose, we resort to Mercer's theorem and the spectral expansion of the integral operator associated to a Mercer kernel. These notions are summarized in the appendix \ref{subsec:mercer}. Therefore, we assume from now on that $\InputSpace\subset\Rset^\Dim$ is compact.

% ######################################
\subsection{Minimizing the norm \texorpdfstring{$\norm{\F}{\RKHS}$}{norm Fdag}\string: the SKRR core} \label{sec:min_norm}

We assume that a family $\{\eigenV_i\}_{i\in\Nset}$ of continuous functions on $\InputSpace$ is given (these are the ``features'' in machine learning techniques, for example), which forms an orthonormal basis of $\Ltwo(\InputSpace, \mu)$, the set of square integrable functions with respect to a Borel probability measure $ \mu$ on $\InputSpace$. It is a Hilbert space endowed with the inner product $\lrangle{\cdot,\cdot}_{\Ltwo}$. We aim to construct a Mercer kernel $\Ker = \sum_{i}\eigenv_i \eigenV_i \otimes \eigenV_i$ where the eigenvalues $\{\eigenv_i > 0\}_{i\in\Nset}$ are to be chosen such that, if $\F\in\RKHS$, its norm $\norm{\F}{\RKHS}$ is minimal. By \Cref{eq:norm_RKHS}, the latter reads:
\begin{equation}
    \norm{\F}{\RKHS}^{2} = \sum\limits_{i=0}^{+\infty} \frac{{\absolute{\F_i}{}^{2}}}{\eigenv_i}\,,
\end{equation}
where $\F_i = \lrangle{ \F,\eigenV_i}_{\Ltwo}$, $i\in\Nset$, such that $\sum_i \absolute{\F_i}{}^2 < +\infty$ because $\F\in\Ltwo(\InputSpace, \mu)$. The minimization problem thus reads:
\begin{equation}\label{eq:min_sigma}
        \min\limits_{\{\eigenv_i\}_{i}} \sum\limits_{i}\frac{\absolute{\F_i}{}^2}{\eigenv_i} \; \text{subjected to } \sum\limits_{i}\eigenv_i = \kappa\,,
\end{equation}
where $0 < \kappa < + \infty$; this condition arises from \Cref{eq:TK_trace_class}. The minimization problem \pref{eq:min_sigma} can be solved by the method of Lagrange multipliers. Let $\nu$ be a Lagrange multiplier and $\{\eigenv^{\star}_i\}_{i\in\Nset}$ be the solution of \Cref{eq:min_sigma}, one has:
\begin{equation*}\label{eq:lagrange}
    \begin{split}
        \frac{\partial}{\partial\eigenv_k}\left[\sum\limits_{i}\frac{\absolute{\F_i}{}^2}{\eigenv^{\star}_i} + \nu\left(\sum\limits_{i}\eigenv^{\star}_i  - \kappa\right)\right] &= 0 \quad \text{ for } k = 0,1,2,\dots\,,  \\
        \frac{\partial}{\partial\nu}\left[\sum\limits_{i}\frac{\absolute{\F_i}{}^2}{\eigenv^{\star}_i} + \nu\left(\sum\limits_{i}\eigenv^{\star}_i  - \kappa\right)\right] &= 0\,.
    \end{split}
\end{equation*}
Then one finds that:
\begin{equation}
        \nu = \frac{1}{\kappa^2}\left(\sum\limits_{i}\absolute{\F_i}{}\right)^2,\quad \eigenv^{\star}_i = \frac{|{\F_i}|}{\sqrt{\nu}}, \quad i = 0,1,2,\dots\,,
\end{equation}
or:
\begin{equation}\label{eq:finish_lagrange}
	\eigenv^{\star}_i =  \frac{\kappa|{\F_i}|}{\sum\limits_{j}\absolute{\F_j}{}},\quad i = 0,1,2,\dots
\end{equation}
In this way we built a Mercer kernel $\Kers = \sum_i \eigenv^{\star}_i \eigenV_i \otimes \eigenV_i$ which minimizes the norm  $\norm{\F}{\RKHS}$. Practically, the sum extends up to a finite rank $\basisnumb$. It remains to compute the expansion coefficients $\{\F_i\}_{i\in\Nset}$ (the "feature extraction" step, say): two approaches based on regression are outlined in the next section. \amend{We note at this stage that although the family $\{\eigenV_i\}_{i\in\Nset}$ may depend on the measure $\mu$, the RKHS $\RKHS$ associated to $\Ker$ does not; see for example \cite[Prop. 11.17]{Paulsen2016} and the comment after the proof of this proposition, and \Cref{rk:independent} in the appendix \ref{subsec:mercer}}.

% ######################################
\subsection{Computing the expansion coefficients}\label{sec:proj_coeff}

Given the orthonormal basis $\{\eigenV_i\}_{i\in\Nset}$ of $\Ltwo(\InputSpace, \mu)$ and $\F\in\Ltwo(\InputSpace, \mu)$, one has thus the following expansions:
\begin{equation*}
    \F(\Xobsbj) = \sum\limits_{i=0}^{+\infty}\lrangle{ \F, \eigenV_i}_{\Ltwo} \eigenV_i(\Xobsbj) = \Yobs_j\,,\quad j = 1,\dots\Nobs\,,
\end{equation*}
because the functions  $\{\eigenV_i\}_{i\in\Nset}$ are continuous. Let $\basisset$ be a set of indices such that $\# \basisset = \basisnumb$ and let the corresponding orthonormal family $\{\eigenV_k\}_{k\in\basisset} \subset \{\eigenV_i\}_{i\in\Nset}$ in $\Ltwo(\InputSpace, \mu)$ be the basis $\Basis^{\basisnumb}$. Let $\bm{\Theta}$ be the $\Nobs \times \basisnumb$ measurement matrix and $\mathbf{c}$ be the expansion coefficients vector in $\Rset^\basisnumb$ such that:
\begin{equation}
    \bm{\Theta} =
    \begin{bmatrix}
        \eigenV_{k_1}(\Xobsb_1) & \cdots  & \eigenV_{k_\basisnumb}(\Xobsb_1)\\
        \vdots & \ddots & \vdots\\
        \eigenV_{k_1}(\Xobsb_\Nobs) & \cdots  & \eigenV_{k_\basisnumb}(\Xobsb_\Nobs)
        \end{bmatrix}, \quad
        \mathbf{c} = \begin{pmatrix}
            c_{k_1} \\
            \vdots \\
            c_{k_\basisnumb}
            \end{pmatrix}
        =
        \begin{pmatrix}
            \lrangle{ \F, \eigenV_{k_1}}_{\Ltwo} \\
            \vdots \\
            \lrangle{\F, \eigenV_{k_\basisnumb}}_{\Ltwo}
            \end{pmatrix}\,;
\end{equation}
then one arrives at the following $\Nobs \times \basisnumb$ system:
\begin{equation}\label{eq:illposed}
    \Yobsb = \bm{\Theta} \mathbf{c} + \bm{\epsimax}\,,
\end{equation}
where $\bm{\epsimax} = (\epsimax_1,\ldots \epsimax_{\Nobs})^\itr$ is an error vector with $\norm{\bm{\epsimax}}{2} \leq \epsimax$ accounting for the truncation of the ground truth function $\F$ on the orthonormal set of $\basisnumb$ vectors $\{\eigenV_k\}_{k\in\basisset}$, and possible noise.

\subsubsection{Least-squares regression}

We first assume that $\Nobs \geq \basisnumb$. Finding the expansion coefficients $\mathbf{c}$ can be done through a regression approach formulated as a least-squares minimization problem, that is, solving the following problem:
\begin{equation}
    \cstarb = \argmin_{\mathbf{h}\in\Rset^{\basisnumb}} \left(\bm{\Theta} \mathbf{h} - \Yobsb \right)^\itr\left(\bm{\Theta} \mathbf{h} - \Yobsb\right)\,.
\end{equation}
This approach is detailed in \emph{e.g.} \cite{Hadigol18} and references therein. Here, we are more interested in the case where $\Nobs < \basisnumb$ or even $\Nobs \ll \basisnumb$, namely when the number of possible ``features'' $\basisnumb$ is way more larger than the number of observations $\Nobs$ of $\F$. This is the topic of the following section.

\subsubsection{Sparse regression}\label{sec:chap3_sparse_recons}

We now assume that $\Nobs \ll \basisnumb$, and that the ``features'' are actually chosen such that the ground truth function $\F$ is expected to be sparse or nearly sparse on this basis: many components of the vector $\mathbf{c}$ of its expansion coefficients are negligible. Such expansion is known as compressible in the terminology of compressed sensing, or compressive sampling (CS) \cite{Candes06,Candes08_intro,Donoho06}. Thus one introduces the sparsity $\sparsity$ defined by:
\begin{equation}
	\sparsity = \# \left\{i; \absolute{c_i} > \delta \right\}\,,
\end{equation}
where $\delta > 0$ is some tolerance, and assume that $\sparsity \ll \basisnumb$. In other words, only a small number of vectors within the basis $\Basis^{\basisnumb}$ is relevant to reconstructing the ground truth function $\F$ without much loss, and this number is that sparsity $\sparsity$. Then \Cref{eq:illposed} in this context may be solved by adaptative methods such as least angle regression as in \emph{e.g.} \cite{Blatman2011}, or by non-adaptive methods such as the following convex $\ell_1$-minimization known as Basis Pursuit Denoising (BPDN) \cite{Chen06}:
\begin{equation}\label{eq:BPDN}
    \min\limits_{\mathbf{h}\in\Rset^{\basisnumb}} \norm{\mathbf{h}}{1} \;\text{ subjected to}\; \norm{\bm{\Theta} \mathbf{h} - \Yobsb}{2} \leq \epsimax\,,
\end{equation}
where $ \norm{\mathbf{h}}{p}=(\sum_{j=1}^\basisnumb\absolute{h_j}^p)^\frac{1}{p}$, $p>0$. BPDN is non-adapted because it identifies both the sparsity pattern, that is the order of the negligible components in the sought vector $\mathbf{c}$, and the leading components at the same time. This is clearly a desirable feature for practical industrial applications. Therefore this approach is favored in the subsequent numerical examples. The conditions on the measurement matrix $\bm{\Theta}$ for which \Cref{eq:BPDN} yields a unique solution, and associated recovery bounds are analyzed in \emph{e.g.} \cite{Candes05,Candes08,Candes08_intro,Candes11} for the noiseless ($\epsimax=0$) and noisy ($\epsimax>0$) cases.

It should be noted that in practical applications, the sparsity of $\F$ is typically seen \emph{a posteriori} and not \emph{a priori}. If the orthonormal set of vectors $\left\{ \eigenV_k \right\}_{k\in\basisset}$ and $\epsimax$ are well chosen, \Cref{eq:BPDN} will yield a sparse solution which approximates well the ground truth function $\F$, \emph{i.e.} a solution where only a few terms are non vanishing. In this paper, the selection of the value $\epsimax$ is done arbitrarily but it can actually be chosen through cross-validation \cite{Boufounos07, Doostan11, Ward09}, for example.

\subsection{Sparse SKRR algorithm}\label{sec:SSKRR}

We propose the following algorithm that we coin Sparse Spectral Kernel Ridge Regression (SSKRR) which couples the sparse reconstruction by, say, $\ell_1$-minimization presented in the foregoing section, and the KRR approximation detailed in \Cref{subsec:chap3_krr}. The main idea of this algorithm is to minimize the RKHS norm of the ground truth function $\F$ with respect to the eigenvalues of a Mercer kernel, which are obtained by \Cref{eq:finish_lagrange} where the expansion coefficients $\{\F_i\}_{i\in\basisset}$ in a finite basis $\Basis^{\basisnumb}$ are computed by $\ell_1$-minimization. The procedure is sketched in \Cref{algo:SKRR_sparse}. It is organized as follows. First, starting from an orthonormal set of $\basisnumb$ vectors $\Basis^{\basisnumb} \equiv \{\eigenV_k\}_{k\in\basisset} $ in $\Ltwo(\InputSpace, \mu)$, $\Nobs$ observations of $\F$, and the parameter $\epsimax$, the BPDN minimization of \Cref{eq:BPDN} is solved, yielding a solution $\cstarb$. If the set $\Basis^{\basisnumb}$ of ``features'' is well chosen, only a limited number of terms in the vector $\cstarb$ is not close to zero. Then one builds the KRR approximation \pref{eq:solution_KRR}, which allows us to get a prediction of the ground truth function $\F$ at an unobserved location $\x$ with a nugget $\lambda$ and using the kernel obtained at the previous step. One of the main advantages of the proposed algorithm is that it provides the prediction variance $\sigma^{2}(\x)$ at the unobserved point $\x$, as in the Gaussian Process Regression framework.

Obviously, the algorithm will strongly depend on the performance of $\ell_1$-minimization to approximate the expansion coefficients $\{\F_i\}_{i\in\basisset}$. Also, several remarks can be made about the parameter $\kappa$ in \Cref{eq:finish_lagrange}. The SKRR approximation at an unobserved point $\x$ is:
\begin{equation*}
\FapproxNs{\lambda}(\x) = \Kerbs(\x, \Xobsb) \left(\Kerbs(\Xobsb, \Xobsb) + \lambda\Id_\Nobs\right)^{-1} \Yobsb\,,
\end{equation*}
and using \Cref{eq:finish_lagrange} one arrives at:
    \begin{equation*}
        \begin{aligned}
            \FapproxNs{\lambda}(\x)% &= \sum\limits_{k\in\basisset} \frac{\kappa|{\F_k}|}{\sum\limits_{j\in \basisset}|{\F_j}|} \eigenV_k(\x) \otimes \eigenV_k(\Xobsb) \left(\sum\limits_{k\in\basisset} \frac{\kappa|{\F_k}|}{\sum\limits_{j\in \basisset}|{\F_j}|} \eigenV_k(\Xobsb)\otimes \eigenV_k(\Xobsb) + \lambda\Id_\Nobs\right)^{-1} \Yobsb \\
            &= \sum\limits_{k\in\basisset} \frac{|{\F_k}|}{\sum\limits_{j\in \basisset}|{\F_j}|} \eigenV_k(\x) \otimes \eigenV_k(\Xobsb)  \left(\sum\limits_{k\in\basisset} \frac{|{\F_k}|}{\sum\limits_{j\in \basisset}|{\F_j}|} \eigenV_k(\Xobsb) \otimes \eigenV_k(\Xobsb)  + \frac{\lambda}{\kappa}\Id_\Nobs\right)^{-1} \Yobsb\,.
        \end{aligned}
    \end{equation*}
So, one can see that the SKRR approximation only depends on the ratio between the nugget $\lambda$ and the parameter $\kappa$. Likewise for the prediction variance of \Cref{eq:sigma_GPR}, one has:
\begin{align*}
        \begin{split}
            \sigma^2(\x)    &= \Kers(\x,\x) - \Kerbs(\x, \Xobsb)\left(\Kerbs(\Xobsb, \Xobsb) + \lambda\Id_\Nobs\right)^{-1}\Kerbs(\Xobsb, \x) \\
                            &= \kappa \Biggl[\sum\limits_{k\in\basisset} \frac{|{\F_k}|}{\sum\limits_{j\in \basisset}|{\F_j}|}  \eigenV_k(\x) \otimes \eigenV_k(\x) - \sum\limits_{k\in\basisset} \frac{|{\F_k}|}{\sum\limits_{j\in \basisset}|{\F_j}|}  \eigenV_k(\x) \otimes \eigenV_k(\Xobsb)\\ &\mathrel{\phantom{=}} \times\left(\sum\limits_{k\in\basisset} \frac{|{\F_k}|}{\sum\limits_{j\in \basisset}|{\F_j}|} \eigenV_k(\Xobsb) \otimes \eigenV_k(\Xobsb) + \frac{\lambda}{\kappa}\Id_\Nobs\right)^{-1}\sum\limits_{k\in\basisset} \frac{|{\F_k}|}{\sum\limits_{j\in \basisset}|{\F_j}|}  \eigenV_k(\Xobsb) \otimes \eigenV_k(\x)\Biggr]\,.
        \end{split}
\end{align*}
The parameter $\kappa$ fixing the trace of the integral operator with kernel $\Kers$ can be understood as a scaling factor on the prediction variance. At last, when the nugget $\lambda$ vanishes one arrives at:
\begin{equation}\label{eq:SORS}
\FapproxNs{}(\x)=\sum_{j\in\basisset}\sum_{k\in\basisset}\frac{\absolute{\F_k}}{\absolute{\F_j}}\lrangle{\eigenV_j(\Xobsb),\eigenV_k(\Xobsb)}_{\Nobs}\lrangle{\eigenV_j(\Xobsb),\Yobsb}_{\Nobs}\eigenV_k(\x)\,,
\end{equation}
where $\lrangle{\eigenV_j(\Xobsb),\eigenV_k(\Xobsb)}_{\Nobs}=\sum_{i=1}^\Nobs\eigenV_j(\Xobsb_i)\eigenV_k(\Xobsb_i)$ and $\lrangle{\eigenV_j(\Xobsb),\Yobsb}_{\Nobs}=\sum_{i=1}^\Nobs\eigenV_j(\Xobsb_i)\Yobs_i$.

{\scriptsize
\begin{algorithm}[h!]
%\TitleOfAlgo{Kernel Ridge Regression through $\ell_1$-minimization.}
    \SetAlgoLined
    \KwIn{Basis $\left\{\eigenV_k\right\}_{k\in\basisset}$, $\Nobs$ observations of $\F$, parameters $\epsimax$, $\lambda$ and $\kappa$.}
    \KwOut{SSKRR approximation $\FapproxNs{\lambda}$ and prediction variance $\sigma^{2}(\x)$.}
    Build the observation matrix $\bm{\Theta}$ from the basis $\left\{\eigenV_k\right\}_{k\in\basisset}$\;
    Solve the BPDN minimization to find the projection coefficients $\cstarb$\string: $\cstarb = \argmin\limits_{\mathbf{h}\in\Rset^{\basisnumb}} \norm{\mathbf{h}}{1} \text{ subjected to } \norm{\bm{\Theta} \mathbf{h} - \Yobsb}{2} \leq \epsimax$\;
    Solve $ \bm{\eigenv}^{\star} = \argmin\limits_{\{\eigenv_{k}\}_{k\in\basisset}} \norm{\fapprox}{\RKHS}^{2} = \argmin\limits_{\{\eigenv_{k}\}_{k\in\basisset}} \sum\limits_{k\in\basisset}\frac{{\cstar_k}^2}{\eigenv_k}$ subjected to $\sum\limits_{k\in\basisset} \eigenv_{k} = \kappa$ following \Cref{eq:finish_lagrange}\;
    Form the new kernel as $\Kers(\x,\y) = \sum\limits_{k\in\basisset} \eigenv_{k}^{\star} \eigenV_k(\x)\otimes\eigenV_k(\y)$\;
    Obtain the SSKRR approximation and the prediction variance at an unobserved point $\x$ by:
	\begin{equation} \label{eq:SSKRR_approx}
	\begin{split}
	\FapproxNs{\lambda}(\x) & = \Kerbs(\x, \Xobsb) \left(\Kerbs(\Xobsb, \Xobsb) + \lambda\Id_\Nobs\right)^{-1} \Yobsb\,, \\
	 \sigma^2(\x) &= \Kers(\x,\x) - \Kerbs(\x, \Xobsb)\left(\Kerbs(\Xobsb, \Xobsb) + \lambda\Id_\Nobs\right)^{-1}\Kerbs(\Xobsb, \x)\,.
	\end{split}
	\end{equation}
   
    \caption{Sparse Spectral Kernel Ridge Regression (SSKRR).}
    \label{algo:SKRR_sparse}
\end{algorithm}
}

%\clearpage

\subsection{Non-sparse SKRR algorithm}\label{sec:NSKRR}

If the ground truth function $\F$ is not sparse on the basis $\Basis^{\basisnumb}$, one can compute the expansion coefficients $\mathbf{c}$ by projection since $\Basis^{\basisnumb}$ is orthonormal. We propose the following procedure sketched in \Cref{algo:SKRR_non_sparse} where projections are carried out iteratively using iterated surrogate approximations to mimic the ground truth function $\F$. We call this algorithm Non-sparse Spectral Kernel Ridge Regression (NSKRR).

{\scriptsize
\begin{algorithm}[h!]
%\TitleOfAlgo{Kernel Ridge Regression through reconstruction by interpolation.}
    \SetAlgoLined
    \KwIn{Basis $\left\{\eigenV_k\right\}_{k\in\basisset}$, eigenvalues $\{\eigenv_{k}^{(0)}\}_{k\in\basisset}$, $\Nobs$ observations of $\F$, parameters $\lambda$ and $\kappa$, and the number of iterations $\Nstep$.}
    \KwOut{NSKRR approximation $\GapproxNLamb{\lambda}^{(\Nstep)}(\x)$ and prediction variance $\sigma^{2}(\x)$.}
    Initialization\string: At step $n = 0$, define the initial kernel as $\Ker^{(0)}(\x,\y) = \sum\limits_{k\in\basisset} \eigenv_{k}^{(0)} \eigenV_k(\x)\otimes\eigenV_k(\y)$\;
    \For{$n \gets 1$ \KwTo $\Nstep$}{
    Approximate $\F$ by its NSKRR approximation: $\GapproxNLamb{\lambda}^{(n-1)}(\x) = \Kerb^{(n-1)}(\x, \Xobsb) \left(\Kerb^{(n-1)}(\Xobsb, \Xobsb) + \lambda\Id_\Nobs\right)^{-1} \Yobsb$\;
    \For{$k\in\basisset$}{
    Compute $c_{k}^{(n-1)} = \lrangle{\smash{{\GapproxNLamb{\lambda}^{(n-1)}}, \eigenV_k}}_{\Ltwo}$\;
    }
    Solve $ \bm{\eigenv}^{(n)} = \argmin\limits_{\{\eigenv_{k}\}_{k\in\basisset}} \norm{\smash{\GapproxNLamb{\lambda}^{(n-1)}}}{\RKHS}^{2} =  \argmin\limits_{\{\eigenv_{k}\}_{k\in\basisset}} \sum\limits_{k\in\basisset}\frac{(c_{k}^{(n-1)})^2}{\eigenv_k}$ subjected to $\sum\limits_{k\in\basisset} \eigenv_{k} = \kappa$ \;
    Form the new kernel as $\Ker^{(n)}(\x,\y) = \sum\limits_{k\in\basisset} \eigenv_{k}^{(n)} \eigenV_k(\x)\otimes\eigenV_k(\y)$\;
    }
    Obtain the NSKRR approximation and the prediction variance at an unobserved point $\x$ by:
	\begin{equation}\label{eq:NSKRR_approx}
	\begin{split}
	\GapproxNLamb{\lambda}^{(\Nstep)}(\x) &= \Kerb^{(\Nstep)}(\x, \Xobsb) \left(\Kerb^{(\Nstep)}(\Xobsb, \Xobsb) + \lambda\Id_\Nobs\right)^{-1} \Yobsb\,, \\
        (\sigma^{(\Nstep)}(\x))^{2} &= \Ker^{(\Nstep)}(\x,\x) - \Kerb^{(\Nstep)}(\x, \Xobsb)\left(\Kerb^{(\Nstep)}(\Xobsb, \Xobsb) + \lambda\Id_\Nobs\right)^{-1}\Kerb^{(\Nstep)}(\Xobsb, \x)\,.
        \end{split}
	\end{equation}
   
    \caption{Non-sparse Spectral Kernel Ridge Regression (NSKRR).}
    \label{algo:SKRR_non_sparse}
\end{algorithm}
}

It is worth mentioning at this stage that the SKRR algorithms we propose here are not competing with the KF algorithms of \Cref{sec:chap3_KF}. Actually the nugget $\lambda$ in \Cref{eq:SSKRR_approx} or \Cref{eq:NSKRR_approx} could be determined by a parametric KF algorithm once a "best" kernel is found in the sense of \pref{eq:best_min_Kers}. This is the approach retained in the numerical example of \Cref{subsec:example_UMRIDA_3D} below. \amend{Also the minimal conditions on kernels for our algorithms to apply are the ones stated in \Cref{df:MercerKernel} of the appendix \ref{subsec:mercer}. Non-smooth triangular kernels, for example, fulfill these conditions. Our approach does not require any explicit knowledge of a kernel function, though.}

%-----------% Section PCE %-----------%
\subsection{Polynomial Chaos Expansion}\label{subsec:PCE}

The remaining question is\string: how to choose a basis $\{\eigenV_i\}_{i\in\mathbb{N}}$ of $\Ltwo(\InputSpace, \mu)$? In this section, we focus on orthonormal polynomial bases. Such kind of representations are referred to as Polynomial Chaos (PC) expansions in the case where $\mu$ is a Gaussian probability measure \cite{Ghanem91, Wiener38}. For more general probability measures, they are called generalized Polynomial Chaos (gPC) expansions \cite{Ernst12, LeMaitre10, Soize04, Xiu02}. \amend{gPC expansions with non-Gaussian probability measures are considered in the synthetic numerical examples of \Cref{sec:numerical_ex} and the aerodynamic example of \Cref{subsec:example_UMRIDA_3D} below for comparisons with our proposed algorithms.} Besides, the use of polynomial bases with Gaussian process regression is illustrated in \emph{e.g.} \cite{Yan18}. \amend{We also note that mixtures of models may be worth considering to improve the accuracy of surrogates, as in Polynomial Chaos-Kriging for example \cite{Schobi2015}. We leave that possibility to future works.}

In the context of gPC expansions, the input variables $\Xobsb$ with values in $\InputSpace \subset \Rset^{\Dim}$ are assumed to be mutually independent random variables with probability distribution $\mu$. One can then build a gPC surrogate model $\FapproxPCE$ of $\F$ by a standard $\Ltwo$ projection on a finite dimensional subspace of $\Ltwo(\InputSpace, \mu)$ spanned by a truncated orthonormal family of $\Dim$-variate polynomials up to total order $p$ denoted by $\smash{\Basis^\basisnumb}=\smash{\{\phi_\mathbf{i}\}_{\norm{\mathbf{i}}{1}\leq p}}$, where $\basisnumb=\smash{\binom{p+\Dim}{p}}$.
Here $\mathbf{i}=\smash{(i_1,i_2,\dots i_\Dim)}$ is a multi-index in $\smash{\Nset^\Dim}$ and $\smash{\norm{\mathbf{i}}{1}} = \smash{\sum_{j=1}^{\Dim}i_j}$. These $\Dim$-variate polynomials read:
\begin{equation}\label{eq:phi_multi}
\phi_\mathbf{i}(\x) = \prod\limits_{j=1}^{\Dim}\phi_{i_j}^{(j)}(x_j)\,,
\end{equation}
where $\x = \smash{(x_1, x_2, \dots x_\Dim)}$, and $\smash{\{\phi^{(j)}_i\}_{i\in\Nset}}$ are the univariate orthonormal polynomials with respect to the law of the $j$-th input variable. Renumbering the $ \basisnumb$ polynomials in the truncated family $\Basis^\basisnumb$ with a single index $k\in\basisset=\{0\leq k\leq \basisnumb-1\}$, the orthonormality condition reads:
\begin{equation}\label{eq:ortho_property}
\lrangle{\phi_j,\phi_k}_{\Ltwo}= \int_\InputSpace\phi_j(\x)\phi_k(\x)\mu(d\x) = \delta_{jk}\,,\quad j,k\in\basisset\,,
\end{equation}
where $\delta_{jk}$ is the Kronecker symbol such that $\delta_{jk} = 1$ if $j = k$, and $\delta_{jk} = 0$ otherwise. Consequently, the gPC surrogate model $\FapproxPCE$ using the truncated basis $\Basis^\basisnumb$ reads:
\begin{equation}\label{eq:G_PCE}
\FapproxPCE(\x) = \sum_{k\in\basisset}\F_k\phi_k(\x) = \sum_{k\in\basisset}\lrangle{\F,\phi_k}_{\Ltwo} \phi_k(\x)\,.
\end{equation}
Such an expansion is amenable to a direct computation of moments, \emph{e.g.} expectation and variance: $\esp(\FapproxPCE)=\lrangle{\FapproxPCE,1}_{\Ltwo}=\F_0$, and
\begin{equation}\label{eq:gPCmoments}
\var(\FapproxPCE)=\lrangle{\FapproxPCE-\esp(\FapproxPCE),\FapproxPCE-\esp(\FapproxPCE)}_{\Ltwo}=\sum_{k=1}^{\basisnumb-1}\F_k^2\,,
\end{equation}
resorting to the orthonormality condition \pref{eq:ortho_property}. Higher-order moments and Sobol' sensitivity indices can be computed along the same lines \cite{Savin17,Sudret08}.

This gPC surrogate can be compared to the surrogate \pref{eq:SORS} with $\eigenV_k\equiv\phi_k$:
\begin{equation*}
\FapproxNs{}(\x)=\sum_{j\in\basisset}\sum_{k\in\basisset}\frac{\absolute{\F_k}}{\absolute{\F_j}}\lrangle{\phi_j(\Xobsb),\phi_k(\Xobsb)}_{\Nobs}\lrangle{\phi_j(\Xobsb),\Yobsb}_{\Nobs}\phi_k(\x)\,.
\end{equation*}
Both approaches are considered in the numerical examples below, using however a nugget $\lambda>0$. In gPC expansion, the expansion coefficients $\F_k = \smash{\lrangle{\F,\phi_k}_{\Ltwo}}$ are usually computed by a numerical quadrature rule:
\begin{equation}\label{eq:quad_nodes}
    \F_k \simeq \F_{k,q} = \sum_{l=1}^{q} \weight_l \F(\node_l)\phi_k(\node_l)\,,
\end{equation}
where $\{\weight_l\}_{l=1}^{q}$ are positive weights and $\{\node_l\}_{l=1}^{q}$ are nodes in $\InputSpace$. The number of nodes $q$ that is needed depends on the selected rule. A classical Gauss quadrature rule requires $q$ nodes to exactly integrate univariate polynomials up to order $2q - 1$. If one quadrature node is fixed, a Gauss-Radau (GR) rule is obtained, which exactly integrates univariate polynomials up to order $2q - 2$. If two quadrature nodes are fixed, a Gauss-Lobatto (GL) rule is obtained, which exactly integrates univariate polynomials up to order $2q- 3$. Thus using quadrature rules one needs about $\Nobs\approx\floor{\frac{p}{2}}^{\Dim}$ sampling points to exactly integrate $\Dim$-variate polynomials of total order $p$. For complex models with high dimensional input spaces, the expansion \pref{eq:G_PCE} can be unaffordable; this is the so-called curse of dimensionality. Sparse quadrature rules based on Smolyak's algorithm can be used to circumvent this limitation \cite{Smolyak63}. In practical examples though, the ground truth function $\F$ is often sparse or nearly sparse owing to a ``sparsity-of-effects'' principle \cite{Montgomery2004} whereby the vector $\mathbf{c} = \smash{(c_k)_{k\in\basisset}} \equiv \smash{(\F_k)_{k\in\basisset}}$ of the expansion coefficients of the polynomial surrogate $\FapproxPCE$ has many negligible components \cite{Chkifa15, Rabitz99}. In these situations, they can be evaluated within the framework of compressed sensing outlined in \Cref{sec:chap3_sparse_recons}. This is the approach retained in \emph{e.g.} \cite{Doostan11, Mathelin12}; see also \cite{Hampton16} and references therein, or \cite{Savin16} for an application to aerodynamics.

%-----------% Section Numerical examples %-----------%

\section{Synthetic numerical examples}\label{sec:numerical_ex}

The foregoing algorithms are first applied on two synthetic test functions: the three-dimensional ($\Dim = 3$) Ishigami function in \Cref{subsec:example_Ishigami}, and the ten-dimensional ($\Dim = 10$) Rosenbrock function in \Cref{subsec:example_Rosenbrock}. %Then we consider a complex aerodynamic test case, namely a two-dimensional airfoil in transonic flow with three geometrical and operational variable inputs ($\Dim=3$), in \Cref{subsec:example_UMRIDA_3D}.
For these examples and for comparison purposes, a surrogate $\FapproxN$ is built using actually four different methods: (i) a fully tensorized gPC surrogate model \pref{eq:G_PCE} where the expansion coefficients are obtained by fully tensorized GL quadrature nodes in \pref{eq:quad_nodes}; (ii) a sparse gPC surrogate model \pref{eq:G_PCE} where the expansion coefficients are obtained by solving the problem \pref{eq:BPDN}; (iii) a classical KRR surrogate model \pref{eq:solution_KRR} using a Gaussian kernel \pref{eq:Gaussian_kernel_ARD}; and finally (iv) a SSKRR surrogate model obtained by \Cref{algo:SKRR_sparse}.

The $\Nobs$ observations of the ground truth function $\F$ (Ishigami or Rosenbrock) are obtained by Latin Hypercube Sampling (LHS) with a minimax criterion using the $\python$ package \smt\ \cite{Bouhlel19}. For the fully tensorized gPC surrogate model, the number of observations $\Nobs$ is chosen in order to exactly integrate the orthonormality condition of \Cref{eq:ortho_property} for polynomials of total order $p$. We recall that given $q$ nodes, the GL quadrature rule exactly integrates univariate polynomials of order $2q-3$. The surrogate models are subsequently tested on a test set consisting of $\NT$ observations. Therefore, we have two sets:
\begin{itemize}
    \item The learning set which consists of $\Nobs$ observations: $\left( \Xobsb, \Yobsb = \F(\Xobsb) \right)$;
    \item The test set which consists of $\NT$ observations: $\left( \Xobsb_{\NT}, \Yobsb_{\NT} = \F(\Xobsb_{\NT}) \right)$.
\end{itemize}
The test set can be understood as an unseen set and it is used to validate the surrogate models. For both test functions, we replicate the four surrogate models through ten independent runs. As we are considering synthetic functions, we are not limited in the choice of the size of the test set. Therefore, we choose $\NT = 1\times10^{5}$ observations taken at random. One can notice that we do not use a validation set, because we know the ground truth functions $\F$ and we assume that the test set is large enough to ensure generalization. 

The performance of each surrogate model $\FapproxN$ is quantified by computing the empirical Normalized Root Mean Square Error $\NRMSE$ defined by:
\begin{equation}\label{eq:error_NRMSE}
    \NRMSE = \sqrt{\frac{\sum\limits_{i=1}^{\NT}\left(\Yobs_i - \FapproxN(\Xobsb_i)\right)^{2}}{\sum\limits_{i=1}^{\NT}\Yobs_i^2}}\,,
\end{equation}
and the empirical Root Mean Square Error $\RMSE$ defined by:
\begin{equation}\label{eq:error_RMSE}
    \RMSE = \sqrt{\frac{1}{\NT}\sum\limits_{i=1}^{\NT}\left(\Yobs_i - \FapproxN(\Xobsb_i)\right)^{2}}\,.
\end{equation}
The only difference between $\NRMSE$ and $\RMSE$ is that $\RMSE$ is divided by $\sum_{i=1}^{\NT}\Yobs_i^2$ in order to remove any scaling factor of the ground truth function $\F$. In addition, the prediction coefficient $\Qtwo$, or coefficient of determination \cite{Blatman2011}, defined by\string:
\begin{equation}
    \Qtwo = 1 - \frac{\sum\limits_{i=1}^{\NT}\left(\Yobs_i - \FapproxN(\Xobsb_i)\right)^2}{\sum\limits_{i=1}^{\NT}\left( \Yobs_i - \displaystyle\frac{1}{\NT} \sum\limits_{i=1}^{\NT} \Yobs_i \right)^2}
\end{equation}
is computed. A prediction coefficient $\Qtwo$ close to one indicates that the surrogate model is accurate over the $\NT$ test samples. In other words, the closer $\Qtwo$ is to one, the more accurate the surrogate model is. All the three metrics above are computed on the test set. The results concerning the errors $\RMSE$ and $\NRMSE$ are presented using box plots. In more details, the central horizontal line is the median value over the ten independent runs, and the edges of the boxes correspond to the 25th $q_{25}$ and 75th $q_{75}$ percentiles. The circles are the outliers defined as being either smaller than $q_{25} - 1.5(q_{75} - q_{25})$, or larger than $q_{75} + 1.5(q_{75} - q_{25})$.

Here in this paper, the Spectral Projected Gradient Algorithm (SPGL1) developed by van den Berg and Friedlander in $\python$ \cite{Berg08, Berg11} is considered in order to compute the solution of \pref{eq:BPDN}. This algorithm is based on primal-dual interior point methods. In order to tune the nugget $\lambda$ of the kernel of the SSKRR surrogate \pref{eq:SSKRR_approx} in \Cref{algo:SKRR_sparse}, we use a grid search algorithm on the $\NT$ data points with the error $\RMSE$ as the metric. That is, we compute $\RMSE$ with respect to $\lambda$ and then select the parameter $\lambda = \lambdamin$ corresponding to the minimum of $\RMSE$. This can be done because the ground truth function $\F$ is known and inexpensive to evaluate. In addition, the parameter $\kappa$ of the kernel of the SSKRR surrogate is set to $\kappa = \var(\Yobsb)$. Finally, the parameters $\bm{\theta} = (\lambda_{\mathrm{KRR}}, \{\gamma_i\}_{i=1}^{\Dim})$ of the Gaussian kernel of the KRR surrogate are determined using the parametric KF algorithm presented in \Cref{subsec:chap3_P_KF_algo}. %\Cref{subsubsec:KF_MSES_param}.
% ######################################

\subsection{Ishigami function}\label{subsec:example_Ishigami}

The Ishigami function \cite{Ishigami99} is commonly used for benchmarking global sensitivity analyses and uncertainty quantification. The analytic expression of this three-dimensional ($\Dim=3$) function is:
\begin{equation}\label{eq:Ishigami_func}
    \F(\Xobsb) = \sin(\Xobs_1) + a\sin^2(\Xobs_2) + b\Xobs_3^4\sin(\Xobs_1),
\end{equation}
with $a = 7$, $b = 0.1$ \cite{Marrel09}, and $\Xobsb = (\Xobs_1, \Xobs_2, \Xobs_3) \in [-\pi, \pi]^{3}$. The input variables $\Xobsb$ are assumed to be mutually independent and uniformly distributed:
\begin{equation}\label{eq:Ishigami_law}
    \Xobs_i \sim \mathcal{U}(-\pi, \pi)\,,\quad i = 1, 2, 3\,.
\end{equation}
The expectation and variance of $\F$ are given by:
\begin{equation}
    \esp(\F) = \frac{a}{2}\,,\quad \var (\F) = \frac{1}{2} + \frac{a^2}{8} + \frac{b^2\pi^8}{18} + \frac{b\pi^4}{2}\,.
\end{equation}
We select polynomials of total order up to $p = 10$ to form the polynomial basis $\Basis^\basisnumb$, which corresponds to $\basisnumb={p + \Dim \choose \Dim} = {10+3 \choose 3} = 286$ multi-dimensional Legendre polynomials. The latter are indeed orthonormal with respect to the uniform probability distribution. $\Basis^\basisnumb$ is considered for the construction of the fully tensorized gPC, the sparse gPC, and the SSKRR surrogates. For the fully tensorized gPC surrogate model, $q = 12^3 = 1728$ GL quadrature nodes are needed to exactly recover the orthonormality condition of \Cref{eq:ortho_property} since we have chosen a total order $p = 10$. For the sparse gPC and the SSKRR surrogates, two learning sets with $\Nobs = 50$ and $\Nobs = 100$ observations of the ground truth function $\F$ are considered to test the influence of $\Nobs$ on the recovery of the expansion coefficients by the BPDN minimization \pref{eq:BPDN}. The values $\Nobs = 50$ and $\Nobs = 100$ are chosen because they are significantly lower than the size of the polynomial basis $\Basis^\basisnumb$, while $\Nobs = 100$ is shown to yield stable solutions of \pref{eq:BPDN} in \Cref{sec:sparsity_Ishigami} below. Also $\epsimax = 1\times10^{-6}$ is chosen there. Finally, the KRR surrogate is built using a learning set with $\Nobs = 100$ observations.

\subsubsection{Sparsity on Legendre polynomials}\label{sec:sparsity_Ishigami}

\begin{figure}[h!]
    \centering
    \subfigure[$\Nobs = 50$ observations.]{\includegraphics[width=0.45\textwidth]{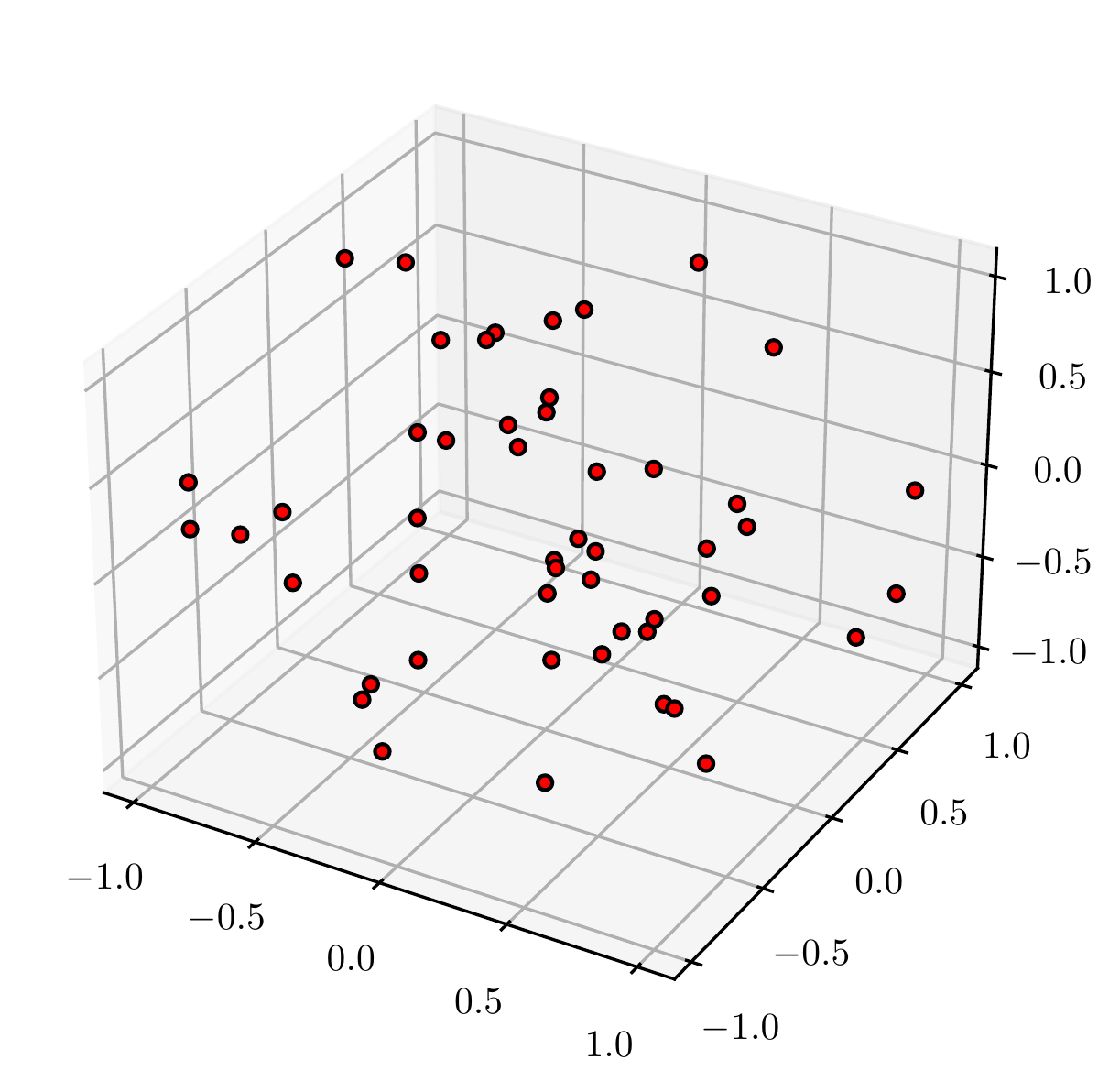} \label{subfig:Ishigami_LHS_seed_0_50obs}}
    \subfigure[$\Nobs = 100$ observations.]{\includegraphics[width=0.45\textwidth]{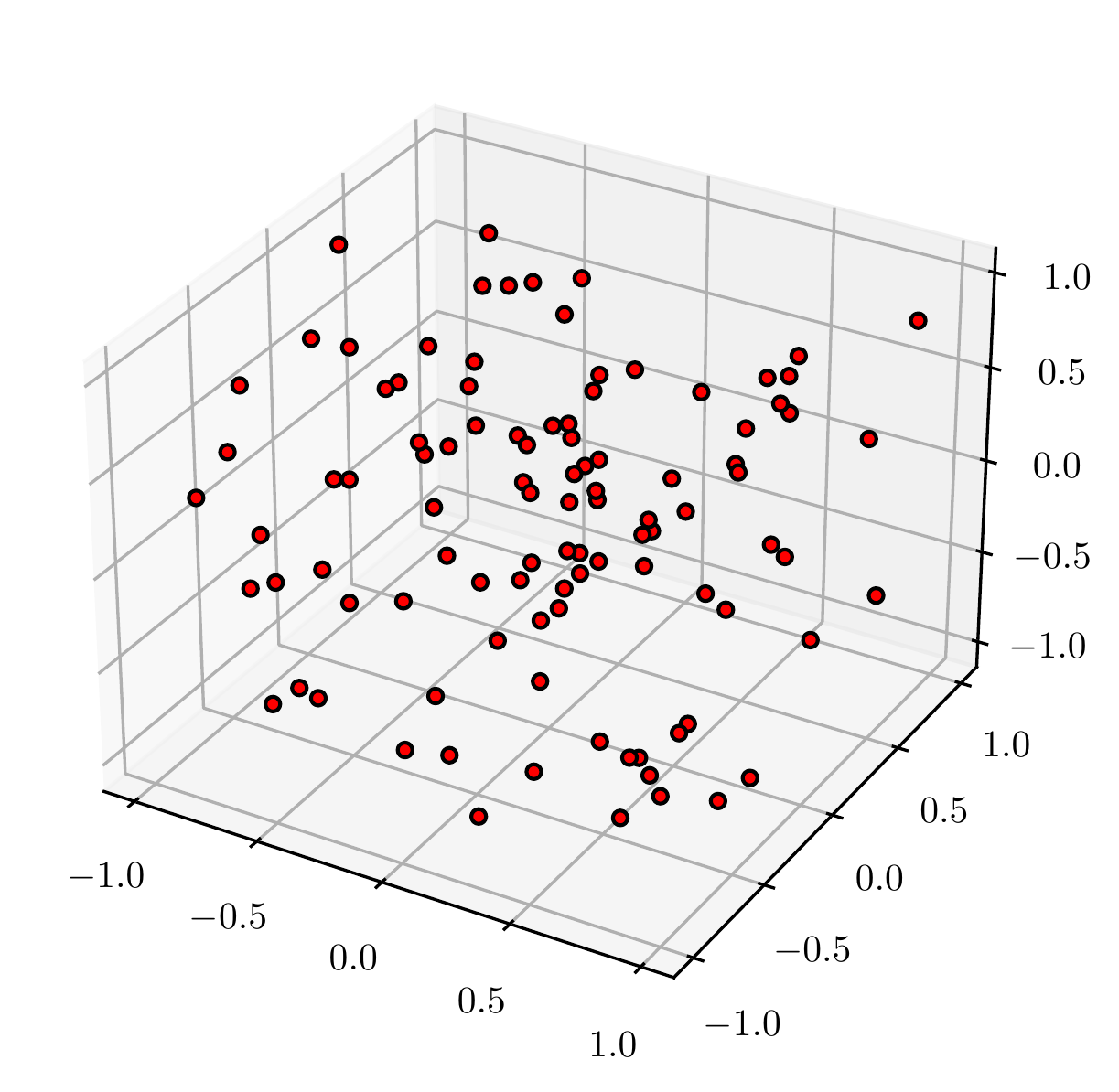} \label{subfig:Ishigami_LHS_seed_0_100obs}}
    \caption{One run of the LHS positions with (a) $\Nobs = 50$ and (b) $\Nobs = 100$ observations of $\F$.}
    \label{fig:Ishigami_LHS_seed_0}
\end{figure}

The first step is to determine the sparsity as it is observed \emph{a posteriori}. In that respect, we increase the number of observations in the learning set from $\Nobs = 50$ to $\Nobs = 100$ and keep track of the expansion coefficients $\cstarb$ solving \Cref{eq:BPDN} that do not significantly change over ten independent runs of the positions of the observations in either set. One run of the positions of the observations can be seen on \Cref{fig:Ishigami_LHS_seed_0} for $\Nobs = 50$ and $\Nobs = 100$ observations of $\F$.

\begin{figure}
	\centering
	\includegraphics[width=0.8\textwidth]{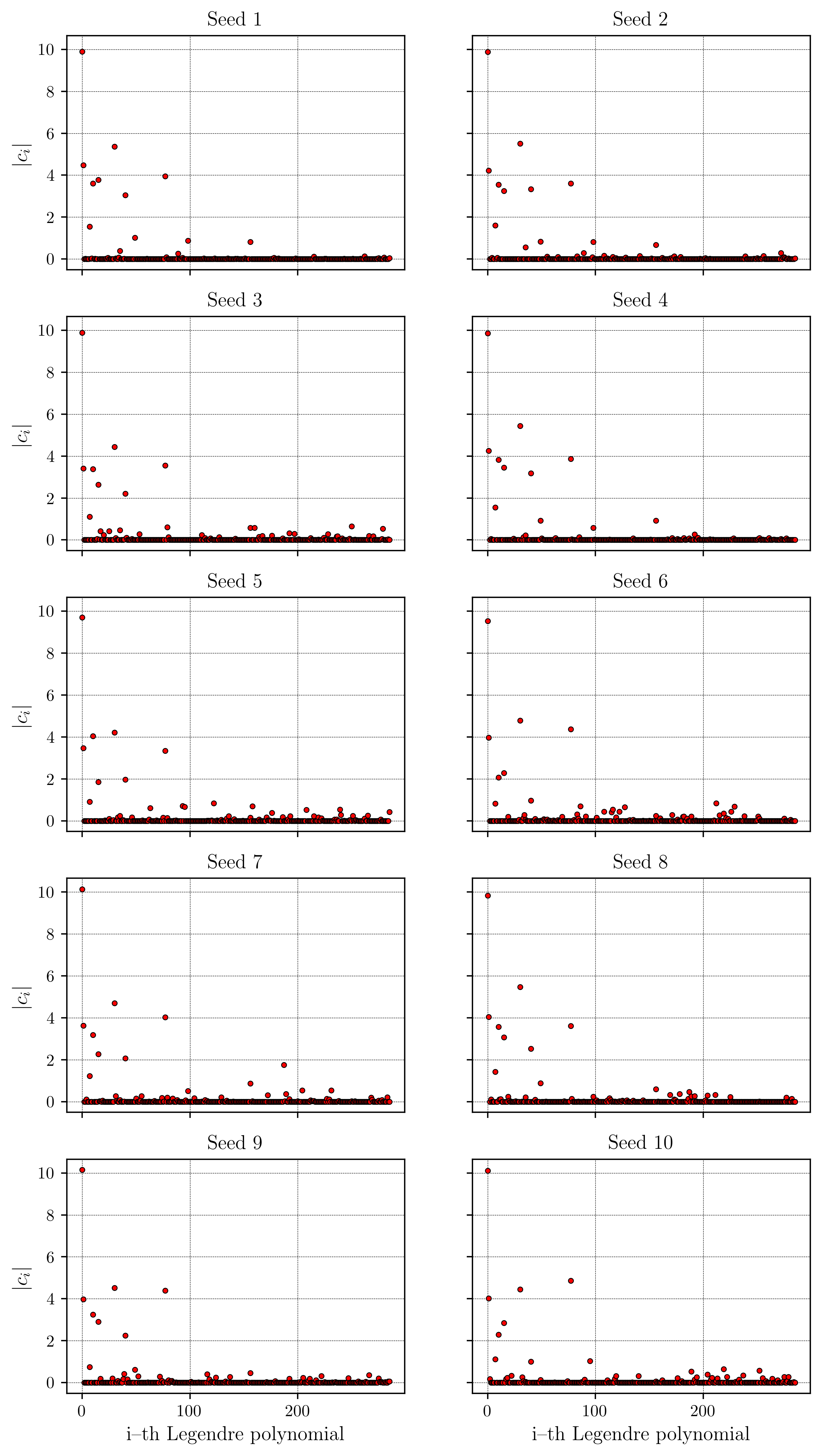}
	\caption{Evolution of the expansion coefficients $\cstarb$ with respect to the random samplings of the positions of the observations with $\Nobs = 50$ observations of $\F$.}
	\label{fig:Ishigami_cstar_50}
\end{figure}

\begin{figure}
	\centering
	\includegraphics[width=0.8\textwidth]{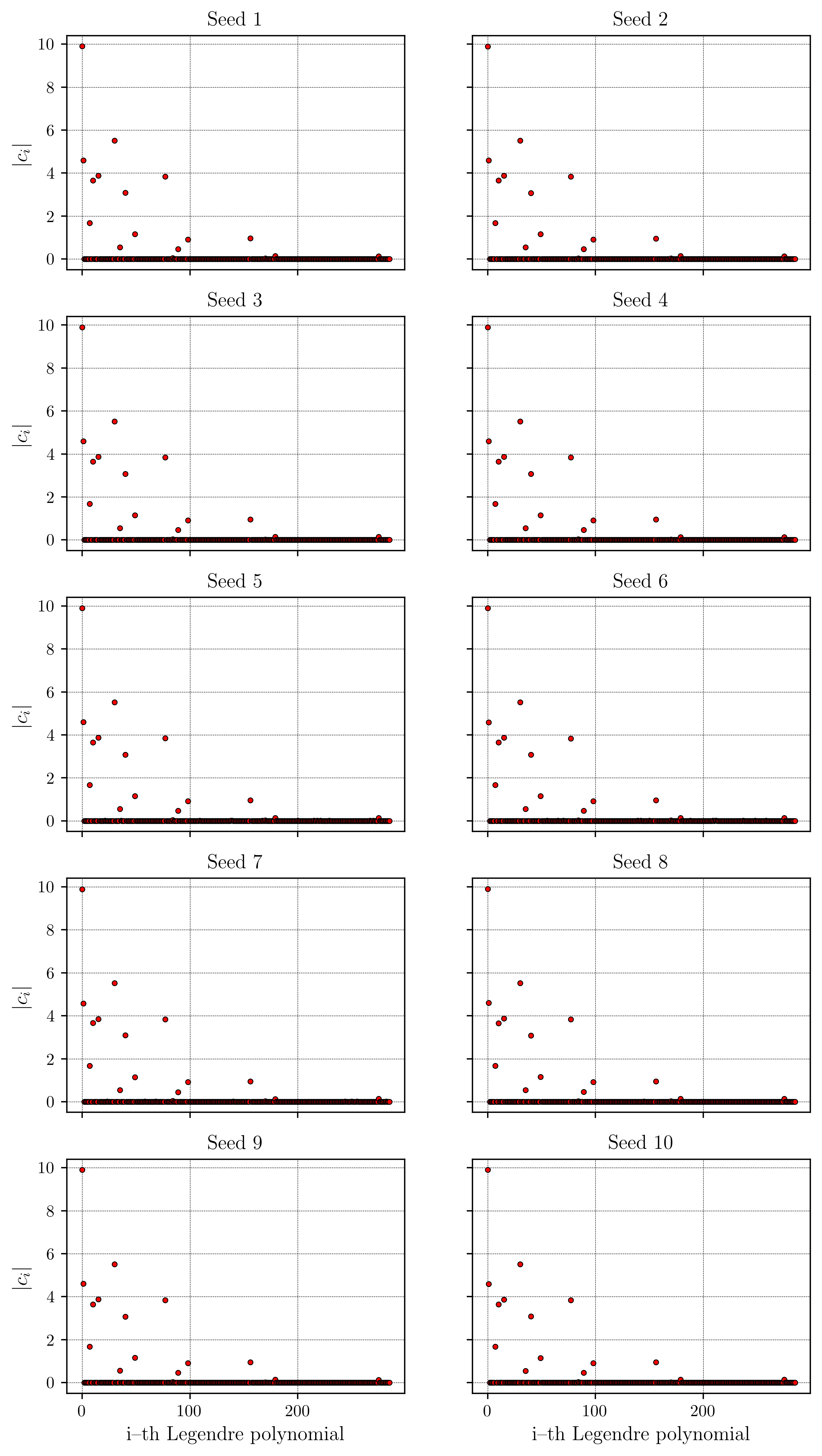}
	\caption{Evolution of the expansion coefficients $\cstarb$ with respect to the random samplings of the positions of the observations with $\Nobs = 100$ observations of $\F$.}
	\label{fig:Ishigami_cstar_100}
\end{figure}

The evolution of the expansion coefficients $\cstarb$ with respect to the random sampling of the positions of the observations are shown on \Cref{fig:Ishigami_cstar_50} and \Cref{fig:Ishigami_cstar_100}, with $\Nobs = 50$ and $\Nobs = 100$ observations of $\F$ respectively. One can notice that for $\Nobs = 50$, the coefficients $\cstarb$ are fluctuating greatly from one sampling to another; see for instance the difference between the first and fifth seeds. Now, looking at \Cref{fig:Ishigami_cstar_100}, where $\Nobs = 100$ observations are used, one can notice that the coefficients $\cstarb$ are similar from one sampling to another, and that the sparsity is about $\sparsity \approx 15$. %The average of the coherence parameter $\coherence(\bm{\Theta})$ over the ten samplings is then $\coherence(\bm{\Theta})\approx 5.89$. Thus, \Cref{th:noiseless_sampling} yields $\Nobs \gtrsim \Nobss \approx 500$ observations up to a constant $C$.
Note that such a sparsity is expected as the sine function can be well approximated by polynomials. A common observation is that $\Nobs\gtrsim 4\sparsity$ observations are usually enough for a successful recovery of $\cstarb$ by \Cref{eq:BPDN} (see for example \cite{Candes08_intro}). From this rule of thumb and from now on we choose a learning set with $\Nobs = 100$ observations of $\F$ to construct its sparse gPC, KRR, and SSKRR surrogate models. The fully tensorized gPC surrogate is constructed using $q=1728$ quadrature nodes to compute the expansion coefficients.

\subsubsection{Comparisons between the surrogate models}

\begin{figure}[H]
	\centering
    \includegraphics[width=1.0\textwidth]{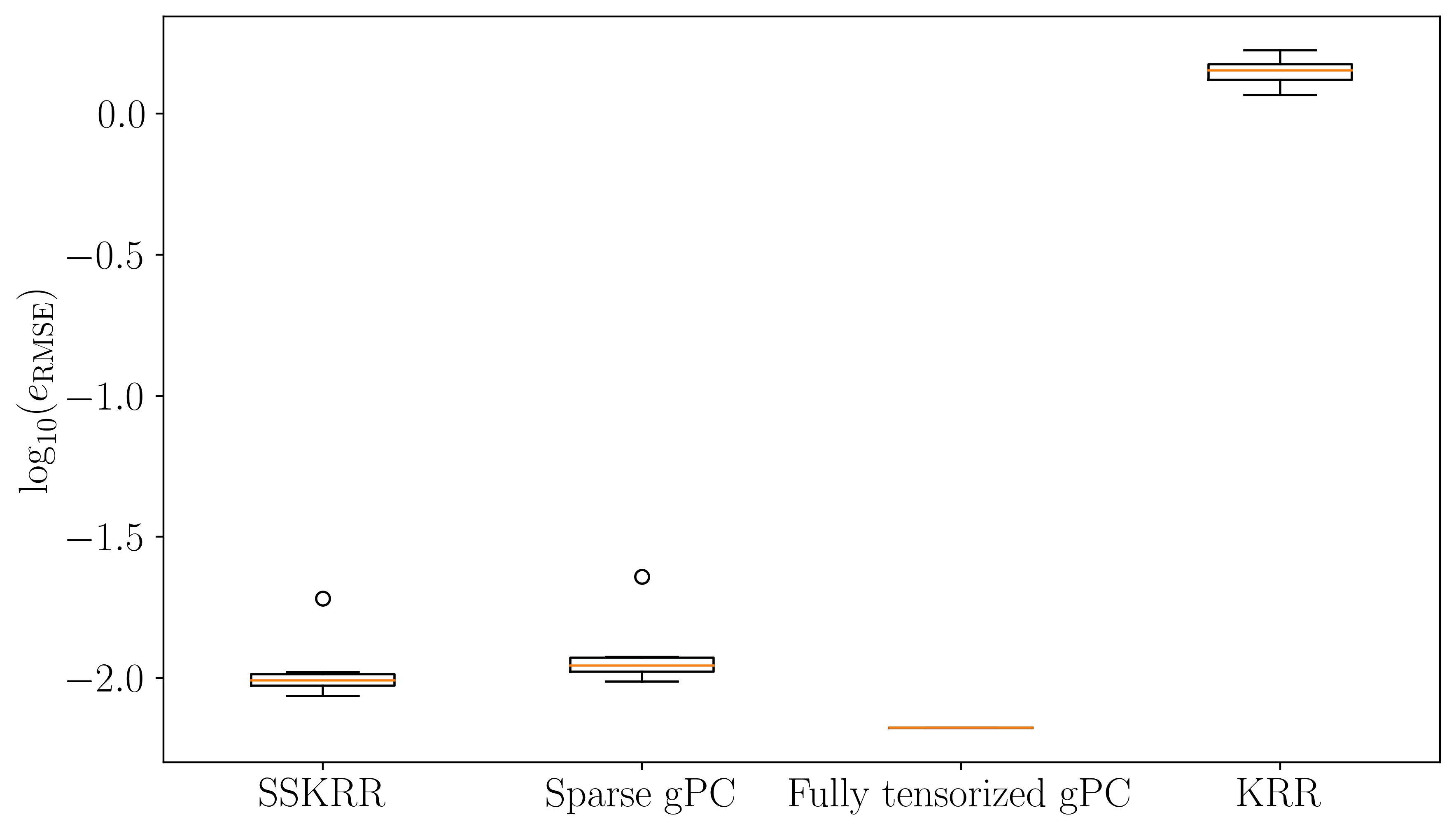}
    \caption{The empirical root mean square error $\RMSE$ over the ten independent runs for the Ishigami function.}
	\label{fig:Ishigami_RMSE}
\end{figure}

\begin{figure}[H]
	\centering
    \includegraphics[width=1.0\textwidth]{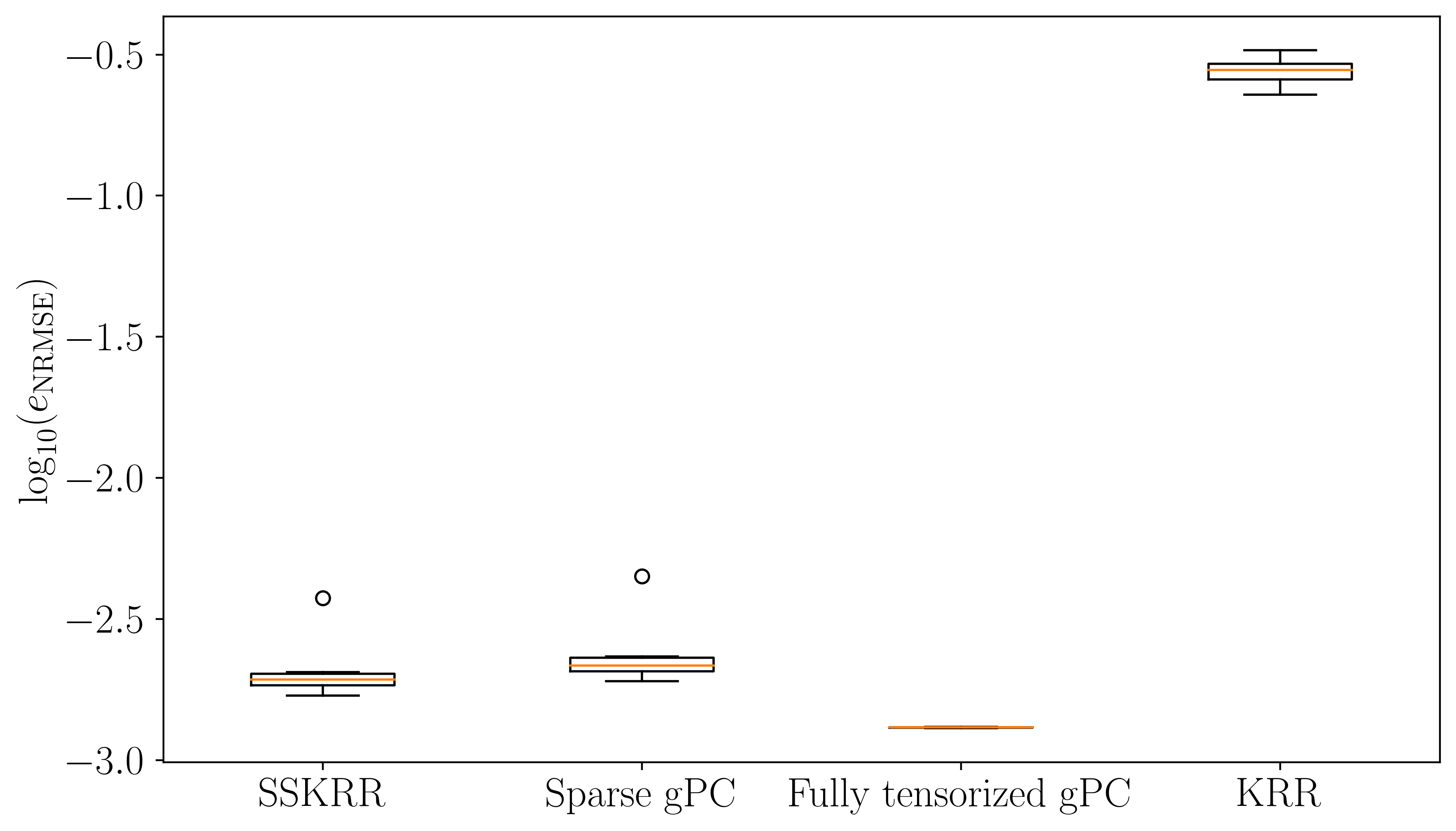}
    \caption{The empirical normalized root mean square error $\NRMSE$ over the ten independent runs for the Ishigami function.}
	\label{fig:Ishigami_NRMSE}
\end{figure}

We now compare the performance of each surrogate model using ten independent runs with $\NT = 1\times10^{5}$ samples in the test set. A comparison of the RMSE error $\RMSE$ and the NRMSE error $\NRMSE$ over the ten independent runs is shown on \Cref{fig:Ishigami_RMSE} and \Cref{fig:Ishigami_NRMSE}. The prediction coefficient $\Qtwo$ for each surrogate model is given on \Cref{tab:Ishigami_Qtwo}. One can see that the surrogate model obtained by SSKRR \Cref{algo:SKRR_sparse} performs slightly better than the sparse gPC surrogate model while the KRR surrogate model performs way worse than the others. The fully tensorized gPC surrogate model ("Full gPC" in \Cref{tab:Ishigami_Qtwo}) performs slightly better than the SSKRR surrogate model but at a much higher computational cost. Indeed, we only used $\Nobs = 100$ observations of $\F$ to obtain the expansion coefficients $\cstarb$ by $\ell_1$-minimization while $q = 1728$ observations of $\F$ are needed to obtain the expansion coefficients for the fully tensorized gPC surrogate model. Notice the circles on \Cref{fig:Ishigami_RMSE} and \Cref{fig:Ishigami_NRMSE}\string: they are outliers. It shows that even with $\Nobs = 100$ observations of $\F$, one can still deviate from recovering the true expansion coefficients. Moreover, the low values of the errors $\RMSE$ and $\NRMSE$ of both approaches can be explained by the fact that the Ishigami function is a smooth function consisting of sine functions, which can be well approximated by polynomials over a bounded domain.

\begin{table}[H]
    \centering
    \begin{tabular} {|c||c|c|c|}
        \hline
        \rowcolor{blue!20} \multicolumn{4}{ |c| }{Ishigami function} \\
        \hline\hline\hline
        \rowcolor{black!20} & Median & Minimum & Maximum \\
            \hline\hline
            Full gPC & $0.999997$ & $0.999997$ & $0.999997$ \\
            \hline
            Sparse gPC & $0.999991$ & $0.999962$ & $0.999993$ \\
            \hline
       		KRR & $0.853325$ & $0.797717$ & $0.902461$ \\
            \hline
            SSKRR & $0.999993$ & $0.999973$ & $0.999995$  \\
        \hline
    \end{tabular}
    \caption{Prediction coefficient $\Qtwo$ over the ten independent runs for the Ishigami function.}
    \label{tab:Ishigami_Qtwo}
\end{table}

\begin{figure}
    \includegraphics[width=0.8\textwidth]{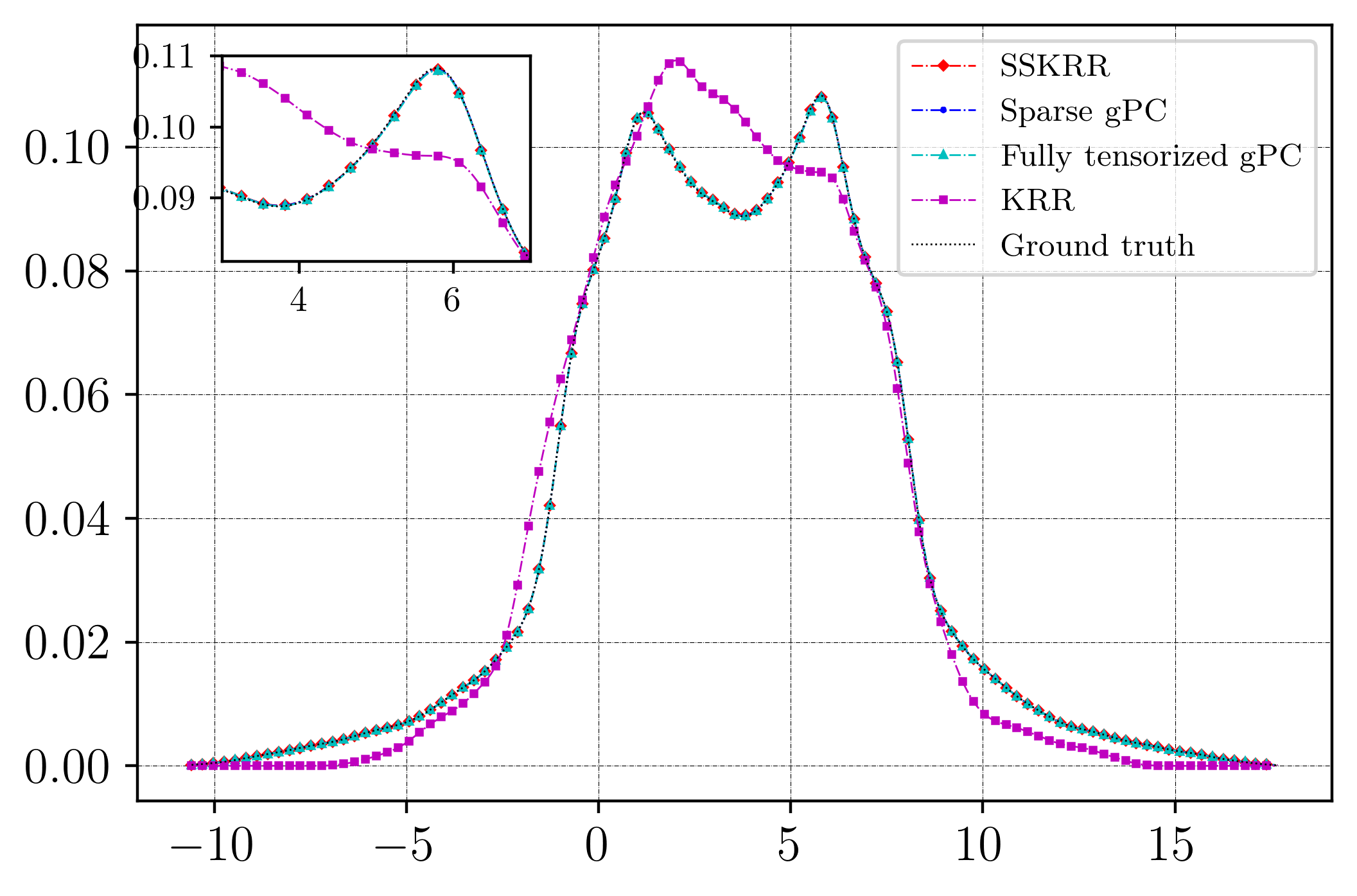}
    \centering
    \caption{PDFs obtained for one run with $\Nobs = 100$ observations of $\F$. The PDFs were obtained from the $\NT= 1\times10^{5}$ data points and then smoothing out by a normal kernel density function.}
    \label{fig:Ishigami_PDF}
\end{figure}

We finally compare several quantities of interest of the surrogate models using the test set over the ten independent runs, namely: the expectation, the variance, and the Kullback-Leibler (KL) divergence. The expectations and variances of the fully tensorized and sparse gPC surrogate models are obtained using the expansion coefficients directly; see \Cref{eq:gPCmoments}. The KL divergence $\DKL$ is computed between each surrogate model and the ground truth function $\F$ by first estimating the PDF from the $\NT = 1\times10^{5}$ observations and then smoothing out the resulting histograms by a normal kernel density function \cite{Wand95}. An example of such PDFs obtained for one run is shown on \Cref{fig:Ishigami_PDF}. \amend{Here the PDFs obtained from the SSKRR, sparse gPC, and fully tensorized gPC surrogates are all superimposed onto the true PDF, whereas the PDF obtained from the KRR surrogate does not fit it well}. The comparison between the surrogate models can be seen on \Cref{fig:Ishigami_MeanVarKL}. One can notice that we obtain roughly the same expectations, variances, and KL divergences with the SSKRR, sparse gPC, and fully tensorized gPC ("Full gPC" in \Cref{fig:Ishigami_MeanVarKL}) surrogate models. These values are close to the exact values; see the last column. Nevertheless, for probabilistic quantities of interest such as the expectation or the variance, the gPC surrogate models perform better because these quantities of interest can be directly computed from the expansion coefficients, while they were estimated from a Monte-Carlo simulation for the KRR and SSKRR surrogates. The KRR surrogate with Gaussian kernel gives the worst results as hinted by the errors $\RMSE$ and $\NRMSE$ computed previously.

\begin{table}
    \centering
    \begin{tabular} {|c||c|c|c||c|}
        \hline
        \rowcolor{blue!20} \multicolumn{5}{ |c| }{Ishigami function} \\
        \hline\hline\hline
        \multicolumn{5}{ |c| }{Expectation} \\
        \hline\hline
        \rowcolor{black!20} & Mean & Minimum & Maximum & Exact \\
        \hline\hline
            Full gPC & $3.500$ & $3.500$ & $3.500$ & $3.500$ \\
            \hline
            Sparse gPC & $3.500$ & $3.498$ & $3.501$ & $3.500$ \\
            \hline
            KRR & $3.451$ & $3.292$ & $3.760$ & $3.500$ \\
            \hline
            SSKRR & $3.501$ & $3.490$ & $3.516$ & $3.500$ \\
            \hline\hline
            \multicolumn{5}{ |c| }{Variance} \\
            \hline\hline
            Full gPC & $13.844$ & $13.844$ & $13.844$ & $13.845$\\
            \hline
            Sparse gPC & $13.819$ & $13.804$ & $13.834$ & $13.845$ \\
            \hline
            KRR & $11.211$ & $10.198$ & $12.165$ & $13.845$\\
            \hline
            SSKRR & $13.775$ & $13.590$ & $13.902$ & $13.845$ \\
            
            \hline\hline
            \multicolumn{5}{ |c| }{Kullback-Leibler divergence} \\
            \hline\hline
            Full gPC & $2.126\times10^{-6}$ & $2.037\times10^{-6}$ & $2.176\times10^{-6}$ & --\\
            \hline
            Sparse gPC & $1.512\times10^{-6}$ & $9.276\times10^{-7}$ & $2.752\times10^{-6}$ & --\\
            \hline
            KRR & $0.0186$ & $0.0134$ & $0.0314$ & --\\
            \hline
            SSKRR & $1.735\times10^{-6}$ & $1.033\times10^{-6}$ & $2.562\times10^{-6}$ & -- \\
        \hline
    \end{tabular}
    \caption{Expectation, variance, and KL divergence over the ten independent runs for the different surrogate models of the Ishigami function.}
	\label{fig:Ishigami_MeanVarKL}
\end{table}

\subsection{Rosenbrock function}\label{subsec:example_Rosenbrock}

The Rosenbrock function is an analytical function widely used in benchmarks for optimization \cite{Rosenbrock60}. It is non-convex and reads:
\begin{equation}\label{eq:Rosenbrock_func}
    \F(\Xobsb) = \sum\limits_{i=1}^{\Dim-1}\left[100\left(\Xobs_{i+1} - \Xobs_{i}^{2}\right)^2 + \left(1-\Xobs_{i}\right)^2\right]
\end{equation}
with $\Dim = 10$ and $\Xobsb \in [-2, 2]^{10}$. The input variables $\Xobsb$ are assumed to be mutually independent and uniformly distributed:
\begin{equation}\label{eq:Rosenbrock_law}
    \Xobs_i \sim \mathcal{U}(-2, 2)\,,\quad i = 1,\dots 10\,.
\end{equation}
We proceed as in \Cref{subsec:example_Ishigami}. We select polynomials of total order up to $p = 4$ to form the polynomial basis $\Basis^\basisnumb$, which corresponds to $\basisnumb={p + \Dim \choose \Dim} = {4+10 \choose 10} = 1001$ multi-dimensional Legendre polynomials. The latter are indeed orthonormal with respect to the uniform probability distribution. $\Basis^\basisnumb$ is considered for the construction of the fully tensorized gPC, the sparse gPC, and the SSKRR surrogates. For the fully tensorized gPC surrogate model, $q = 6^{10} = 60,466,176$ GL quadrature nodes are needed to exactly recover the orthonormality property given by \Cref{eq:ortho_property} since we have chosen a total order $p = 4$. The fully tensorized gPC surrogate model is not doable due to the numbers of points needed: this is the curse of dimensionality invoked in \Cref{subsec:PCE}. Therefore this model will not be considered in this example. For the sparse gPC and the SSKRR surrogates, one learning set with $\Nobs = 400$ observations of the ground truth function $\F$ is considered. The value $\Nobs = 400$ is chosen because it is significantly lower than the size of the polynomial basis $\Basis^\basisnumb$, and it yields stable solutions of \pref{eq:BPDN} as detailed in \Cref{sec:sparsity_Rosenbrock} below. Also $\epsimax = 1\times10^{-6}$ is chosen there. Finally, the KRR surrogate is built using the same learning set.

\subsubsection{Sparsity on Legendre Polynomials}\label{sec:sparsity_Rosenbrock}

\begin{figure}
	\centering
	\includegraphics[width=0.8\textwidth]{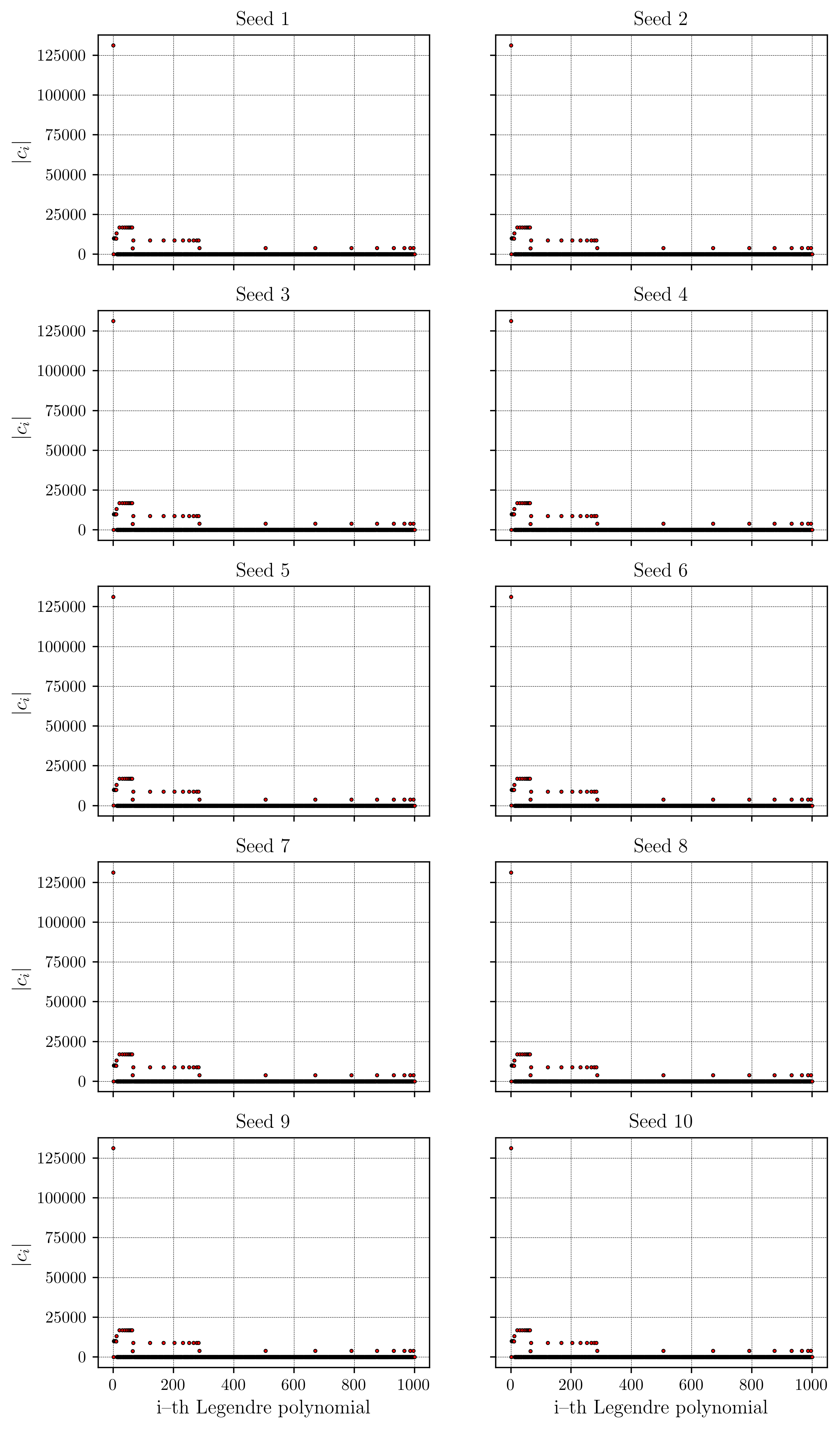}
	\caption{Evolution of the expansion coefficients $\cstarb$ with respect to the random samplings of the positions of the observations with $\Nobs = 400$ observations of $\F$.}
	\label{fig:Rosenbrock_cstar_400}
\end{figure}

To determine the sparsity $\sparsity$ needed to obtain a lower bound on the number of observations required to have a successful recovery of the expansion coefficients, we gradually increase this number and stop when the solution $\cstarb$ of \Cref{eq:BPDN} do not change significantly over ten independent runs of the positions of the observations. The foregoing study is carried out and we find that $\Nobs = 400$ observations are enough to observe sparsity over these ten runs. This justifies our choice of the size of the learning set picked above. The evolution of the expansion coefficients $\cstarb$ with respect to the random samplings of the positions of the observations are shown on \Cref{fig:Rosenbrock_cstar_400}. Here one can see that the sparsity is $\sparsity \approx 38$. Incidentally one can notice that $\sparsity = 38$ is exactly the number of terms in the expression of the Rosenbrock function.

\subsubsection{Comparison between the surrogate models}

\begin{figure}
    \centering
    \includegraphics[width=1.0\textwidth]{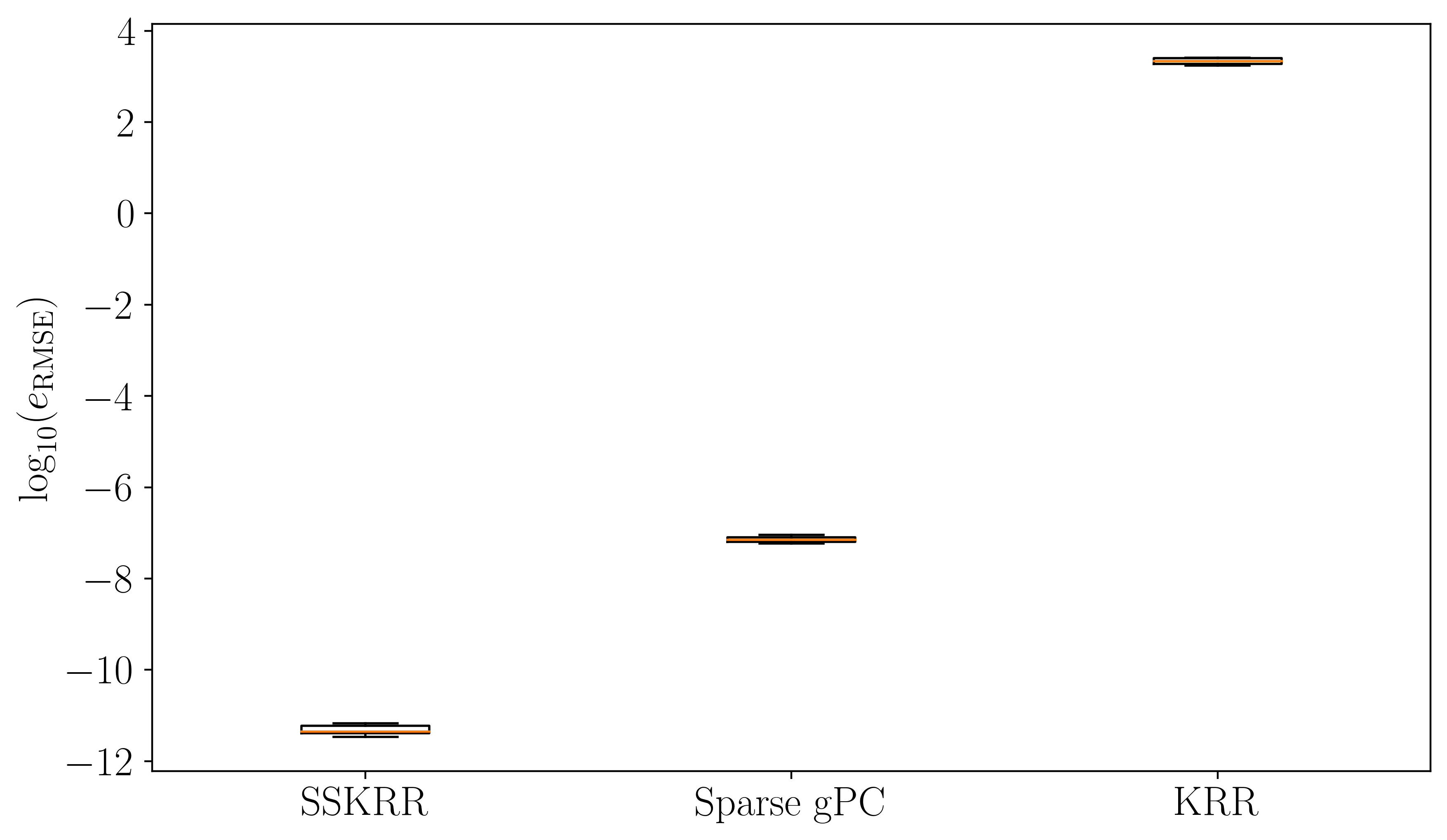} \\
    \caption{The empirical root mean square error $\RMSE$ over the ten independent runs for the Rosenbrock function.}
	\label{fig:Rosenbrock_RMSE}
\end{figure}

\begin{figure}
    \centering
    \includegraphics[width=1.0\textwidth]{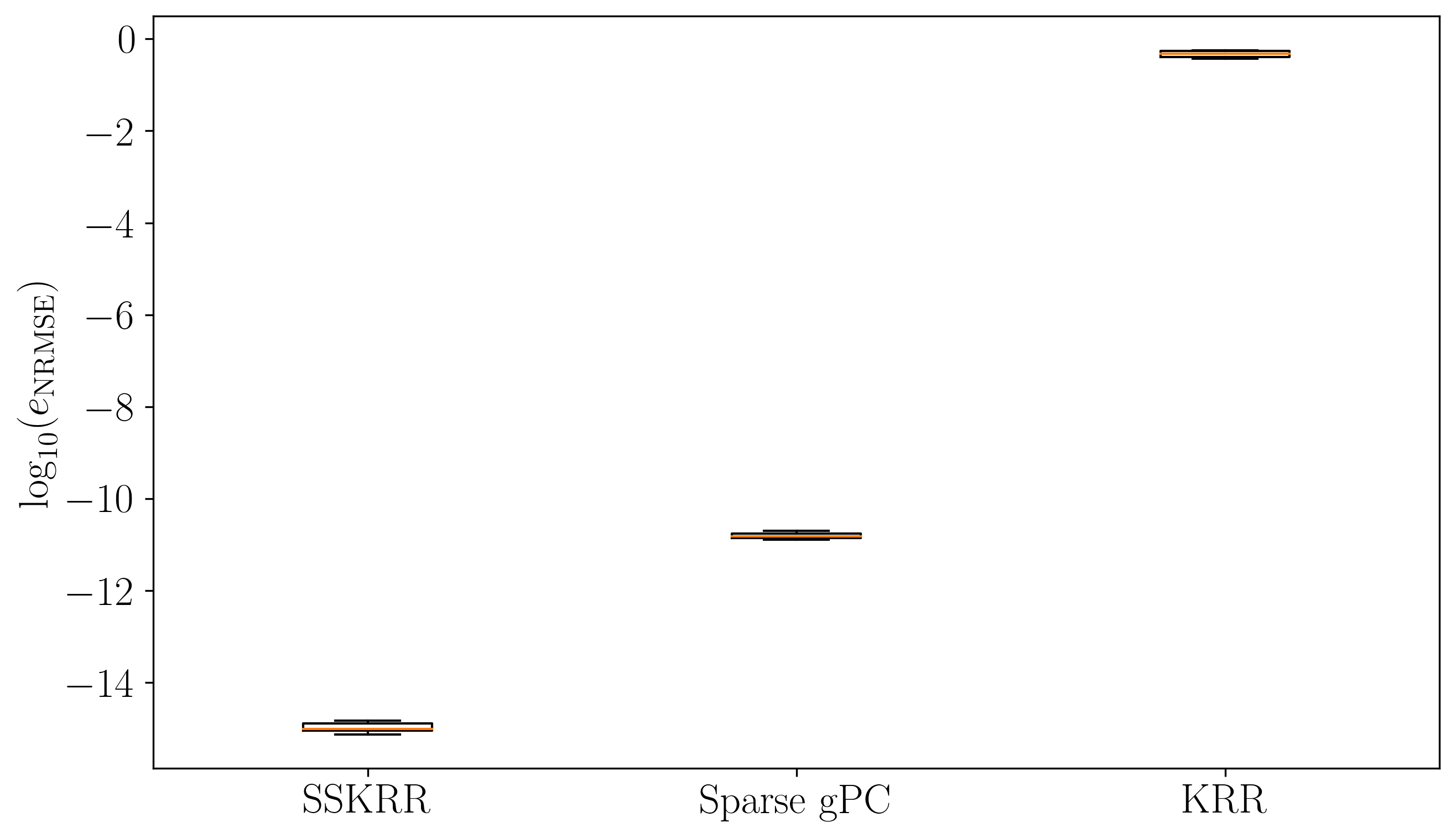}
    \caption{The empirical normalized root mean square error $\NRMSE$ over the ten independent runs for the Rosenbrock function.}
	\label{fig:Rosenbrock_NRMSE}
\end{figure}

We now compare the performance of the sparse gPC, KRR, and SSKRR surrogate models using ten independent runs with $\NT = 1\times10^{5}$ samples in the test set. Here the fully tensorized gPC surrogate model is not feasible due to the considerable number of quadrature nodes needed. A comparison of the RMSE error $\RMSE$ and the NRMSE error $\NRMSE$ over the ten independent runs is shown on \Cref{fig:Rosenbrock_RMSE} and \Cref{fig:Rosenbrock_NRMSE}. The prediction coefficient $\Qtwo$ for each surrogate model is given in \Cref{tab:Rosenbrock_Qtwo}. Notice that the prediction coefficient $\Qtwo$ is sometimes negative for the KRR surrogate model with Gaussian kernel. It means in this case that the mean of the data provides a better approximation than the KRR surrogate. One can see that the SSKRR surrogate obtained by our algorithm performs better than the sparse gPC surrogate while the KRR surrogate performs way worse than the others. One observe that only $400$ observations of $\F$ are indeed enough to obtain nearly exactly the expansion coefficients $\cstarb$ by $\ell_1$-minimization, while $Nq= 6^{10}$ observations of $\F$ would have been needed to obtain them through the fully tensorized method. Moreover, the very low values of $\RMSE$ and $\NRMSE$ of the sparse gPC and SSKRR surrogates can be explained by the fact that the Rosenbrock function is a polynomial expansion. These values are close to machine precision.

\begin{table}[h!]
    \centering
    \begin{tabular} {|c||c|c|c|}
        \hline
        \rowcolor{blue!20} \multicolumn{4}{ |c| }{Rosenbrock function} \\
        \hline\hline\hline
        \rowcolor{black!20} & Median & Minimum & Maximum \\
            \hline\hline
            Full gPC & -- & -- & -- \\
            \hline
            Sparse gPC & $\approx 1.00000$ & $\approx 1.00000$ & $\approx 1.00000$ \\
            \hline
            KRR & $-0.17561$ & $-0.61829$ & $0.27424$ \\
            \hline
            SSKRR & $\approx 1.00000$ & $\approx 1.00000$ & $\approx 1.00000$  \\
        \hline
    \end{tabular}
    \caption{Prediction coefficient $\Qtwo$ over the ten independent runs for the Rosenbrock function.}
    \label{tab:Rosenbrock_Qtwo}
\end{table}

\begin{figure}[t]
 \centering
    \includegraphics[scale = 0.8]{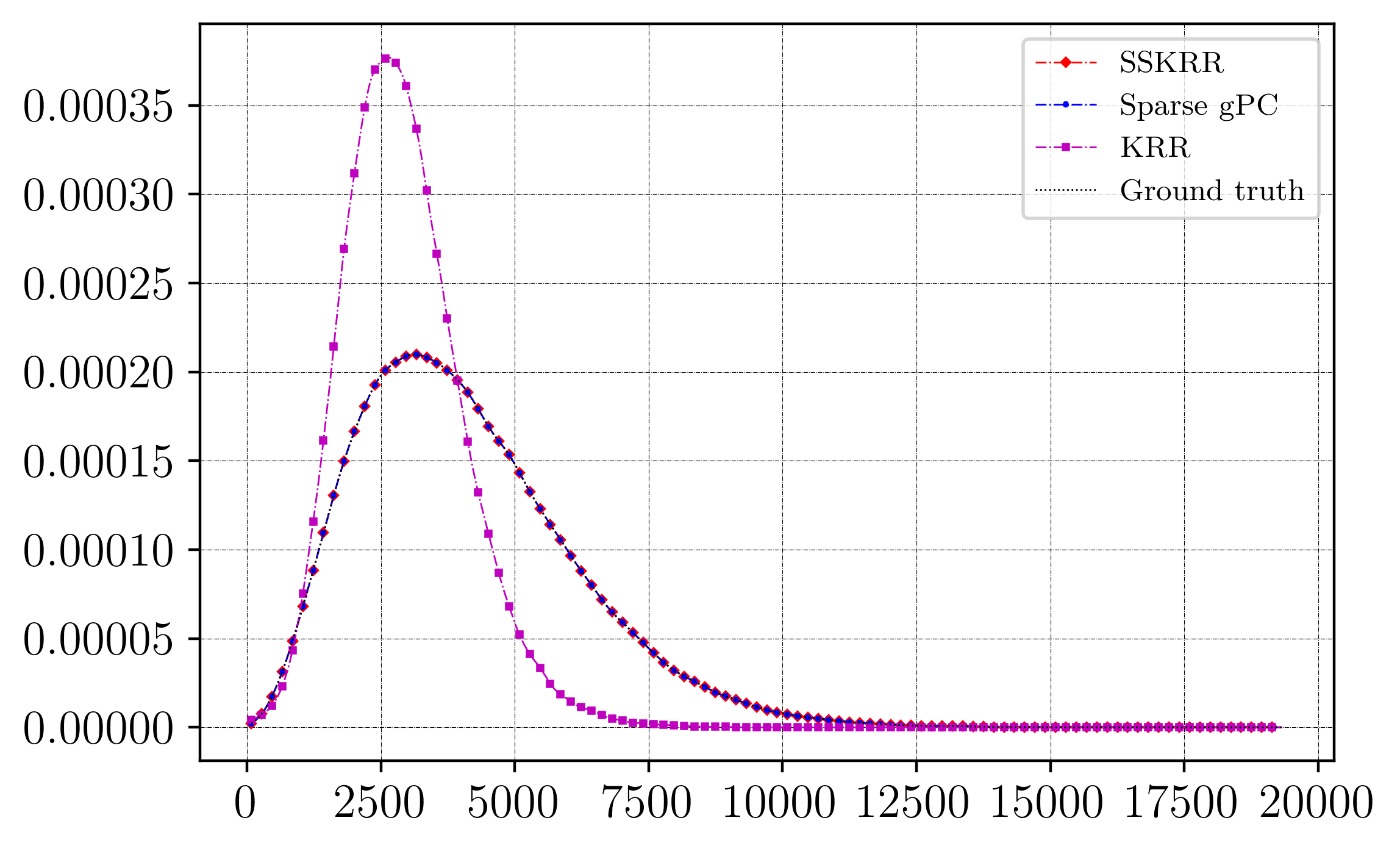}
    \caption{PDFs obtained for one run with $\Nobs = 400$ observations of $\F$. The PDFs were obtained from the $\NT= 1\times10^{5}$ data points and then smoothing out by a normal kernel density function.}
    \label{fig:Rosenbrock_PDF}
\end{figure}

We finally compare several quantities of interest of the surrogate models using the test set over the ten independent runs, namely: the expectation, the variance, and the KL divergence. The expectations and variances of the sparse gPC surrogate models are obtained using the expansion coefficients directly; see \Cref{eq:gPCmoments}. The KL divergence $\DKL$ is computed between each surrogate model and the ground truth function $\F$ by first estimating the PDF from the $\NT = 1\times10^{5}$ observations and then smoothing out the resulting histograms by a normal kernel density function \cite{Wand95}.  An example of such PDFs obtained for one run is shown on \Cref{fig:Rosenbrock_PDF}. \amend{Here the PDFs obtained from the SSKRR and sparse gPC surrogates are superimposed onto the true PDF, whereas the PDF obtained from the KRR surrogate does not fit it well}. The comparison between the surrogate models can be seen on \Cref{fig:Rosenbrock_MeanVarKL}. The exact results are computed by taking the mean of the ground truth function $\F$ on the test set for the ten independent runs. One can notice that we obtain similar expectations, variances, and KL divergences with the SSKRR surrogate compared to the sparse gPC surrogate. It has to be noted that we only needed $400$ observations to obtain the nearly exact expansion coefficients $\cstarb$. The KRR surrogate with Gaussian kernel gives the worst results as hinted by the errors $\RMSE$ and $\NRMSE$ computed previously.

\begin{table}[t]
    \small
    \centering
    \begin{tabular} {|c||c|c|c||c|}
        \hline
        \rowcolor{blue!20} \multicolumn{5}{ |c| }{Rosenbrock function} \\
        \hline\hline\hline
        \multicolumn{5}{ |c| }{Expectation} \\
        \hline\hline
        \rowcolor{black!20} & Mean & Minimum & Maximum & Exact \\
        \hline\hline
            Full gPC & -- & -- & -- & 4101.391 \\
            \hline
            Sparse gPC & $4101.000$ & $4101.000$ & $4101.000$ & 4101.391 \\
            \hline
            KRR & $3162.772$ & $2847.550$ & $3518.393$ & 4101.391\\
            \hline
            SSKRR & $4101.391$ & $4089.939$ & $4122.100$ & 4101.391 \\
            \hline\hline
            \multicolumn{5}{ |c| }{Variance} \\
            \hline\hline
            Full gPC & -- & -- & -- & 4092724.179 \\
            \hline
            Sparse gPC & $4105081.752$ & $4105081.752$ & $4105081.752$ &  4092724.179 \\
            \hline
            KRR & $1484933.564$ & $1128976.833$ & $1811692.945$ & 4092724.179 \\
            \hline
            SSKRR & $4092724.179$ & $4046013.350$ & $4133221.629$ & 4092724.179 \\
            \hline\hline
            \multicolumn{5}{ |c| }{Kullback-Leibler divergence} \\
            \hline\hline
            Full gPC & -- & -- & -- & --\\
            \hline
            Sparse gPC & $\approx0$ & $\approx0$ & $\approx0$ & --\\
            \hline
            KRR & $0.198$ & $0.142$ & $0.323$ & --\\
            \hline
            SSKRR & $\approx0$ & $\approx0$ & $\approx0$ & -- \\
        \hline
    \end{tabular}
    \caption{Expectation, variance, and KL divergence over the ten independent runs for the different surrogate models of the Rosenbrock function.}
	\label{fig:Rosenbrock_MeanVarKL}
\end{table}

\section{Application to the RAE2822 transonic airfoil}\label{subsec:example_UMRIDA_3D}
\newcommand{\lift}{C_L}

\begin{figure}[h!]
  \centering
  \includegraphics[width = 0.7\linewidth]{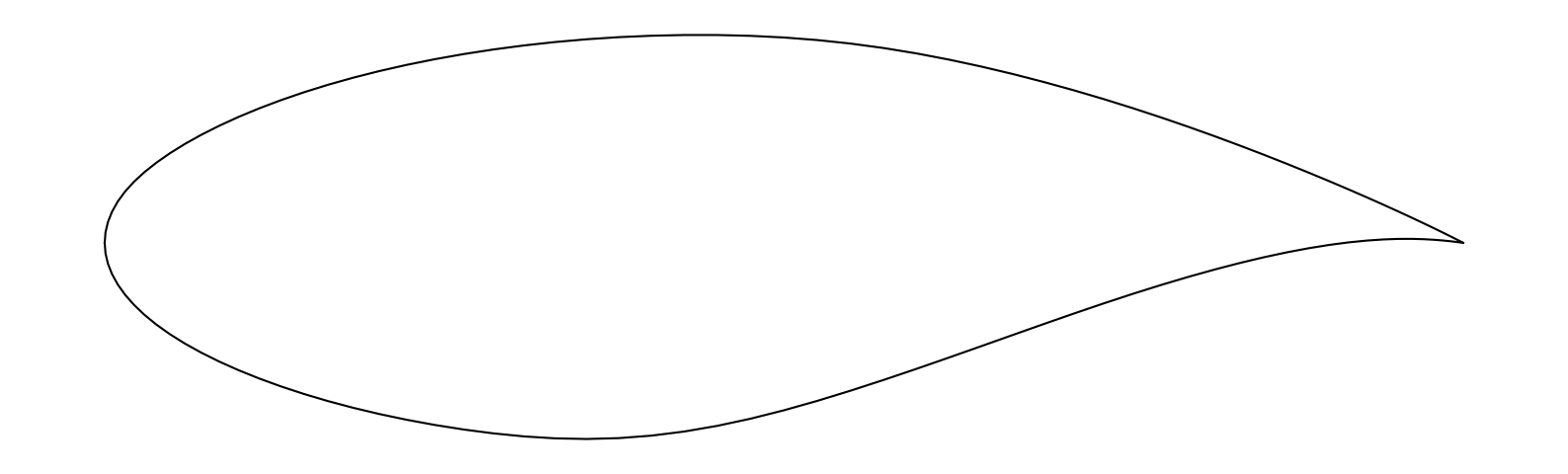}
  \caption{Geometry of the RAE2822 airfoil.}\label{fig:rae288_airfoil_picture}
\end{figure}

We now apply the methods of \Cref{sec:numerical_ex} to a complex aerodynamic test case: the two-dimensional RAE2822 airfoil of which geometry is depicted on \Cref{fig:rae288_airfoil_picture}. The RAE2822 wing profile is a supercritical airfoil which has become a standard test case for turbulence modeling validation in transonic regimes \cite{Cook1979}.
%The quantities of interest are the lift coefficient $C_L$, the drag coefficient $C_D$, and the pitching moment coefficient $C_M$ as functions of
Here we aim to build a surrogate model of the lift coefficient $\lift$---this is the ground truth function $\F$ of \Cref{pb:01}---of that airfoil when some characteristics of the flow and/or the profile are variable and only a finite number of observations of $\lift$ is available. More precisely, three (random) input variables are considered: the free-stream Mach number $M$, the angle of attack $\alpha$, and the thickness-to-chord ratio $r$ of the airfoil. These three parameters define the input vector $\Xobsb = (r,M,\alpha)\in \InputSpace = \InputSpace_1\times\InputSpace_2\times\InputSpace_3$. We compare the performances of four surrogate modeling methods: KF as sketched in \Cref{sec:chap3_KF}, and fully tensorized gPC, sparse gPC \cite{Savin16}, and SKRR as sketched in \Cref{sec:SKKR_algo}. We note that ordinary Kriging has already been applied to the RAE2822 airfoil in a different input space in \cite{Dumont2019}, gradient-enhanced Kriging has been applied to this very profile in \cite{Laurenceau08}, and universal Kriging with a PC expansion of the trend has been applied to the NACA4412 airfoil (another classical example of turbulence modeling validation) in \cite{Wein2019}.

\subsection{Problem setup}

Observations of the ground truth function $\lift$ are obtained by solving the steady-state Reynolds-Averaged Navier-Stokes (RANS) equations together with a Spalart-Allmaras turbulence model closure \cite{Spalart92}. The CFD solver \textit{elsA} \cite{elsa} is used to simulate two-dimensional transonic flows around that airfoil and construct the learning set. The nominal flow conditions correspond to the ones described in \cite[Test case \#6]{Cook1979} together with the correction formulas for the wall interference derived in \cite[pp. 386--387]{Garner1966}, and their slight modifications proposed in \cite{Haase1993}. The operational parameters considered here are thus $ \underline{M} = 0.729$ for the free-stream Mach number, $ \underline{\alpha} = 2.31\degree$ for the angle of attack, and $\underline{\mathrm{Re}} = 6.50\times10^{6}$ for the Reynolds number based on the chord length $c$, fluid velocity, temperature, and molecular viscosity at infinity. They arise from the corrections $\Delta M = 0.004$ and $\Delta \alpha = -0.61\degree$ given in \cite[pp. 130]{Haase1993} for the test case $\#6$ outlined in \cite{Cook1979}, for which $M = 0.725$, $\alpha = 2.92\degree$, and $\mathrm{Re} = 6.50\times10^{6}$. More details about this example and the numerical parameters used for \textit{elsA} runs can be found in \cite{Savin16}.
%Denoting by $(\Xobs_1, \Xobs_2, \Xobs_3) = (r, M, \alpha)$ the random input variables, they are mutually independent and follow Beta laws of the first kind $\beta_{I}(a, b)$\string:
The random input variables are mutually independent and follow Beta distributions of the first kind $\beta_{I}(a, b)$:
\begin{equation*}
    \beta_{I}(x, a, b) = \indic_{[X_\text{l}, X_\text{u}]}(x) \frac{\Gamma(a+b)}{\Gamma(a)\Gamma(b)} \frac{(x-X_\text{l})^{a-1}(X_\text{u}-x)^{b-1}}{(X_\text{u} - X_\text{l})^{a+b-1}}\,,
\end{equation*}
where $a = (4, 4, 4)$, $b = (4, 4, 4)$, and $[X_\text{l}, X_\text{u}]$ is the compact support of the random parameter $X\sim\beta_{I}$. \Cref{tab:input_param} gathers the ranges $\InputSpace_i$, $i=1,2,3$, of each random input variable and their associated parameters $(a,b)$.

\begin{table}[h]
    \begin{center}
        \begin{tabular} {|c||c|c|c|}
            \hline
            \rowcolor{black!20} & $X_{\mathrm{lb}}$ & $X_{\mathrm{ub}}$ & (a,b)\\
            \hline\hline
                $\Xobs_1 = r$ & $0.97\times\underline{r}$ & $1.03\times\underline{r}$ & $(4,4)$ \\
                \hline
                $\Xobs_2 = M$ & $0.95\times\underline{M} $ & $1.05\times\underline{M}$ & $(4,4)$ \\
                \hline
                $\Xobs_3 = \alpha$ & $0.98\times\underline{\alpha}$ & $1.02\times\underline{\alpha}$ & $(4,4)$ \\
            \hline
        \end{tabular}
    \end{center}
    \caption{Range and probability distribution of each input parameter, with $\underline{r} = 1$.}
    \label{tab:input_param}
\end{table}

For this example, surrogates $\FapproxN$ are built for the ground truth function $\lift$ using four different methods: (i) a KRR surrogate model \pref{eq:solution_KRR} using the parametric KF algorithm of \Cref{subsec:chap3_P_KF_algo} to estimate both the nugget and the length scales of a Gaussian base kernel \pref{eq:Gaussian_kernel_ARD}; (ii) a fully tensorized gPC surrogate model \pref{eq:G_PCE} where the expansion coefficients are obtained by tensorized GL quadrature nodes in \pref{eq:quad_nodes}; (iii) a sparse gPC surrogate model \pref{eq:G_PCE} where the expansion coefficients are obtained by solving the problem \pref{eq:BPDN}; and (iv) a SSKRR surrogate model obtained by \Cref{algo:SKRR_sparse} where the nugget is tuned by the parametric KF algorithm. 

\begin{figure}[h]
	\centering
	\includegraphics[width=1.0\textwidth]{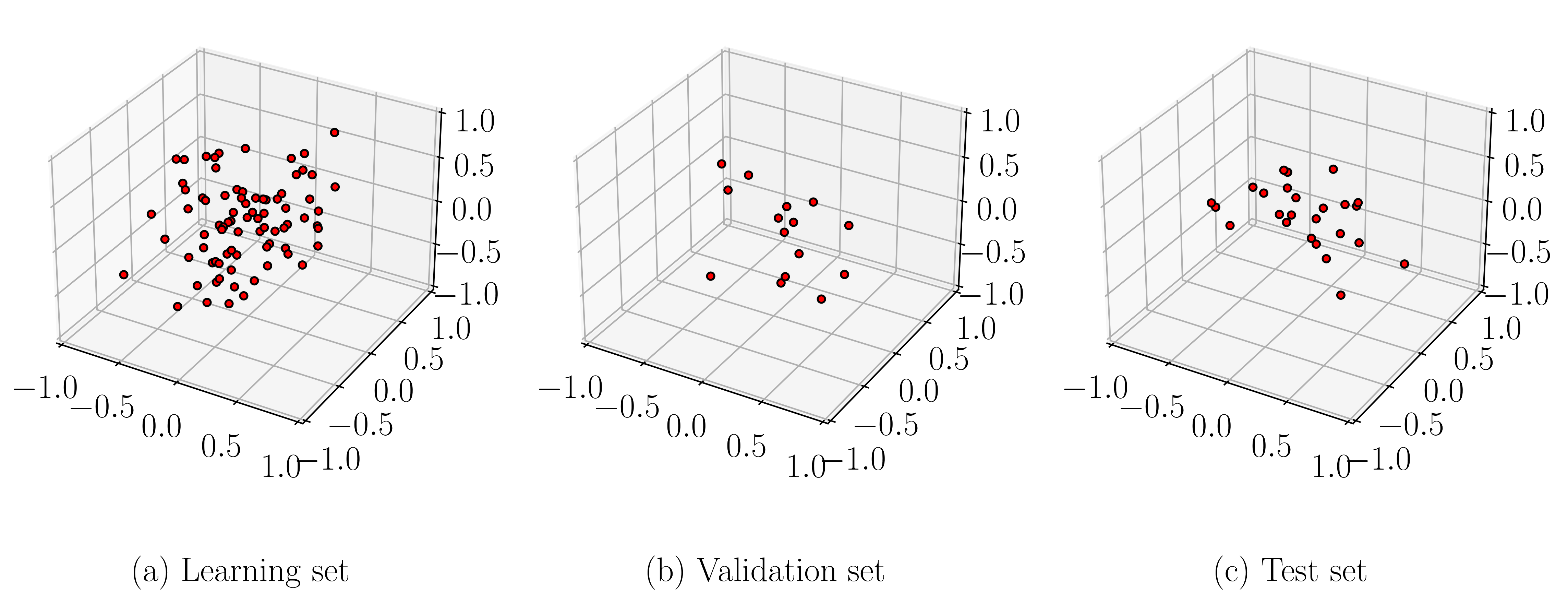}
	\caption{Random sampling points in (a) the learning set, (b) the validation set, and (c) the test set with $\Nobs = 80$, $\NV = 15$, and $\NT = 25$ points used to derive the sparse gPC, KRR, and SSKRR surrogates.}
	\label{fig:UMRIDA_120_random}
\end{figure}

The $\Nobs$ observations of $\lift$ used to construct the sparse gPC, KRR, and SSKRR surrogates are obtained by random trials of the random input variables following Beta distributions. For the fully tensorized gPC surrogate, the number of observations is chosen in order to exactly integrate the orthonormality property given by \Cref{eq:ortho_property} for polynomials of total order up to $p$. We recall that given $q$ nodes, the GL quadrature rule exactly integrates uni-variate polynomials of order $2q-3$. The four surrogates are subsequently validated on a validation set consisting of $\NV$ observations, and tested on a test set consisting of $\NT$ observations where the input variables are again drawn randomly following Beta distributions with parameters as in \Cref{tab:input_param}. These $\NTOT = \Nobs + \NV + \NT$ observations are thus split as follows\string:
\begin{itemize}
    \item The learning set which consists of $\Nobs=80$ observations, that is $67\%$ of the $\NTOT$ observations: $( \Xobsb, \lift(\Xobsb))$; %$\left( \Xobsb, C_L(\Xobsb), C_D(\Xobsb), C_M(\Xobsb)  \right)$;
    \item The validation set which consists of $\NV=15$ observations, that is $12\%$ of the $\NTOT$ observations: $(\Xobsb_{\NV},\lift(\Xobsb_{\NV}))$; %$\left( \Xobsb_{\NV}, C_L(\Xobsb_{\NV}), C_D(\Xobsb_{\NV}), C_M(\Xobsb_{\NV}) \right)$;
    \item The test set which consists of $\NT=25$ observations, that is $21\%$ of the $\NTOT$ observations: $( \Xobsb_{\NT},\lift(\Xobsb_{\NT}))$. %$\left( \Xobsb_{\NT}, C_L(\Xobsb_{\NT}), C_D(\Xobsb_{\NT}), C_M(\Xobsb_{\NT}) \right)$.
\end{itemize}
These different sets are shown on \Cref{fig:UMRIDA_120_random}. Other splitting choices could have been made, for instance the classical $60/20/20$ splitting ($60\%$ for the learning set, $20\%$ for the validation set, $20\%$ for the test set) as in \cite{Hastie2009}.

The performance of each surrogate model is quantified by computing the empirical Normalized Root Mean Square Error $\NRMSE$ of \Cref{eq:error_NRMSE} and the empirical Root Mean Square Error $\RMSE$ of \Cref{eq:error_RMSE} using the validation and test sets. However the knowledge of $\RMSE$ and $\NRMSE$ might not be enough to assess the performance of a surrogate model. Indeed, $\RMSE$ only gives the global error over the whole domain but does not give any information about the distribution. For instance, two similar values of $\RMSE$ for two different surrogate models can be obtained\string: in one case the surrogate provides a reliable approximation of the ground truth function for the majority of the domain but a poor one for a few points, while in another case the other surrogate provides a less reliable approximation of the ground truth function in the entire domain. In that respect, we compute an additional metric, the maximum relative error $\ERRrel$ defined by:
\begin{equation}
    \ERRrel = \max\limits_{i=1,\dots\NT} \left(\frac{\absolute{\Yobs_i - \FapproxN(\Xobsb_i)}}{\absolute{\Yobs_i}}\right)\times100\,.
\end{equation}

Following \cite{Savin16}, we choose a total order up to $p = 8$ which corresponds to $\basisnumb={p + \Dim \choose \Dim} = {8+3 \choose 3} = 165$ multi-dimensional Jacobi polynomials. The latter are indeed orthonormal with respect to the Beta distribution. They constitute the basis $\Basis^\basisnumb$ considered for the construction of the fully tensorized gPC, the sparse gPC, and the SSKRR surrogates. Since $p = 8$, $N = 1000$ GL quadrature nodes are needed to exactly recover the orthonormality property given by \Cref{eq:ortho_property} and are selected to compute the expansion coefficients of the fully tensorized gPC surrogate by \Cref{eq:quad_nodes}. SPGL1 in $\python$\; \cite{Berg08, Berg11} is again considered in order to compute the solution of \pref{eq:BPDN} for the expansion coefficients in the sparse gPC surrogate \pref{eq:G_PCE} and the SSKRR surrogate \pref{eq:SSKRR_approx} obtained by \Cref{algo:SKRR_sparse}. Also $\epsimax = 1\times10^{-5}$ has been chosen in \pref{eq:BPDN}. %The coherence parameter of \Cref{def:coh_param} is $\coherence(\bm{\Theta})\approx 270$ on the learning set which yields a large theoretical lower bound in \Cref{th:noiseless_sampling}.
In \cite{Savin16} it has been observed that $\sparsity\approx 10$ and that $\Nobs = 80$ observations of the learning set yielded satisfactory results below.

The KRR surrogate model is built by the parametric KF algorithm of \Cref{subsec:chap3_P_KF_algo} using the framework \GPytorch\ \cite{Gpytorch}. The optimization of the parameters is done by the optimizer \texttt{Adam} \cite{Kingma2017adam} implemented in \Pytorch\ \cite{Pytorch2019}, which is run to compute the gradients by automatic differentiation \cite{Paszke17}. We initialize the length scales $\gamma_i$, $i=1,2,3$, as:
\begin{equation*}
\gamma_i = \frac{2}{\Nobs(\Nobs-1)}\sum\limits_{j=1}^{\Nobs}\sum\limits_{k=j+1}^{\Nobs} \norm{\Xobsb_j- \Xobsb_k}{2}\,,
\end{equation*}
and the nugget as $\lambda = 1\times10^{-6}$. Also the nugget of the SSKRR surrogate \pref{eq:SSKRR_approx} is tuned by the parametric KF algorithm as well, starting from the same initial guess $\lambda = 1\times10^{-6}$. In addition, the trace parameter $\kappa$ is chosen as $\kappa = \var(\Yobsb)$. The selected values of the parameters from the parametric KF algorithm with $\NF = \Nobs$, $\NC = \NF/2$, and the accuracy $\rho$ defined by \Cref{eq:rho_p}, are chosen as their values at the iteration for which the $\RMSE$ on the $\NV$ observations of the validation set is minimal; see \Cref{fig:UMRIDA_120_CL_KRR-KF} below for the parametric KF algorithm applied to the KRR surrogate, and \Cref{fig:UMRIDA_120_CL_KF} for the parametric KF algorithm applied to the SSKRR surrogate. This choice aims to evade possible overfitting.

A pick-freeze estimator \cite{Janon14, Prieur16} is subsequently used to compute Sobol' main-effect sensitivity indices obtained from the KRR and SSKRR surrogates. A matrix size of $1\times10^{6}$ samples is selected, corresponding to a total number of $(\Dim + 1)\times10^{6} = 4\times10^{6}$ evaluations of the surrogate model. This method is a Monte-Carlo based one and thus it may be difficult to obtain accurate estimates of small sensitivity indices. On the other hand, these indices are directly obtained from the expansion coefficients of the fully tensorized and sparse gPC surrogates \cite{Sudret08}. %; see \Cref{sec:App_UQ}.

%%%%%%%%%%%%%%

\subsection{Results}

\begin{figure}[h!]
    \centering
    \includegraphics[width=0.75\textwidth]{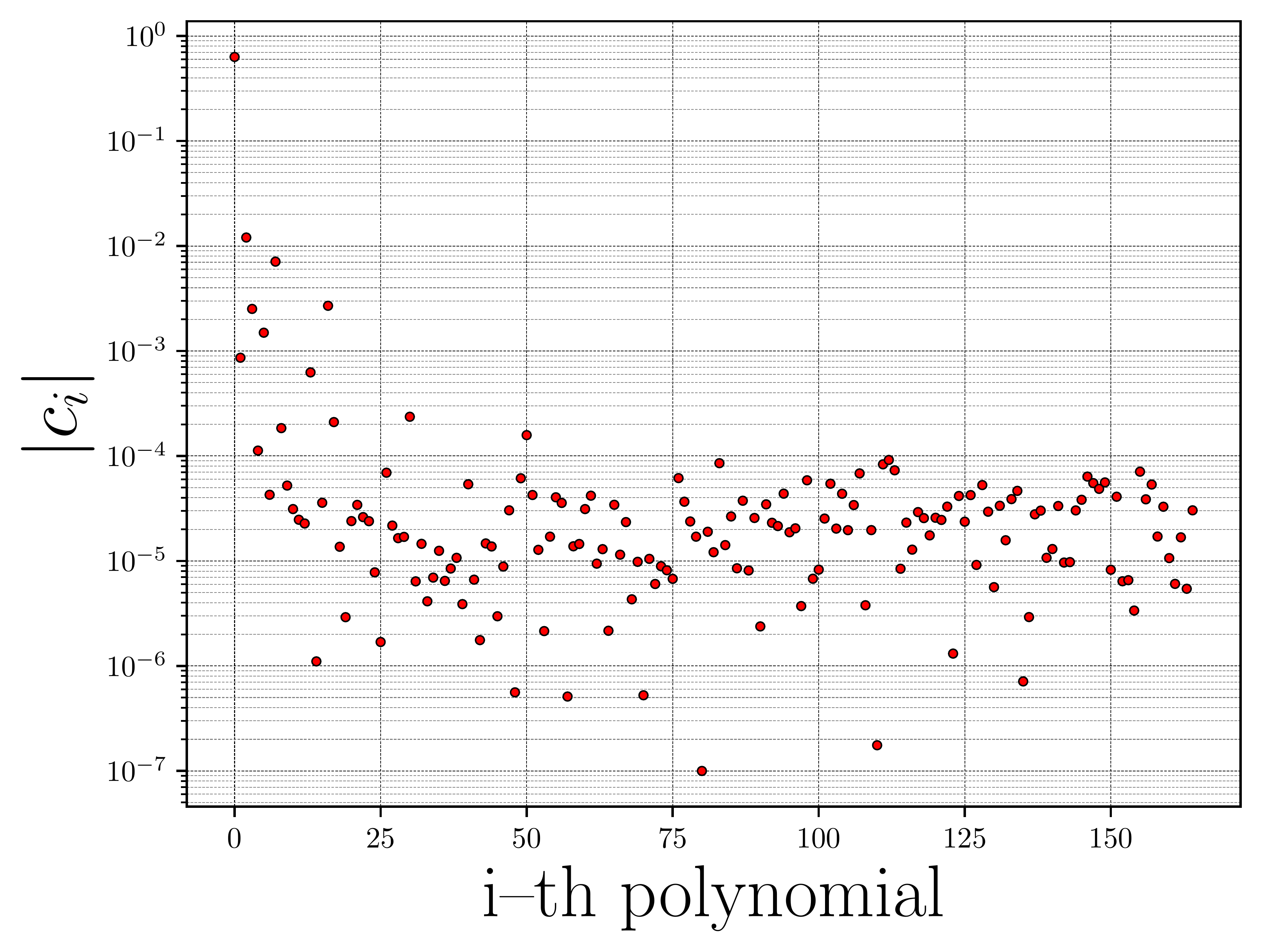}
    \caption{Expansion coefficients $\cstarb$ with $\Nobs = 80$ observations of $\lift$.}
    \label{fig:UMRIDA_120_CL_ci}
\end{figure}

The polynomial expansion coefficients $\cstarb$ of $\lift$ in $\Basis^\basisnumb$ given by $\ell_1$-minimization \pref{eq:BPDN} are shown on \Cref{fig:UMRIDA_120_CL_ci} using the learning set of \Cref{fig:UMRIDA_120_random}. We note that only low order polynomials are relevant. Indeed, one can see that:
\begin{multline}\label{eq:UMRIDA_120_CL_coeff_expan}
    \FapproxPCE(\x) \approx {c}_{1}^{\star}\phi_{(0,0,0)}(\x) + {c}_{2}^{\star}\phi_{(1,0,0)}(\x) + {c}_3^{\star}\phi_{(0,1,0)}(\x) + {c}_4^{\star}\phi_{(0,0,1)}(\x) \\ + {c}_{6}^{\star}\phi_{(1,1,0)}(\x)+ {c}_{8}^{\star}\phi_{(0,2,0)}(\x) + {c}_{17}^{\star}\phi_{(0,3,0)}(\x)\,,
\end{multline}
where $\phi_{(i_1,i_2,i_3)}$ is defined as in \Cref{eq:phi_multi}, and $|{c}_{1}^{\star}| \gg |{c}_{3}^{\star}| \gg |{c}_4^{\star}| \approx |{c}_{6}^{\star}| \approx |{c}_{8}^{\star}| \approx |{c}_{17}^{\star}| \gg |{c}_{2}^{\star}|$. The highest order polynomial of $\FapproxPCE(\x)$ has order $3$ and its expansion coefficient is small compared to the others. 
From \Cref{fig:UMRIDA_120_CL_ci}, the sparsity is observed to be $\sparsity \approx 7$ for a threshold $\delta$ of about $10^{-3}$. %Therefore \Cref{th:noiseless_sampling} yields $N \gtrsim \Nobss \approx 10000$ observations up to a constant $C$, but 
Also we observe in practice that $\Nobs = 80$ observations are enough for an accurate recovery in \pref{eq:BPDN}.

\begin{figure}[h!]
    \centering
    \subfigure[Accuracy $\rho$.]{\includegraphics[width=0.48\textwidth]{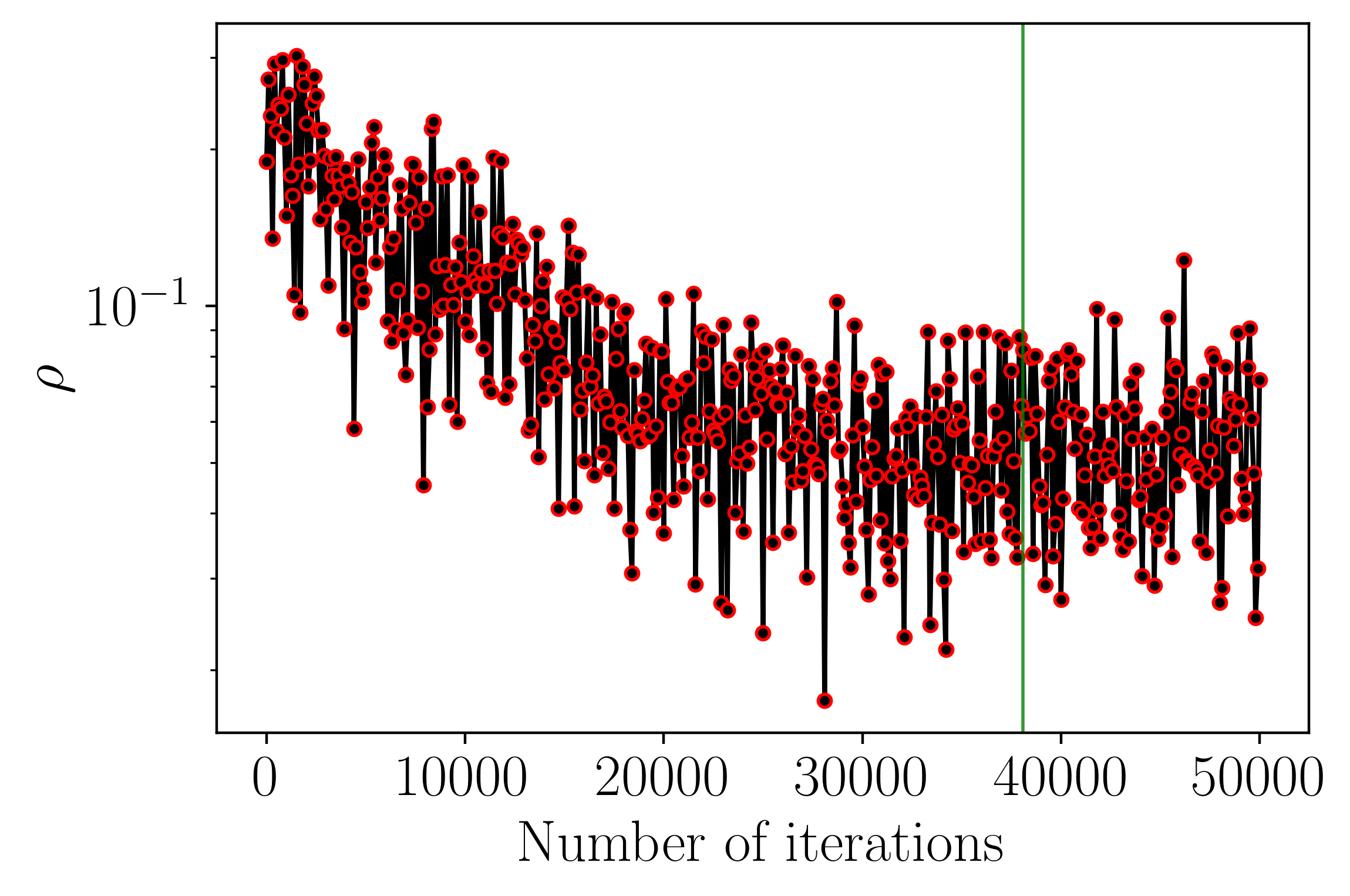}}
    \subfigure[Nugget $\lambda$.]{\includegraphics[width=0.48\textwidth]{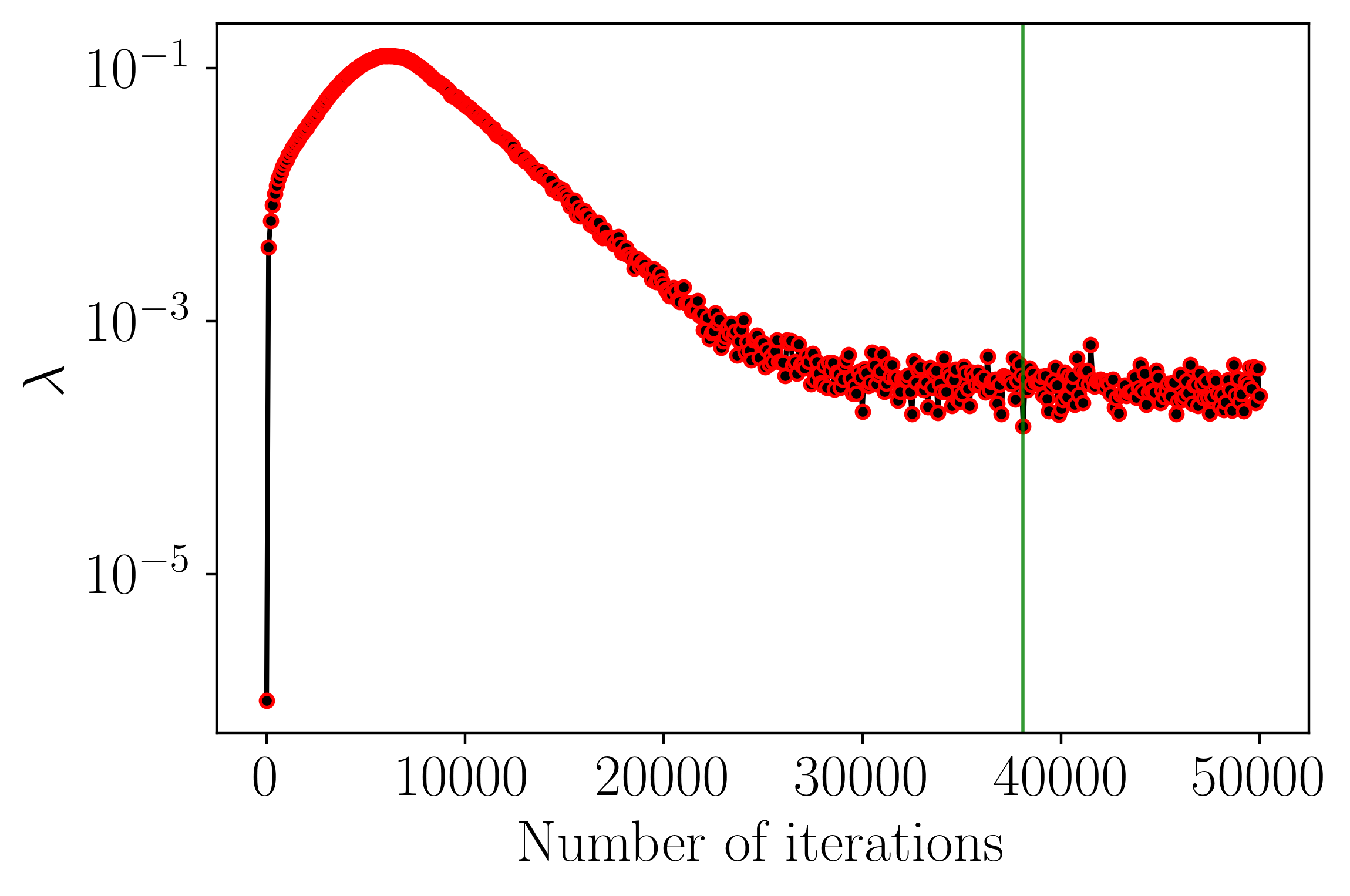}} \\
    \subfigure[Length scale $\gamma_1$.]{\includegraphics[width=0.48\textwidth]{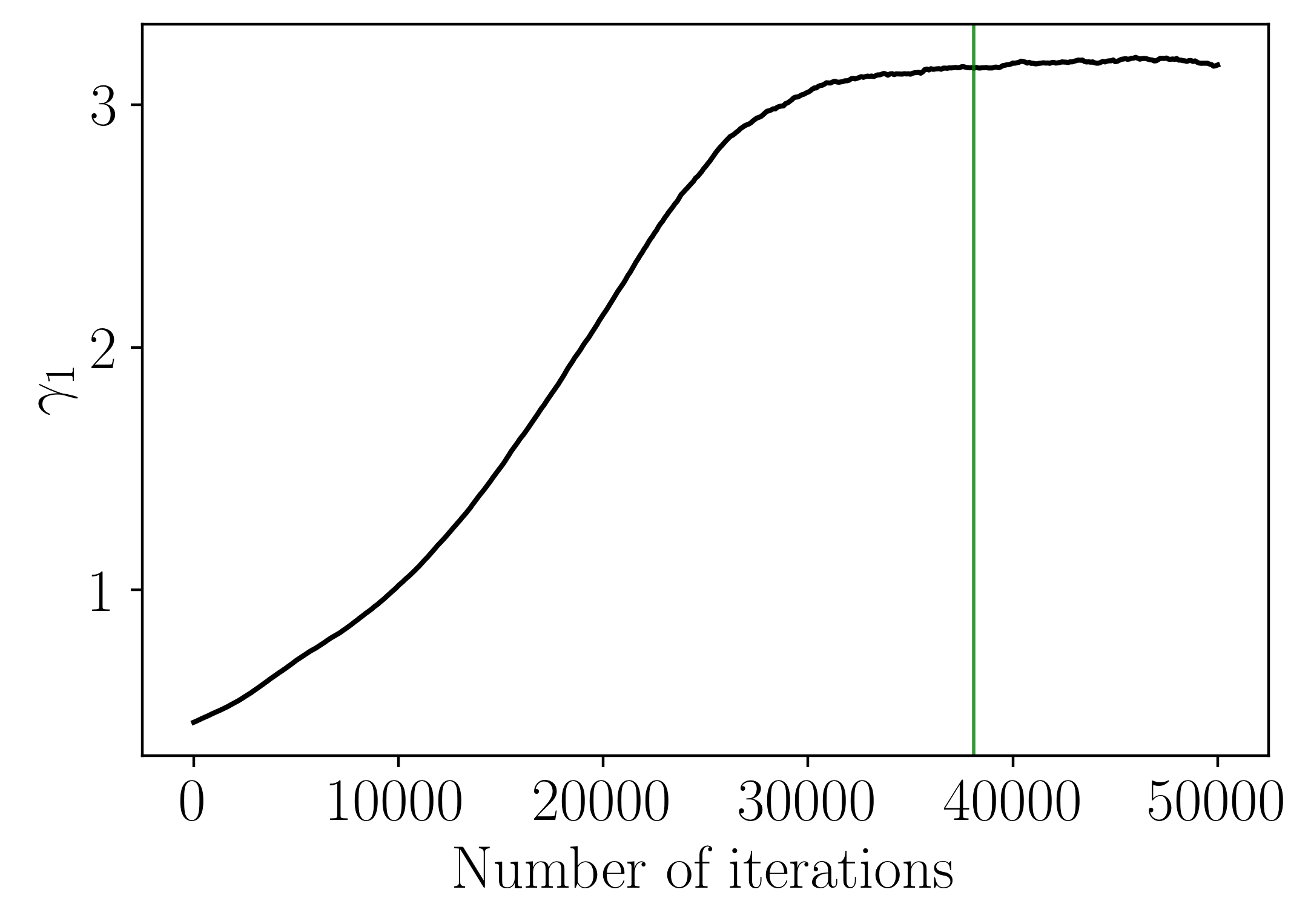}}
    \subfigure[Length scale $\gamma_2$.]{\includegraphics[width=0.48\textwidth]{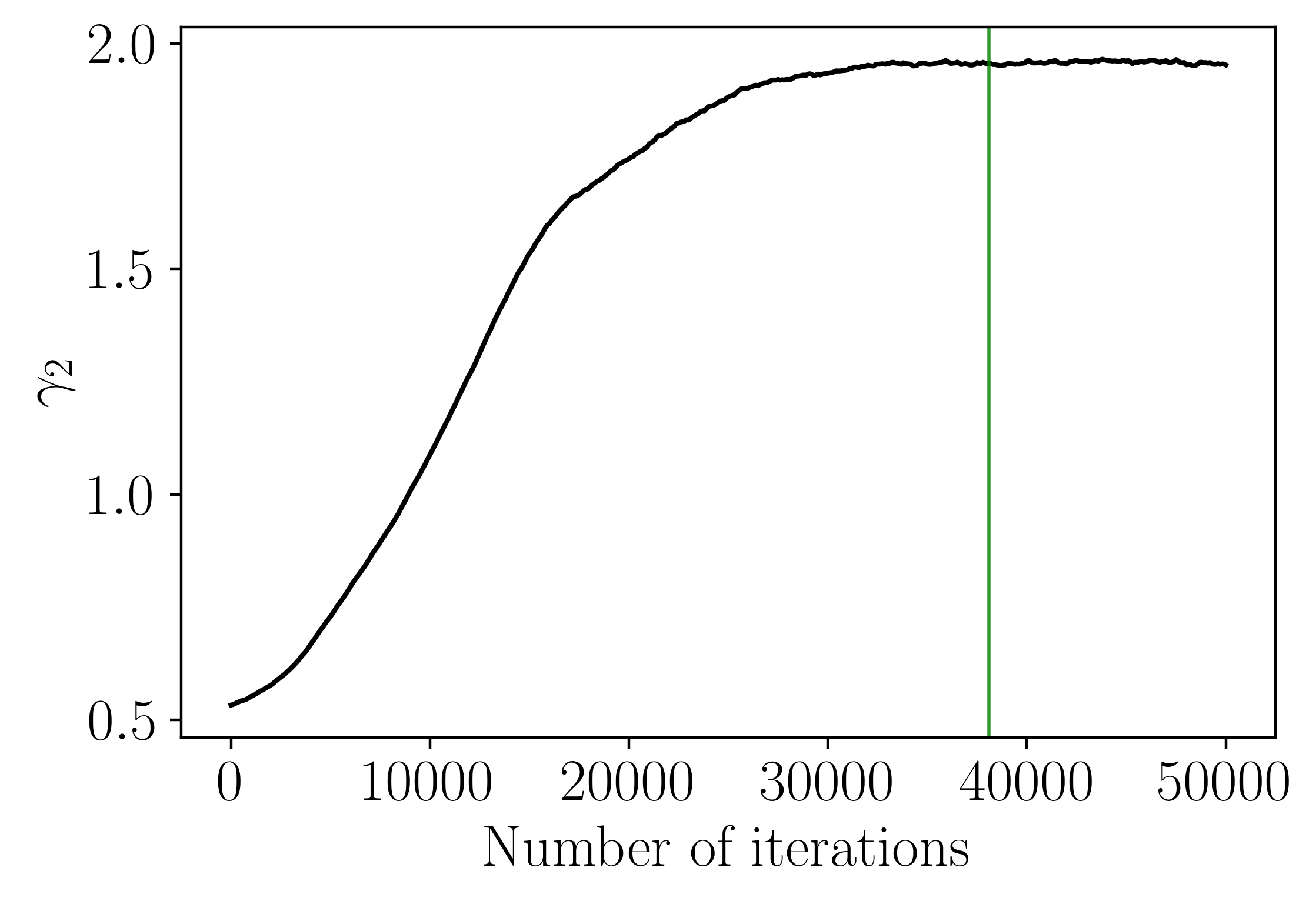}} \\
    \subfigure[Length scale $\gamma_3$.]{\includegraphics[width=0.48\textwidth]{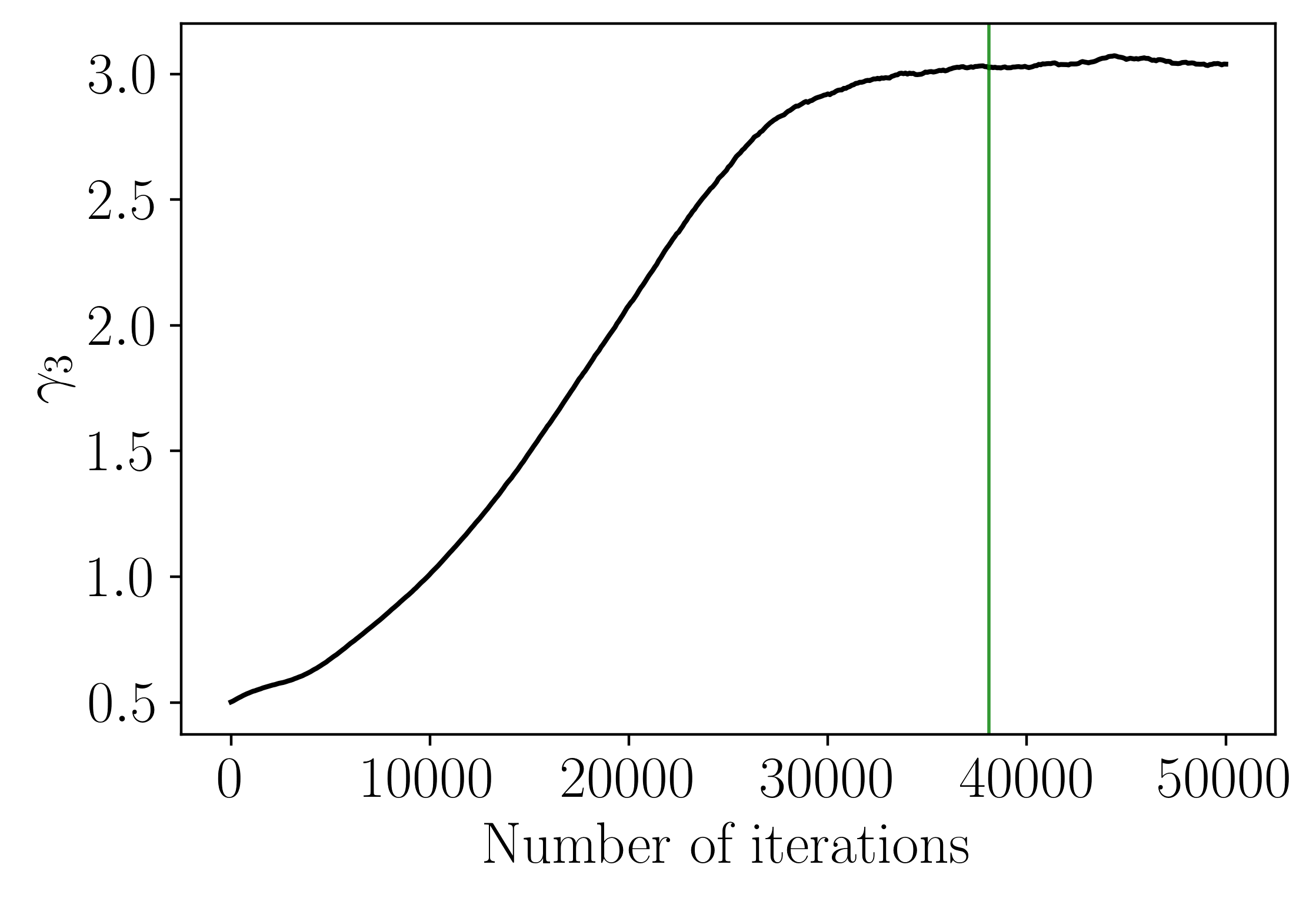}}
    \subfigure[Error $\RMSE$ on the validation set.]{\includegraphics[width=0.48\textwidth]{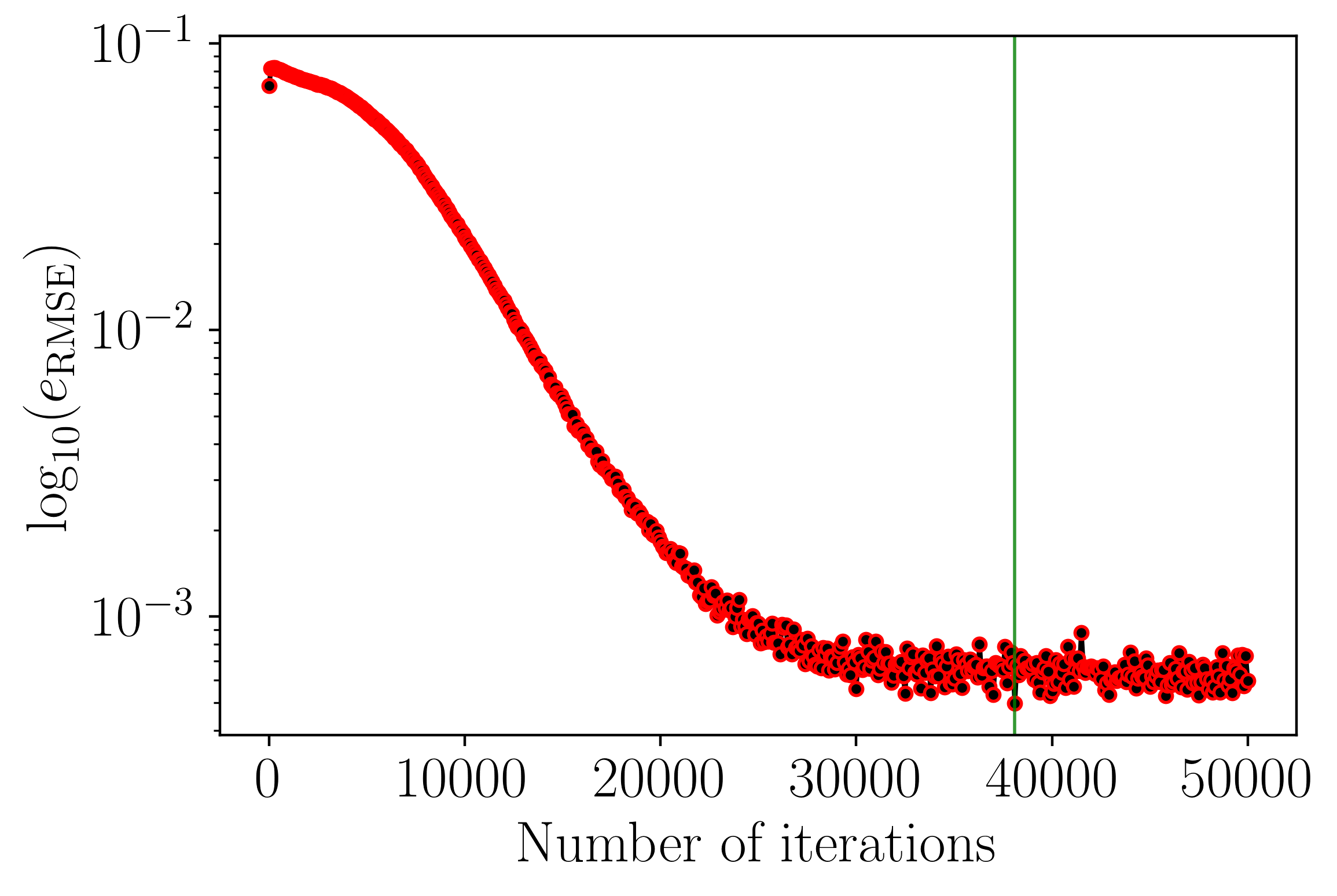}}
    \caption{Evolutions of the accuracy $\rho$, the nugget $\lambda$, the lengthscales $\gamma_i$, $i=1,2,3$, and the error $\RMSE$ on the validation set as functions of the number of iterations for the parametric KF algorithm applied to the KRR surrogate of $\lift$. The green vertical lines correspond to the iteration where the error $\RMSE$ is minimal on the validation set.}\label{fig:UMRIDA_120_CL_KRR-KF}
\end{figure}

\begin{figure}[h!]
    \centering
    \subfigure[Accuracy $\rho$.]{\includegraphics[width=0.48\textwidth]{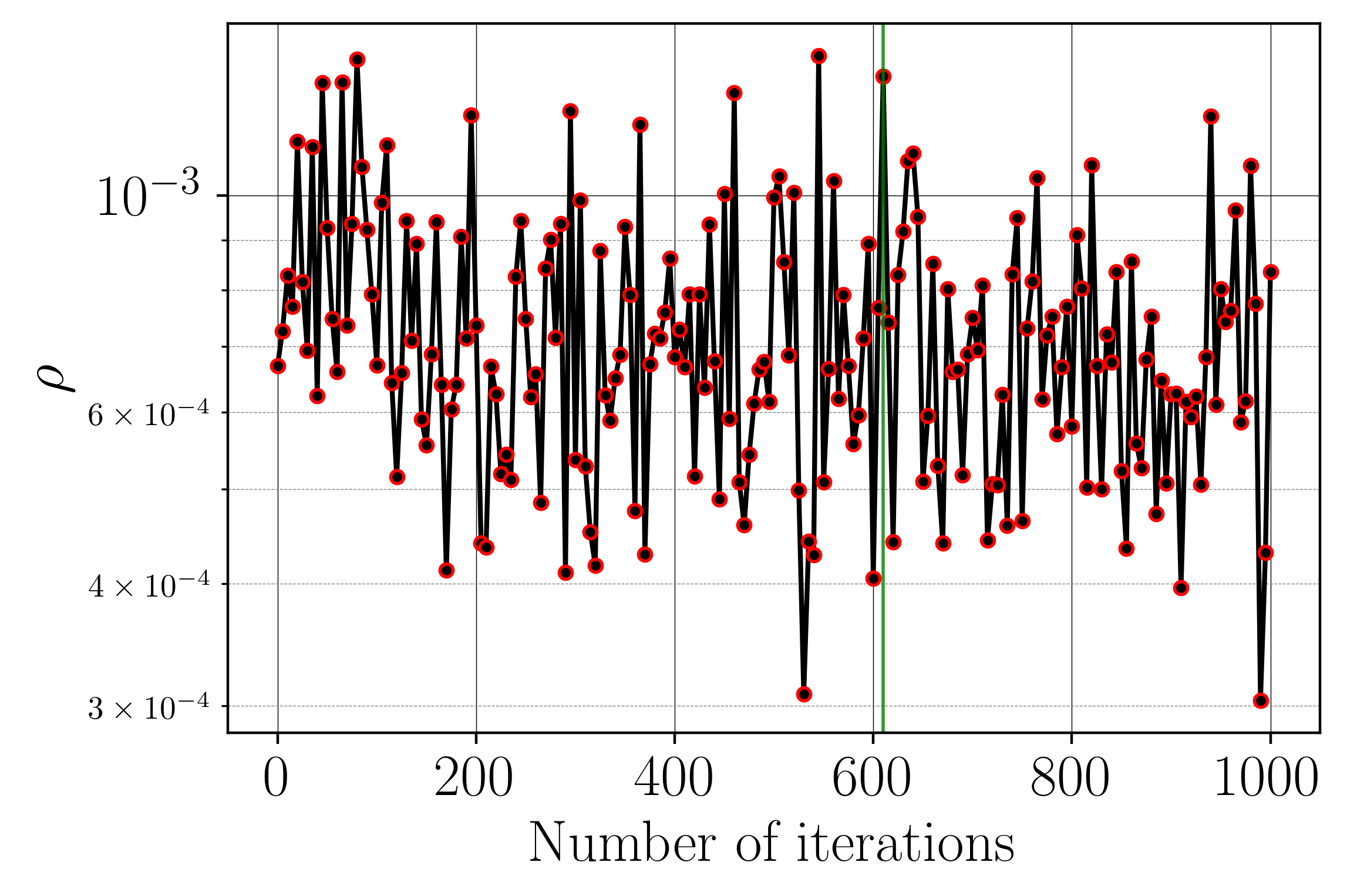}}
    \subfigure[Nugget $\lambda$.]{\includegraphics[width=0.48\textwidth]{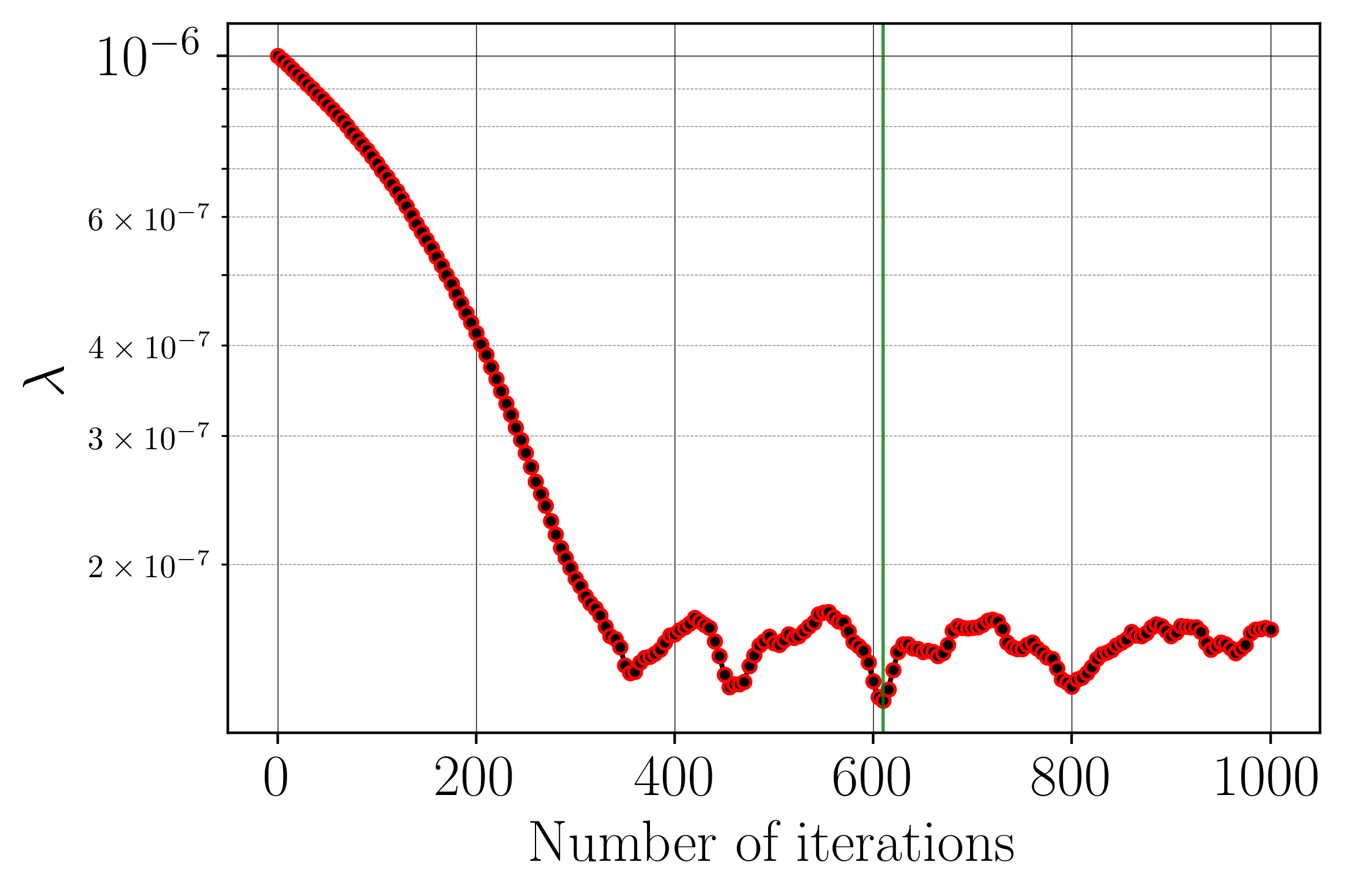}} \\
    \subfigure[Error $\RMSE$ on the validation set.]{\includegraphics[width=0.48\textwidth]{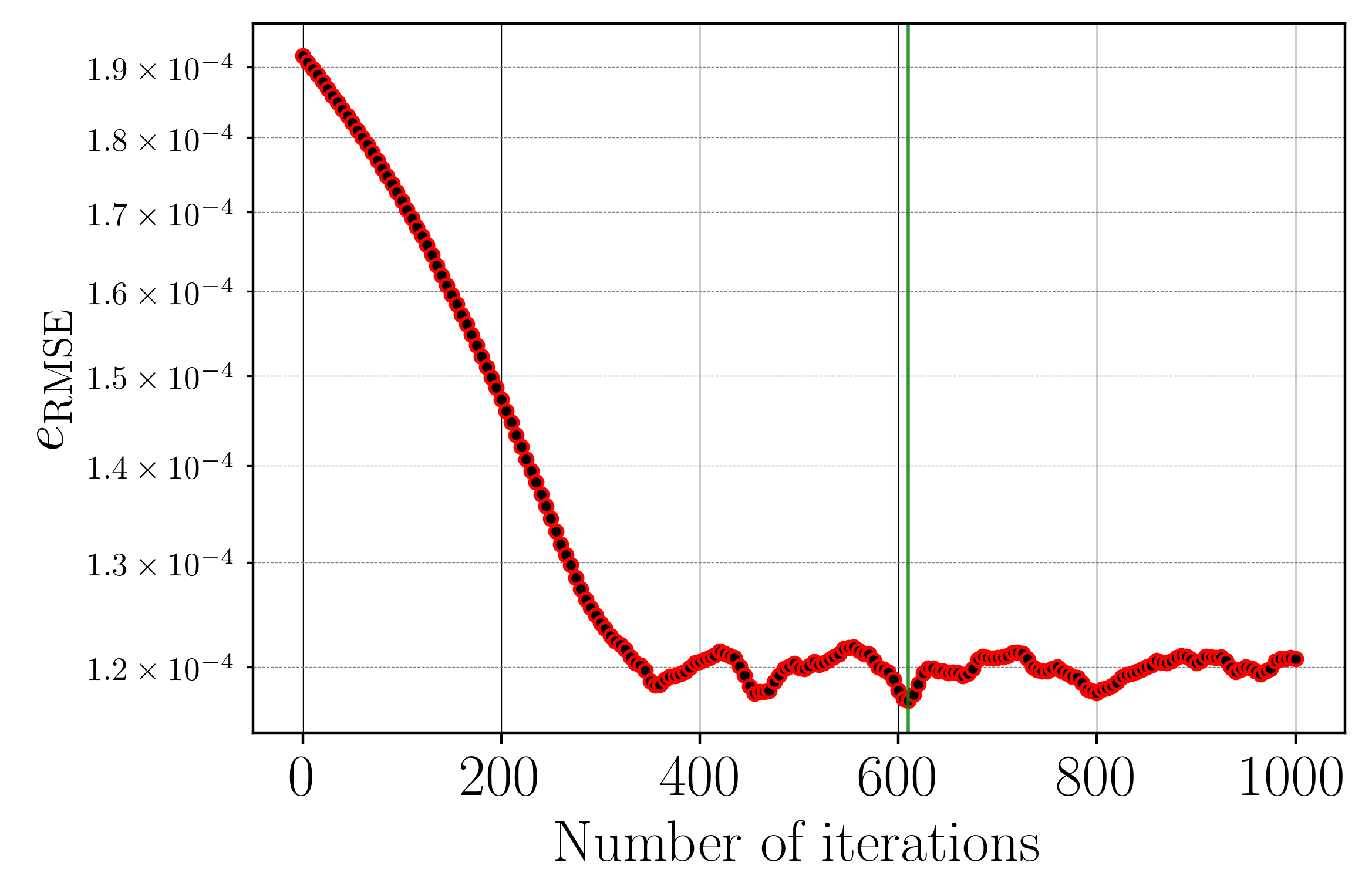}}
    \caption{Evolutions of the accuracy $\rho$, the nugget $\lambda$, and the error $\RMSE$ on the validation set as functions of the number of iterations for the parametric KF algorithm applied to the SSKRR surrogate of $\lift$. The green vertical lines correspond to the iteration where the error $\RMSE$ is minimal on the validation set.}\label{fig:UMRIDA_120_CL_KF}
\end{figure}

\begin{table}[h!]
    \begin{center}
        \begin{tabular} {|c||c|c|c|c|}
            \hline
%            \rowcolor{blue!20} \multicolumn{5}{ |c| }{$\lift$} \\
%            \hline\hline\hline
            \rowcolor{black!20} & SSKRR  & Sparse gPC & Full gPC & KRR\\
            \hline\hline
                $\RMSE$ & $7.574\times10^{-5}$ & $3.715\times10^{-4}$ &$1.159\times10^{-4}$ & $4.425\times\times10^{-4}$\\
                \hline
                $\NRMSE$ & $1.040\times10^{-4}$ & $5.103\times10^{-4}$ &$8.437\times10^{-5}$ & $6.079\times10^{-4}$\\
                \hline
                $\ERRrel$ & $0.0319\%$ & $0.232\%$ & $0.0368\%$ & $0.223\%$\\
                \hline
                $\Qtwo$ & $0.99996$ & $0.99911$ & $0.99995$ & $0.99874$\\
            \hline
        \end{tabular}
    \end{center}
    \caption{Comparison of the errors between the surrogate models of $\lift$ with $\Nobs = 80$ and $\NT = 25$ sampling points.}\label{tab:UMRIDA_120_CL_error}
\end{table}

The evolution of the accuracy $\rho$, the nugget $\lambda$, the length scales $\gamma_i$, $i=1,2,3$, and the error $\RMSE$ on the validation set as functions of the number of iterations of the parametric KF algorithm applied to the KRR surrogate, are shown on \Cref{fig:UMRIDA_120_CL_KRR-KF}. We find $\lambdamin = 1.48\times10^{-4}$ in this case, which corresponds to the green vertical lines on \Cref{fig:UMRIDA_120_CL_KRR-KF}. Likewise, the evolution of the accuracy $\rho$, the nugget $\lambda$, and the error $\RMSE$ on the validation set as functions of the number of iterations of the parametric KF algorithm applied to the SSKRR surrogate, are shown on \Cref{fig:UMRIDA_120_CL_KF}. We find $\lambdamin = 1.30\times10^{-7}$ in this case, which corresponds to the green vertical lines on \Cref{fig:UMRIDA_120_CL_KF}. Here one can notice that for the initial choice of $\lambda = 1\times10^{-6}$, the accuracy $\rho$ is already very small (about $1\times10^{-3}$). Therefore, the decrease in the error $\RMSE$ on the validation set is marginal and the changes in $\lambda$ are not substantial.

\begin{figure}[h!]
\centering\includegraphics[scale = 1]{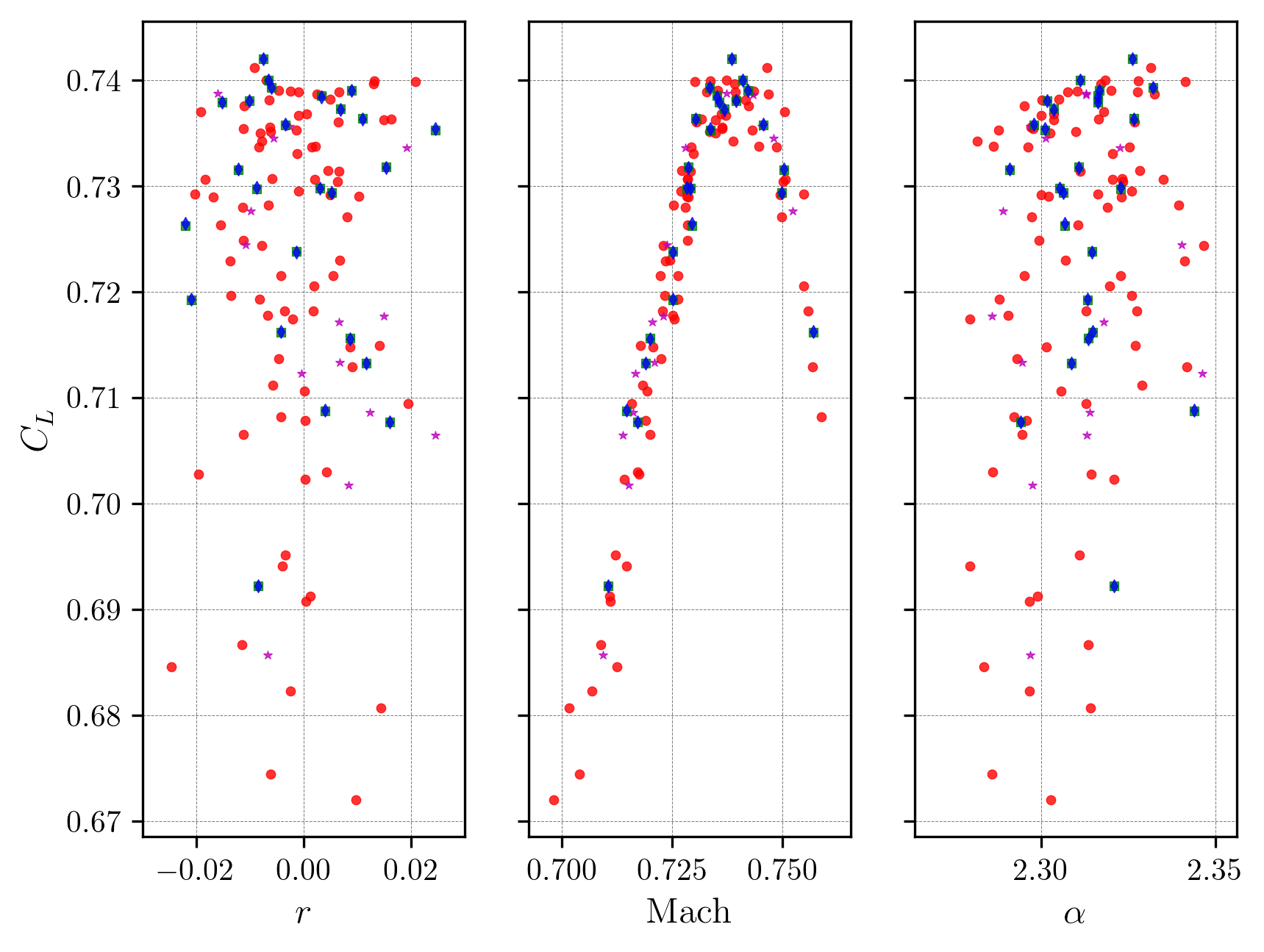}
\caption{Distribution of the difference between the prediction given by SSKRR surrogate and the observations on the test set for $\lift$ with $\Nobs = 80$ and $\NT = 25$ sampling points. The red circles are the observations defining the learning set. The purple stars are the observations defining the validation set. The green squares are the observations defining the test set with their corresponding predictions given by the SSKRR surrogate, depicted by the blue diamonds.}\label{fig:UMRIDA_120_CL_dist_diff}
\end{figure}

\begin{figure}[h!]
\centering\includegraphics[width=0.8\textwidth]{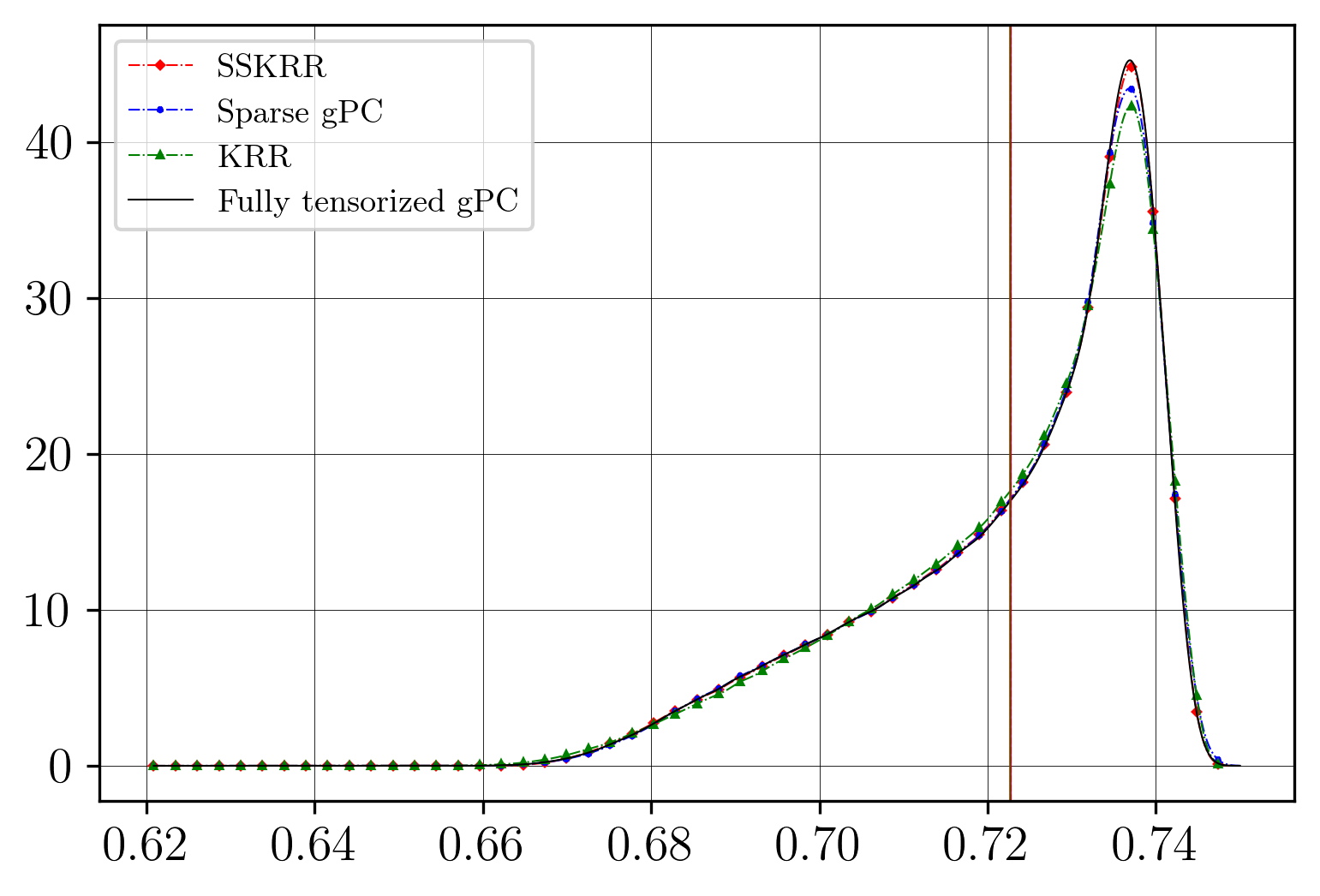}
\caption{The PDFs of $\lift$ using the four surrogate models. The vertical lines correspond to their respective expectation in \Cref{tab:UMRIDA_120_CL_meanvar}.}\label{fig:UMRIDA_120_CL_pdf}
\end{figure}

The comparison of the errors $\RMSE$, $\NRMSE$ and $\ERRrel$ between the surrogate models on the test set are given in \Cref{tab:UMRIDA_120_CL_error} where $\lambda = \lambdamin$. \Cref{fig:UMRIDA_120_CL_dist_diff} shows the values of $\lift$ with respect to the input parameters on the learning set, the verification set, the test set, and the predictions of the SSKRR surrogate. A strong non-linear dependence between $\lift$ and the Mach number $\Mach$ can be seen on \Cref{fig:UMRIDA_120_CL_dist_diff}. The PDFs of $\lift$ using the three surrogate models are estimated from $\NS = 1\times10^{6}$ random data points taken at random following the Beta distributions of \Cref{tab:input_param} and then smoothing out the resulting histograms by a normal kernel density function \cite{Wand95}. They are shown on \Cref{fig:UMRIDA_120_CL_pdf}, together with their corresponding expectation and variance in \Cref{tab:UMRIDA_120_CL_meanvar}. The expectations from the PDFs obtained by each surrogate model are shown on \Cref{fig:UMRIDA_120_CL_pdf} with vertical lines. Notice that we obtain comparable results except at the tails of the distributions and at their peaks. Finally, Sobol' main-effect sensitivity indices are gathered in \Cref{tab:sobol_indices_CL}. As expected from the previous results, the variable $\Xobs_2 = \Mach$ is more influential than the variables $\Xobs_1 = r$ or $\Xobs_3 = \Aoa$ where $\Xobs_1 = r$ has almost no influence on $\lift$. The SSKRR surrogate slightly outperforms the fully tensorized gPC one. Both have a much better performance than the sparse gPC and KRR surrogates.

\begin{table}[h!]
    \begin{center}
    \begin{tabular} {|c||c|c|}
        \hline
%        \rowcolor{blue!20} \multicolumn{3}{ |c| }{$\lift$} \\
%        \hline\hline\hline
        \rowcolor{black!20} & Expectation & Variance \\
            \hline\hline
            Full gPC & $72.273\times10^{-2}$ & $2.787\times10^{-4}$  \\
            \hline
            Sparse gPC & $72.278\times10^{-2}$ & $2.782\times10^{-4}$  \\
            \hline
            SSKRR & $72.274\times10^{-2}$ & $2.777\times10^{-4}$ \\
            \hline
            KRR & $72.266\times10^{-2}$ & $2.802\times10^{-4}$ \\
        \hline
    \end{tabular}
    \end{center}
    \caption{The expectation and variance of $\lift$ estimated from the four surrogates with $\NS=1\times10^{6}$ sampling points.}\label{tab:UMRIDA_120_CL_meanvar}
\end{table}

\begin{table}[h!]
    \begin{center}
        \begin{tabular} {|c||c|c|c|}
            \hline
%            \rowcolor{blue!20} \multicolumn{4}{ |c| }{$\lift$} \\
%            \hline\hline\hline
            \rowcolor{black!20} & $\Xobs_1 = r$ & $\Xobs_2 = \Mach$ & $\Xobs_3 = \Aoa$\\
            \hline\hline
            Full gPC & $0.00345$ & $0.955$ & $0.0286$ \\
            \hline
            Sparse gPC & $0.00358$ & $0.953$ & $0.0298$ \\
            \hline
            SSKRR & $0.00425$ & $0.956$ & $0.0296$ \\
            \hline
            KRR & $0.00392$ & $0.952$ & $0.0316$ \\
            \hline
        \end{tabular}
    \end{center}
    \caption{Sobol' main-effect sensitivity indices of $\lift$ with $\Nobs = 80$ sampling points. They were estimated from a pick-freeze estimator with a matrix size of $1\times10^{6}$ samples for the KRR and SSKRR surrogates, and from the polynomial expansion coefficients for the fully tensorized and sparse gPC surrogates.}
    \label{tab:sobol_indices_CL}
\end{table}

%-----------% Conclusion %-----------%
\section{Conclusions}

In this paper we have devised two classes of algorithms to design a kernel from data in kernel methods aimed to approximate regular functions.
The first class is kernel flow which iteratively learns the parameters of a base kernel in a variant of cross-validation, albeit a non parametric version exists where observations in the dataset are moved along directions that minimize the metric used for learning.
The second class is coined spectral kernel ridge regression. It builds on a suitable representation of the data in a carefully chosen basis to design a kernel which is expanded on that basis. Both approaches can be implemented concurrently. For example, we have used kernel flow to tune the nugget in a regressor derived by the spectral kernel ridge approach. Our numerical experiments suggest that the second class of algorithms increases the accuracy of the obtained approximations.
However it has been shown in some numerical experiments that the accuracy of kernel flow can be improved if it is trained with alternative metrics such as maximum mean discrepancy. Here we have only considered a relative error, and parameterized base kernels instead of mixtures of parameterized base kernels.
The algorithm proposed to address non sparse functions by spectral kernel ridge regression has not been tested so far, so we also aim to explore its performances in future works. Time series are ubiquitous in engineering science and we shall consider the application of kernel methods in transient aerodynamic forecasting as well, with kernels learned from data. In this respect, greedy approaches whereby the accuracy of the surrogates is improved by incrementally adding observations of the ground truth function may be worth considering, with the use of \emph{e.g.} Newton bases.

%-----------% Conclusion %-----------%
\clearpage
\section{Appendices}

\subsection{Reproducing kernel Hilbert space}\label{subsec:chap3_RKHS}

This appendix follows \cite[Chapter 1]{Berlinet2004} and \cite[Chapters 1 and 2]{Paulsen2016}. It is a short reminder on reproducing kernel Hilbert spaces (RKHS). The set of functions from $\InputSpace$ to $\Rset$ is denoted by $\FS(\InputSpace,\Rset)$, which is a vector space over $\Rset$ with the operations of addition and scalar multiplication.
\begin{definition}[RKHS]\label{def:RKHS}
    Let $\InputSpace$ be a non-empty set. A subset $\HS \subseteq \FS(\InputSpace,\Rset)$ is called a RKHS on $\InputSpace$ if
    \begin{itemize}
        \item $\HS$ is a vector subspace of $\FS(\InputSpace,\Rset)$;
        \item $\HS$ is endowed with an inner product $\lrangle{\cdot,\cdot}_{\HS}$, with respect to which $\HS$ is a Hilbert space;
        \item for every $\x\in\InputSpace$, the linear evaluation functional $E_{\x}:\HS\to\Rset$ defined by $E_{\x}(f) = f(\x)$ is bounded\string: $\exists C_{\x} > 0,\; \forall f \in \HS,\; \absolute{f(\x)} = \absolute{E_{\x}(f)} \leq C_{\x} \norm{f}{\HS}$, where $\norm{f}{\HS}=\smash{\sqrt{\lrangle{f,f}_{\HS}}}$. 
    \end{itemize}
\end{definition}
\noindent If $\HS$ is a RKHS, then the Riesz representation theorem shows that the linear evaluation functional $E_{\x}$ is given by the inner product with a unique vector in $\HS$; that is, $\forall\x\in\InputSpace$, $\exists!\Ker(\x, \cdot)\in\HS$ such that $\forall f\in\HS$, $E_{\x}(f) = f(\x)=\lrangle{f,\Ker(\x, \cdot)}_{\HS}$.
\begin{definition}[Reproducing kernel \cite{Paulsen2016}]\label{def:RK}
The function $\Ker(\x, \cdot)$ is called the reproducing kernel for $\x$. The function $\x,\y\mapsto\Ker(\x,\y):\InputSpace\times\InputSpace \to \Rset$ defined by $\Ker(\x, \y) = \lrangle{\Ker(\y, \cdot),\Ker(\x, \cdot)}_{\HS}$ is called the reproducing kernel for  $\HS$.
\end{definition}
\noindent In other words, $\Ker(\x, \y)$ is for $\x,\y$ in $\InputSpace$ the evaluation of the function $\Ker(\y,\cdot)$ of $\HS$ at $\x$. Following the previous definition, we introduce a kernel function as follows:
\begin{definition}[Kernel function \cite{Paulsen2016}]\label{def:Kernel}
    Let $\InputSpace$ be a non-empty set and let $\Ker: \InputSpace\times\InputSpace \to \Rset$ be a function. $\Ker$ is called a kernel function if it is symmetric and positive semi-definite that is, for any $m \geq 1$, for any $(a_1,\dots a_m) \in \Rset^m$, for any $(\x_1,\dots\x_m) \in \InputSpace^{m}$,
    \begin{equation}\label{eq:kernel_pos_semi_def}
        \sum\limits_{i=1}^{m}\sum\limits_{j=1}^{m}a_ia_j\Ker(\x_i, \x_j) \geq 0\,.
    \end{equation}
\end{definition}

\begin{proposition}\label{prop:RK_sym_dp}
    Let $\InputSpace$ be a non-empty set and let $\HS$ be a RKHS on $\InputSpace$ with reproducing kernel $\Ker$. Then $\Ker$ is a kernel function.
\end{proposition}
\noindent Indeed let $m \geq 1$ and $(\x_1,\dots\x_m) \in \InputSpace^m$, $(a_1,\dots a_m) \in \Rset^m$; then one has:
\begin{equation*}
\sum\limits_{i=1}^{m}\sum\limits_{j=1}^{m}a_ia_j\Ker(\x_i, \x_j) = \lrangle{\sum\limits_{j=1}^{m} a_j\Ker(\x_j, \cdot), \sum\limits_{i=1}^{m} a_i\Ker(\x_i, \cdot)}_{\HS} = \norm{\sum\limits_{i=1}^{m} a_i\Ker(\x_i, \cdot)}{\HS}^{2} \geq 0\,.
\end{equation*}
\noindent In addition, let $\x, \y \in \InputSpace$; one has:
    \begin{equation*}
        \Ker(\x, \y) = \lrangle{\Ker(\y, \cdot),\Ker(\x, \cdot) \rangle_{\HS} = \langle \Ker(\x, \cdot),\Ker(\y, \cdot)}_{\HS} = \Ker(\y, \x)\,,
    \end{equation*}
which shows that $\Ker$ is symmetric. From \Cref{prop:RK_sym_dp}, a RKHS $\HS$ defines a reproducing kernel $\Ker$ which is a symmetric kernel function. Although \Cref{prop:RK_sym_dp} is quite elementary, it has a deep converse known as the Moore-Aronszajn theorem.
\begin{theorem}\cite[Theorem 3]{Berlinet2004}, \cite[Theorem 2.14]{Paulsen2016}\label{th:Moore_Aron}
    Let $\InputSpace$ be a non-empty set and let $\Ker: \InputSpace\times\InputSpace \to \Rset$ be a function. If $\Ker$ is a kernel function, then there exists a unique RKHS $\HS$ of functions on $\InputSpace$ such that $\Ker$ is the reproducing kernel of $\HS$.
\end{theorem}
\noindent Therefore, \Cref{prop:RK_sym_dp} and \Cref{th:Moore_Aron} show there is a one-to-one correspondence between RKHS on a set and kernel functions on this set. In this paper, the unique RKHS associated with the kernel function $\Ker$ is denoted by $\RKHS$, and $\lrangle{\cdot,\cdot}_{\RKHS}$ is its inner product with the associated norm $\norm{\cdot}{\RKHS}$:
\begin{definition}
    Given a kernel function $\Ker: \InputSpace\times\InputSpace \to \Rset$, $\RKHS$ denotes the unique RKHS with reproducing kernel $\Ker$.
\end{definition}
\noindent One more useful assumption about the kernel function $\Ker$ is made in this work, namely:
\begin{assumption}[Positive definite]\label{assu:ker_pd}
Let $\InputSpace$ be a non-empty set and let $\Ker: \InputSpace\times\InputSpace \to \Rset$ be a kernel function. $\Ker$, which is thus symmetric, is assumed to be positive definite, or non-degenerate, that is, for any $m \geq 1$, for any ${\bf a}=(a_1,\dots a_m) \in \Rset^m$, ${\bf a} \neq {\bf 0}$, for any $(\x_1,\dots\x_m) \in \InputSpace^{m}$,
    \begin{equation*}
        \sum\limits_{i=1}^{m}\sum\limits_{j=1}^{m}a_ia_j\Ker(\x_i, \x_j) > 0\,.
    \end{equation*}
\end{assumption}

\subsection{Examples of parametric kernel}\label{subsec:examples_kernel}

We denote by $\norm{\x}{p} = \smash{(\sum_{j=1}^{d}\absolute{x_j}^{p})^{\frac{1}{p}}}, \; p > 0$ with $\norm{\x}{0} = \# \{j\;;x_j \neq 0\}$ the $p$-norm of the vector $\x$. For practical cases, many different kernels $\Ker $ are available and for citing the most encountered ones \cite{Rasmussen06, Stein99}, with $\x,\y\in\Rset^\Dim$:
\begin{itemize}
	\item Polynomial kernel defined as 
	\begin{equation}\label{eq:Polynomial_kernel}
		\Ker(\x, \y;\bm{\theta}) = \left(b + \x^\itr\y\right)^{p}\,,
	\end{equation}
	where $b \geq 0$ and $p > 0 $ are parameters, and $\bm{\theta}\equiv(b,p)$;
	\item Gaussian kernel (also known as squared exponential) defined as 
	\begin{equation}\label{eq:Gaussian_kernel}
		\Ker(\x,\y;\theta) = \exp\left(-\frac{\norm{\x - \y}{2}^2}{\gamma^{2}}\right)\,, 
	\end{equation}
	where $\gamma > 0 $ is called the length scale, and $\theta\equiv\gamma$ is a unique parameter. Alternatively, different length scales $\bm{\theta}\equiv(\gamma_i)_{i=1}^\Dim$ can be chosen for the input dimensions\string:
	\begin{equation}\label{eq:Gaussian_kernel_ARD}
		\Ker(\x,\y;\bm{\theta}) = \exp\left(-\sum\limits_{i=1}^{\Dim}\frac{\absolute{x_i - y_i}{}^2}{\gamma_i^{2}}\right)\,;
	\end{equation}
	\item Mat\'{e}rn-like kernels defined as
	\begin{equation}\label{eq:Matern_kernel}
		\Ker(\x,\y;\bm{\theta}) = \frac{2^{1-\nu}}{\Gamma(\nu)}\left(\frac{\sqrt{2\nu}\norm{\x - \y}{2}}{\gamma}\right)^{\nu} B_{\nu}\left(\frac{\sqrt{2\nu}\norm{\x - \y}{2}}{\gamma}\right)\,,
	\end{equation}
	where $\gamma > 0 $ is called the length scale, $\nu$ is a positive parameter, $\bm{\theta}\equiv(\gamma,\nu)$, and $B_{\nu}$ is the modified Bessel function. The most used ones are for $\nu = 3/2$ and $\nu=5/2$. Note that for $\nu \to +\infty$, the Gaussian kernel \pref{eq:Gaussian_kernel} is recovered. Stein in \cite{Stein99} named this type of kernels after the work of Mat\'{e}rn \cite{Matern60};
	\item Rational Quadratic (RQ) kernel defined as 
		\begin{equation}\label{eq:RQ_kernel}
			\Ker(\x,\y;\bm{\theta})  = \left(1 + \frac{\norm{\x - \y}{2}^{2}}{2\alpha\gamma^{2}}\right)^{-\alpha}\,,
		\end{equation}
		with $\alpha > 0$, $\gamma > 0$, and $\bm{\theta}\equiv(\alpha,\gamma)$. The case $\alpha \to +\infty$ corresponds to the Gaussian kernel \pref{eq:Gaussian_kernel}.
\end{itemize}
Stein \cite{Stein99} argues that the Gaussian kernel is too smooth for modeling many physical systems and recommends to use Mat\'{e}rn-like kernels.

\subsection{Mercer's theorem}\label{subsec:mercer}

This appendix is a brief summary on Mercer's framework, which allows us to express a Mercer kernel as a function of eigenvalues and eigenvectors of its associated integral operator.
\begin{definition}\label{df:MercerKernel}
	Let $\InputSpace$ be a compact subset of $\mathbb{R}^{\Dim}$. A function $\Ker : \InputSpace\times\InputSpace\to\Rset$ is called a Mercer kernel if it is continuous, symmetric, and positive semi-definite in the sense of \Cref{eq:kernel_pos_semi_def}.
\end{definition}
\noindent This definition allows us to state Mercer's theorem \cite[Chapter 11]{Paulsen2016}:
\begin{theorem}[Mercer's theorem]\label{th:Mercer}
    Let $\mu$ be a finite Borel measure with support $\InputSpace$ and let $\Ltwo(\InputSpace, \mu)$ be the set of square integrable functions on $\InputSpace$ with respect to $\mu$. Let $\Ker$ be a Mercer kernel on $\InputSpace$ and let $\TK : \Ltwo(\InputSpace, \mu) \to \Ltwo(\InputSpace, \mu)$ be the associated integral operator defined by $\forall f\in\Ltwo(\InputSpace, \mu), \forall \x\in\InputSpace, \TK f(\x) = \int_{\InputSpace}\Ker(\x,\y)f(\y)\mu(d\y)$. Then there exists a countable, orthonormal collection of functions $\{\eigenV_i\}_{i\in\Nset}$ of $\Ltwo(\InputSpace, \mu)$ which are eigenvectors of $\TK$ with associated non-negative eigenvalues $\{\eigenv_i\geq 0\}_{i\in\Nset}$. Moreover, taking the eigenvectors corresponding to the non-zero eigenvalues, they are continuous functions on $\InputSpace$ and $\Ker(\x,\y)$ has the following representation:
    \begin{equation*}
        \Ker(\x,\y) = \sum\limits_{i=0}^{+\infty}\eigenv_i \eigenV_i(\x) \otimes \eigenV_i(\y)\,,
    \end{equation*}
    where the series converges absolutely and uniformly:
    \begin{equation*}
        \lim\limits_{n\to+\infty}\sup\limits_{\x,\y \in \InputSpace}\absolute{\Ker(\x,\y) - \sum\limits_{i=0}^{n}\eigenv_i \eigenV_i(\x) \otimes \eigenV_i(\y)}{} = 0\,.
    \end{equation*}
\end{theorem}
\noindent From Mercer's theorem, $\TK$ is a trace class operator with:
\begin{equation}\label{eq:TK_trace_class}
    \text{Tr}(\TK) = \int\limits_{\InputSpace} \Ker(\x,\x) \mu(d\x) = \sum\limits_{i=0}^{+\infty}\eigenv_i < +\infty\,.
\end{equation}

\begin{remark}
Since $\TK$ is self-adjoint and compact, $\{\eigenV_i\}_{i\in\mathbb{N}}$ is a basis of $\Ltwo(\InputSpace, \mu)$.
We remind that a family of functions $\{\eigenV_i\}_{i\in\Nset}$ is a basis of a Banach space $H$ if $\forall f \in H,\; \exists! \{\alpha_i\}_{i\in\Nset}$ such that $\forall \epsilon > 0$, $\exists N$ such that $\forall n \geq N$
\begin{equation*}
\norm{f - \sum\limits_{i=0}^{n}\alpha_i\eigenV_{i}}{H} \leq \epsilon\,.
\end{equation*}
\end{remark}
\noindent Mercer's theorem also allows us to define explicitly the RKHS $\RKHS$ associated with the kernel $\Ker$, on the condition that $\Ker$ is a Mercer kernel. Indeed, one has \cite[Theorem 11.18]{Paulsen2016}:
\begin{equation*}
    \RKHS = \left\{ f\in\Ltwo(\InputSpace, \mu), f = \sum\limits_{i=0}^{+\infty}\lrangle{ f,\eigenV_i}_{\Ltwo}\eigenV_i \text{ with } \sum\limits_{i=0}^{+\infty} \frac{\lrangle{ f,\eigenV_i}_{\Ltwo}^{2}}{\eigenv_i} < + \infty \right\}\,,
\end{equation*}
where $\lrangle{ \cdot,\cdot}_{\Ltwo}$ is the inner product of $\Ltwo(\InputSpace, \mu)$. The inner product of $\RKHS$ is given as $\forall (f,g) \in\RKHS\times\RKHS$,
\begin{equation*}
%    \lrangle{ f,g }_{\RKHS} = \sum\limits_{i=0}^{+\infty} \frac{f_i g_i}{\eigenv_i},
\lrangle{ f,g }_{\RKHS} = \sum\limits_{i=0}^{+\infty} \frac{\lrangle{ f,\eigenV_i}_{\Ltwo}\lrangle{ g,\eigenV_i }_{\Ltwo}}{\eigenv_i}\,.
\end{equation*}
%where $f_i = \lrangle{ f,\eigenV_i}_{\Ltwo}$ and $g_i = \lrangle{ g,\eigenV_i }_{\Ltwo}$.
Consequently, the norm $\norm{f}{\RKHS}$ reads:
\begin{equation}\label{eq:norm_RKHS}
    \norm{f}{\RKHS}^{2} = \sum\limits_{i=0}^{+\infty} \frac{\lrangle{ f,\eigenV_i}_{\Ltwo}^{2}}{\eigenv_i}\,,
\end{equation}
which gives its expression in terms of the eigenvalues and eigenvectors of $\Ker$. One can notice that $\lrangle{ f,\eigenV_i}_{\Ltwo} = \eigenv_i\lrangle {f,\eigenV_i }_{\RKHS}$, so one also has:
\begin{equation*}
    \lrangle{ f,g }_{\RKHS} = \sum\limits_{i=0}^{+\infty} \eigenv_i \lrangle{f,\eigenV_i}_{\RKHS} \lrangle{g,\eigenV_i}_{\RKHS}\,.
\end{equation*}
\begin{remark}\label{rk:independent}
    The functional space $\RKHS$ does not depend on the measure $\mu$, actually. Only the eigenvectors $\{\eigenV_i\}_{i\in\mathbb{N}}$ and the eigenvalues $\{\eigenv_i\geq 0\}_{i\in\mathbb{N}}$ do.
\end{remark}
\begin{remark}
    It can be shown that $\{\sqrt{\eigenv_i}\eigenV_i\}_{i\in\mathbb{N}}$ is an orthonormal basis of $\RKHS$; see \cite[Theorem 11.18]{Paulsen2016}.
\end{remark}

%-----------% Bibliography %-----------%
\bibliographystyle{plain}

\begin{thebibliography}{99}\label{pg:biblio}

\bibitem{Berke93} \href{https://doi.org/10.1016/0045-7949(93)90435-G}{L. Berke, S. N. Patnaik, and P. L. N. Murthy, Optimum design of aerospace structural components using neural networks, \emph{Comput. Struct.}, \textbf{48}(6)\string:1001--1010, 1993}

\bibitem{Berkooz93} \href{https://doi.org/10.1146/annurev.fl.25.010193.002543}{G. Berkooz, P. Holmes, and J. L. Lumley, The proper orthogonal decomposition in the analysis of turbulent flows, \emph{Annu. Rev. Fluid Mech.}, \textbf{25}\string:539--575, 1993.}

\bibitem{Berlinet2004} \href{https://doi.org/10.1007/978-1-4419-9096-9}{A. Berlinet and C. Thomas-Agnan, \emph{Reproducing Kernel Hilbert Spaces in Probability and Statistics}, Springer, New York NY, 2004.}

\bibitem{Blatman2011} \href{https://doi.org/10.1016/j.jcp.2010.12.021}{G. Blatman and B. Sudret, Adaptive sparse polynomial chaos expansion based on {\it least angle regression}, \emph{J. Comput. Phys.}, \textbf{230}(6):2345--2367, 2011.}

\bibitem{Boufounos07} \href{https://doi.org/10.1109/SSP.2007.4301267}{P. Boufounos, M. F. Duarte, and R. G. Baraniuk, Sparse signal reconstruction from noisy compressive measurements using cross validation, In \emph{SSP'07: Proceedings of the 2007 IEEE/SP 14th Workshop on Statistical Signal Processing, 26-29 August 2007, Madison WI}, pp 299--303, 2007.}

\bibitem{Bouhlel19} \href{https://doi.org/10.1016/j.advengsoft.2019.03.005}{M. A. Bouhlel, J. T. Hwang, N. Bartoli, R. Lafage, J. Morlier, and J. R. R. A. Martins, A Python surrogate modeling framework with derivatives, \emph{Adv. Eng. Softw}, \textbf{135}:102662, 2019.}

\bibitem{Bui04} \href{https://doi.org/10.2514/1.2159}{T. Bui-Thanh, M. Damodaran, and K. Willcox, Aerodynamic data reconstruction and inverse design using Proper Orthogonal Decomposition, \emph{AIAA J.}, \textbf{42}(8)\string:1505--1516, 2004.}

\bibitem{elsa} \href{https://doi.org/10.1051/meca/2013056}{L. Cambier, S. Heib, and S. Plot, The Onera \textit{elsA} CFD software: Input from research and feedback from industry, \emph{Mechanics \& Industry}, \textbf{14}(3)\string:159--174, 2013.}

\bibitem{Candes05} \href{https://doi.org/10.1109/TIT.2005.858979}{E. J. Cand{\`{e}}s and T. Tao, Decoding by linear programming, \emph{IEEE Trans. Inf. Theory}, \textbf{51}(12)\string:4203--4215, 2005}

\bibitem{Candes06} \href{https://doi.org/10.1002/cpa.20124}{E. J. Cand{\`{e}}s, J. K. Romberg, and T. Tao, Stable signal recovery from incomplete and inaccurate measurements, \emph{Comm. Pure Appl. Math.}, \textbf{59}(8)\string:1207--1223, 2006.}

\bibitem{Candes08} \href{https://doi.org/10.1016/j.crma.2008.03.014}{E. J. Cand{\`{e}}s, The restricted isometry property and its implications for compressed sensing, \emph{C. R. Math.}, \textbf{346}(9-10):589--592, 2008.}

\bibitem{Candes08_intro} \href{https://doi.org/10.1109/MSP.2007.914731}{E. J. Cand{\`{e}}s and M. B.Wakin, An introduction to compressive sampling, \emph{IEEE Signal Process. Mag.}, \textbf{25}(2)\string:21--30, 2008.}

\bibitem{Candes11} \href{https://doi.org/10.1109/TIT.2011.2161794}{E. J. Cand{\`{e}}s and Y. Plan, A probabilistic and RIPless theory of compressed sensing, \emph{IEEE Trans. Inf. Theory}, \textbf{57}(11):7235--7254, 2011.}

\bibitem{Chatterjee2000} \href{https://www.jstor.org/stable/24103957}{A. Chatterjee, An introduction to the proper orthogonal decomposition, \emph{Current Science}, \textbf{78}(7)\string:808--817, 2000.}

\bibitem{Chen06} \href{https://doi.org/10.1137/S1064827596304010}{S. S. Chen, D. L. Donoho, and M. A. Saunders, Atomic decomposition by basis pursuit, \emph{SIAM J. Sci. Comput.}, \textbf{20}(1):33--61, 2006.}

\bibitem{Chen21} \href{https://doi.org/10.1090/mcom/3649}{Y. Chen, H. Owhadi, and A. M. Stuart, Consistency of empirical Bayes and kernel flow for hierarchical parameter estimation, \emph{Math. Comput.}, \textbf{90}\string:2527--2578, 2021.}

\bibitem{Chen21b} \href{https://doi.org/10.1016/j.jcp.2021.110668}{Y. Chen, B. Hosseini, H. Owhadi, and A. M. Stuart, Solving and learning nonlinear PDEs with Gaussian processes, \emph{J. Comput. Phys.}, \textbf{447}\string:110668, 2021.}

\bibitem{Chkifa15} \href{https://doi.org/10.1016/j.matpur.2014.04.009}{A. Chkifa, A. Cohen, and C. Schwab, Breaking the curse of dimensionality in sparse polynomial approximation of parametric PDEs, \emph{J. Math. Pures Appl.}, \textbf{103}(2)\string:400--428, 2015.}

\bibitem{Cook1979} \href{http://eda-ltd.com.tr/caeeda\_doc/AGARD-AR-138.pdf}{P. H. Cook, M. A. McDonald, and M. C. P. Firmin, Aerofoil {RAE} 2822---Pressure distributions, and boundary layer and wake measurements, In \emph{Experimental Data Base for Computer Program Assessment. {AGARD} Advisory Report No. 138}, NATO, May 1979.}

\bibitem{Cressie1993} \amend{\href{https://doi.org/10.1002/9781119115151}{N. A. C. Cressie, \emph{Statistics for Spatial Data}, John Wiley \& Sons, New York NY, 1993.}}

\bibitem{Darcy21} \href{}{M. Darcy, B. Hamzi, J. Susiluoto, A. Braverman, and H. Owhadi, Learning dynamical systems from data: a simple cross-validation perspective, part II: nonparametric kernel flows, 2021.}

\bibitem{Davis2002} \amend{\href{https://www.wiley.com/en-us/9780471172758}{J. C. Davis, \emph{Statistics and Data Analysis in Geology}, 3rd Edition, John Wiley \& Sons, New York NY, 2002.}}

\bibitem{Donoho06} \href{https://doi.org/10.1109/TIT.2006.871582}{D. L. Donoho, Compressed sensing, \emph{IEEE Trans. Inf. Theory}, \textbf{52}(4)\string:1289--1306, 2006.}

\bibitem{Doostan11} \href{https://doi.org/10.1016/j.jcp.2011.01.002}{A. Doostan and H. Owhadi, A non-adapted sparse approximation of PDEs with stochastic inputs, \emph{J. Comput. Phys.}, \textbf{230}(8)\string:3015--3034, 2011.}

\bibitem{Dumont2019} \href{https://doi.org/10.1007/978-3-319-77767-2_14}{A. Dumont, J.-L. Hantrais-Gervois, P.-Y. Passaggia, J. Peter, I. Salah el Din, and {\'E}. Savin, Ordinary kriging surrogates in aerodynamics, In \emph{Uncertainty Management for Robust Industrial Design in Aeronautics (C. Hirsch, D. Wunsch, J. Szumbarski, \L. {\L}aniewski-Wo{\l\l}k, and J. Pons-Prats, eds.)}, pp. 229--245, Springer, Cham, 2019.}

\bibitem{Ernst12} \href{https://doi.org/10.1051/m2an/2011045}{O. G. Ernst, A. Mugler, H.-J. Starkloff, and E. Ullmann, On the convergence of generalized polynomial chaos expansions, \emph{ESAIM\string: M2AN}, \textbf{46}(2)\string:317--339, 2012.}

\bibitem{Forrester09} \href{https://doi.org/10.1016/j.paerosci.2008.11.001}{A. I. J. Forrester and A. J. Keane, Recent advances in surrogate-based optimization, \emph{Prog. Aerosp. Sci.}, \textbf{45}(1)\string:50--79, 2009.}

\bibitem{Gpytorch} \href{https://proceedings.neurips.cc/paper/2018/file/27e8e17134dd7083b050476733207ea1-Paper.pdf}{J. Gardner, G. Pleiss, K. Q. Weinberger, D. Bindel, and A. G. Wilson, GPyTorch: Blackbox matrix-matrix Gaussian process inference with GPU acceleration, In \emph{Advances in Neural Information Processing Systems 31}, NeurIPS 2018, Montreal, 2018.}

\bibitem{Garner1966} \href{}{H. C. Garner, E. W. E. Rogers, W. E. A. Acum, and E. C. Maskell, Subsonic wind tunnel wall corrections, {AGARD}o-graph 109, NATO, 1966.}

\bibitem{Ghanem91} \href{https://doi.org/10.1007/978-1-4612-3094-6}{R. G. Ghanem and P. D. Spanos, \emph{Stochastic Finite Elements\string: A Spectral Approach}, Springer, New York NY, 1991.}

\bibitem{Goodfellow16} \href{https://mitpress.mit.edu/books/deep-learning}{I. Goodfellow, Y. Bengio, and A. Courville, \emph{Deep Learning}, MIT Press, Cambridge MA, 2016.}

\bibitem{Haase1993} \href{https://doi.org/10.1007/978-3-663-14131-0}{W. Haase, F. Bradsma, E. Elsholz, M. Leschziner, and D. Schwamborn, \emph{{EUROVAL}---{A}n {E}uropean {I}nitiative on {V}alidation of {CFD} {C}odes}, Vieweg Verlag, Wiesbaden, 1993.}

\bibitem{Hadigol18} \href{https://doi.org/10.1016/j.cma.2017.12.019}{M. Hadigol and A. Doostan, Least squares polynomial chaos expansion\string: A review of sampling strategies, \emph{Comput. Methods Appl. Mech. Eng.}, \textbf{332}\string:382--407, 2018.}

\bibitem{Hampton16} \href{https://doi.org/10.1007/978-3-319-11259-6_67-1}{J. Hampton and A. Doostan, Compressive sampling methods for sparse polynomial chaos expansions, In \emph{Handbook of Uncertainty Quantification (R. Ghanem, D. Higdon, and H. Owhadi, eds.)}, pp. 827--855, Springer, Cham, 2016.}

\bibitem{Hamzi21_dyn} \href{https://doi.org/10.1016/j.physd.2020.132817}{B. Hamzi and H. Owhadi, Learning dynamical systems from data: A simple cross-validation perspective, part I: Parametric kernel flows, \emph{Physica D}, \textbf{421}\string:132817, 2021.}

\bibitem{Hamzi21_forecast} \href{https://doi.org/10.1098/rspa.2021.0326}{B. Hamzi, R. Maulik, and H. Owhadi, Simple, low-cost and accurate data-driven geophysical forecasting with learned kernels, \emph{Proc. R. Soc. A}, \textbf{477}(2252)\string:20210326, 2021.}

\bibitem{Hastie2009} \href{https://doi.org/10.1007/978-0-387-84858-7}{T. Hastie, R. Tibshirani, and J. Friedman, \emph{The Elements of Statistical Learning}, Springer-Verlag, New York NY, 2009.}

\bibitem{Ishigami99} \href{https://doi.org/10.1109/ISUMA.1990.151285}{T. Ishigami and T. Homma, An importance quantification technique in uncertainty analysis for computer models, In \emph{Proceedings First International Symposium on Uncertainty Modeling and Analysis, 3-5 December 1990, College Park MD}, pp. 398--403, 1990.}

\bibitem{Janon14} \href{https://doi.org/10.1051/ps/2013040}{A. Janon, T. Klein, A. Lagnoux, M. Nodet, and C. Prieur, Asymptotic normality and efficiency of two Sobol index estimators, \emph{ESAIM\string: PS}, \textbf{18}\string:342--364, 2014.}

\bibitem{Kadri16} \href{http://jmlr.org/papers/v17/11-315.html}{H. Kadri, E. Duflos, P. Preux, S. Canu, A. Rakotomamonjy, J. Audiffren, Operator-valued kernels for learning from functional response data, \emph{J. Mach. Learn. Res.}, \textbf{17}(20):1--54, 2016.}

\bibitem{Karniadakis21} \href{https://doi.org/10.1038/s42254-021-00314-5}{G. E. Karniadakis, I. G. Kevrekidis, L. Lu, P. Perdikaris, S. Wang, and L. Yang, Physics-informed machine learning, \emph{Nat. Rev. Phys.}, \textbf{3}(6)\string:422--440, 2021.}

\bibitem{Kingma2017adam} \href{https://arxiv.org/abs/1412.6980}{D. P. Kingma and J. Ba, Adam: A method for stochastic optimization, arXiv.org:1412.6980, 2017.}

\bibitem{Kleijnen09} \href{https://doi.org/10.1016/j.ejor.2007.10.013}{J. Kleijnen, Kriging metamodeling in simulation: a review, \emph{Eur. J. Oper. Res.}, \textbf{192}(3)\string:707--716, 2009.}

\bibitem{Kosambi43} \href{https://doi.org/10.1007/978-81-322-3676-4_15}{D. D. Kosambi, Statistics in function space, \emph{J. Indian Math. Soc.}, \textbf{7}\string:76--88, 1943.}

\bibitem{Laurenceau08} \href{https://doi.org/10.2514/1.32308}{J. Laurenceau and P. Sagaut, Building efficient response surfaces of aerodynamic functions with Kriging and Cokriging, \emph{AIAA J.}, \textbf{46}(2)\string:498--507, 2008.}

\bibitem{LeMaitre10} \href{https://doi.org/10.1007/978-90-481-3520-2}{O. Le Ma\^{\i}tre and O. Knio, \emph{Spectral Methods for Uncertainty Quantification. With Applications to Computational Fluid Dynamics}, Springer, Dordrecht, 2010.}

\bibitem{Marrel09} \href{https://doi.org/10.1016/j.ress.2008.07.008}{A. Marrel, B. Iooss, B. Laurent, and O. Roustant, Calculations of Sobol indices for the Gaussian process metamodel, \emph{Reliab. Eng. Syst. Safety}, \textbf{94}(3)\string:742--751, 2009.}

\bibitem{Matern60} \href{https://doi.org/10.1007/978-1-4615-7892-5}{B. Mat\'{e}rn, \emph{Spatial Variation}, Springer-Verlag, New York NY, 1960.}

\bibitem{Mathelin12} \href{https://doi.org/10.4208/cicp.151110.090911a}{L. Mathelin and K. A. Gallivan, A compressed sensing approach for partial differential equations with random input data, \emph{Commun. Comput. Phys.}, \textbf{12}(4)\string:919--954, 2012.}

\bibitem{Mathelin12b} \href{https://doi.org/10.1007/s00162-011-0235-9}{L. Mathelin, L. Pastur, and O. Le Ma\^{\i}tre, A compressed-sensing approach for closed-loop optimal control of nonlinear systems, \emph{Theor. Comput. Fluid Dyn.}, \textbf{26}(1-4)\string:319--337, 2012.}

\bibitem{Matheron63} \amend{\href{https://doi.org/10.2113/gsecongeo.58.8.1246}{G. Matheron, Principles of geostatistics, \emph{Economic Geology}, \textbf{58}(8)\string:1246--1266, 1963.}}

\bibitem{Micchelli77} \href{https://doi.org/10.1007/978-1-4684-2388-4_1}{C. A. Micchelli and T. J. Rivlin, A survey of optimal recovery, In \emph{Optimal Estimation in Approximation Theory (C. A. Micchelli and T. J. Rivlin, eds.)}, pp 1--54, Springer, Boston MA, 1977.}

\bibitem{Micchelli04} \href{https://doi.org/10.5555/2976040.2976156}{C. A. Micchelli and M. Pontil, Kernels for multi-task learning, In \emph{NIPS'04: Proceedings of the 17th International Conference on Neural Information Processing}, 921--928, 2004.}

\bibitem{Montgomery2004} \href{https://www.wiley.com/en-us/9781119492443}{D. C. Montgomery, \emph{Design and Analysis of Experiments}, John Wiley and Sons, New York NY, 2004.}

\bibitem{Nguyen99} \href{https://doi.org/10.1016/S0307-904X(99)00006-2	}{T. Nguyen-Thien and T. Tran-Cong, Approximation of functions and their derivatives: A neural network implementation with applications, \emph{Appl. Math. Model}, \textbf{23}(9)\string:687--704, 1999.}

\bibitem{Nouy10} \href{https://doi.org/10.1007/s11831-010-9054-1}{A. Nouy, Proper generalized decompositions and separated representations for the numerical solution of high dimensional stochastic problems. \emph{Arch. Comput. Methods Eng.}, \textbf{17}\string:403--434, 2010.}

\bibitem{Scovel19} \href{https://doi.org/10.1017/9781108594967}{H. Owhadi and C. Scovel, \emph{Operator-Adapted Wavelets, Fast Solvers, and Numerical Homogenization: From a Game Theoretic Approach to Numerical Approximation and Algorithm Design}, Cambridge University Press, Cambridge, 2019.}

\bibitem{Owhadi19} \href{https://doi.org/10.1016/j.jcp.2019.03.040}{H. Owhadi and G. R. Yoo, Kernel Flows: From learning kernels from data into the abyss, \emph{J. Comput. Phys.}, \textbf{389}\string:22--47, 2019.}

\bibitem{Owhadi20} \href{https://arxiv.org/abs/2008.03920}{H. Owhadi, Do ideas have shape? Plato's theory of forms as the continuous limit of artificial neural networks, arXiv\string:2008.03920, 2020.}

\bibitem{Paszke17}\href{https://openreview.net/pdf?id=BJJsrmfCZ}{A. Paszke, S. Gross, S. Chintala, G. Chanan, E. Yang, Z. DeVito, Z., Lin, A. Desmaison, L. Antiga, and A. Lerer, Automatic differentiation in PyTorch, In \emph{NIPS 2017 Workshop Autodiff}, 2017.}

\bibitem{Pytorch2019} \href{http://papers.neurips.cc/paper/9015-pytorch-an-imperative-style-high-performance-deep-learning-library.pdf}{A. Paszke, S. Gross, F. Massa, A. Lerer, J. Bradbury, G. Chanan, T. Killeen, Z. Lin, N. Gimelshein, L. Antiga, A. Desmaison, A. Kopf, E. Yang, Z. DeVito, M. Raison, A. Tejani, S. Chilamkurthy, B. Steiner, L. Fang, J. Bai, Junjie, and S. Chintala, PyTorch: An imperative style, high-performance deep learning library, In \emph{Advances in Neural Information Processing Systems 32 (H. Wallach, H. Larochelle, A. Beygelzimer, F. d'Alch\'{e}-Buc, E. Fox, and R. Garnett, eds.)}, pp. 8024--8035, 2019.}

\bibitem{Paulsen2016} \href{https://doi.org/10.1017/CBO9781316219232}{V. I. Paulsen and M. Raghupathi, \emph{An Introduction to the Theory of Reproducing Kernel Hilbert Spaces}, Cambridge University Press, Cambridge, 2016.}

\bibitem{Prieur16} \href{https://doi.org/10.1007/978-3-319-11259-6_35-1}{C. Prieur and S. Tarantola, Variance-based sensitivity analysis\string: theory and estimation algorithms, In \emph{Handbook of Uncertainty Quantification (R. Ghanem, D. Higdon, and H. Owhadi, eds.)}, pp. 1217--1239, Springer, Cham, 2016.}

\bibitem{Queipo05} \href{https://doi.org/10.1016/j.paerosci.2005.02.001}{N. V. Queipo, R. T. Haftka, W. Shyy, T. Goel, R. Vaidyanathan, and P. K. Tucker, Surrogate-based analysis and optimization, \emph{Prog. Aerosp. Sci.}, \textbf{41}(1)\string:1--28, 2005.}

\bibitem{Rabitz99} \href{https://doi.org/10.1016/S0010-4655(98)00152-0}{H. Rabitz, \"O. F. Ali\c{s}, J. Shorter, and K. Shim, Efficient input-output model representations, \emph{Comput. Phys. Commun.}, \textbf{117}(1-2)\string:11--20, 1999.}

\bibitem{Raissi18} \href{https://doi.org/10.1137/17M1120762}{M. Raissi, P. Perdikaris, and G. E. Karniadakis, Numerical Gaussian processes for time-dependent and nonlinear partial differential equations, \emph{SIAM J. Sci. Comput.}, \textbf{40}(1)\string:A172--A198, 2018.}

\bibitem{Rasmussen06} \href{https://doi.org/10.7551/mitpress/3206.001.0001}{C. E. Rasmussen and C. K. I. Williams, \emph{Gaussian Processes for Machine Learning}, MIT Press, Cambridge MA, 2006.}

\bibitem{Rosenbrock60} \href{https://doi.org/10.1093/comjnl/3.3.175}{H. H. Rosenbrock, An automatic method for finding the greatest or least value of a function, \emph{The Computer Journal}, \textbf{3}(3)\string:175--184, 1960.}

\bibitem{Sacks89} \href{https://doi.org/10.1214/ss/1177012413}{J. Sacks, W. J. Welch, T. J. Mitchell, and H. P. Wynn, Design and analysis of computer experiments, \emph{Stat. Sci.}, \textbf{4}(4)\string:409--423, 1989.}

\bibitem{Santner03} \href{https://doi.org/10.1007/978-1-4757-3799-8}{T. J. Santner, B. J. Williams, and W. I. Notz, \emph{The Design and Analysis of Computer Experiments}, Springer-Verlag, New York NY, 2003.}

\bibitem{Savin16} \href{https://doi.org/10.2514/6.2016-0433}{\'{E}. Savin, A. Resmini, and J. Peter, Sparse polynomial surrogates for aerodynamic computations with random inputs, In \emph{18th AIAA Non-Deterministic Approaches Conference, 4-8 January 2016, San Diego CA}, AIAA paper \#2016-0433, 2016.}

\bibitem{Savin17} \href{https://doi.org/10.1002/nme.5505}{\'{E}. Savin and B. Faverjon, Computation of higher-order moments of generalized polynomial chaos expansions, \emph{Int. J. Numer. Methods Eng.}, \textbf{111}(12)\string:1192--1200, 2017.}

\bibitem{Schobi2015} \amend{\href{https://doi.org/10.1615/Int.J.UncertaintyQuantification.2015012467}{R. Sch\"obi, B. Sudret, and J. Wiart, Polynomial-chaos-based Kriging, \emph{Int. J. Uncertainty Quantification}, \textbf{5}(2)\string:171--193, 2015.}}

\bibitem{Scholkopf01} \href{https://mitpress.mit.edu/books/learning-kernels}{B Sch\"{o}lkopf and A. J. Smola, \emph{Learning with Kernels. Support Vector Machines, Regularization, Optimization, and Beyond}, MIT Press, Cambridge MA, 2001.}

\bibitem{Schwai04} \href{https://dl.acm.org/doi/10.5555/2976040.2976192}{A. Schwaighofer, V. Tresp, and K. Yu, Learning Gaussian process kernels via hierarchical bayes, In \emph{NIPS'04: Proceedings of the 17th International Conference on Neural Information Processing Systems}, pp. 1209--1216, December 2004. }

\bibitem{Simpson01} \href{https://doi.org/10.1007/PL00007198}{T. W. Simpson, J. D. Poplinski, P. N. Koch, and J. K. Allen, Metamodels for computer-based engineering design: survey and recommendations, \emph{Eng. Comput.}, \textbf{17}(2)\string:129--150, 2001.}

\bibitem{Smolyak63}\href{http://mi.mathnet.ru/eng/dan/v148/i5/p1042 }{S.A. Smolyak, Quadrature and interpolation formulas for tensor products of certain classes of functions, \emph{Dokl. Akad. Nauk SSSR}, \textbf{148}(5):1042--1045, 1963.}

\bibitem{Soize04} \href{https://doi.org/10.1137/S1064827503424505}{C. Soize and R. G. Ghanem, Physical systems with random uncertainties: Chaos representations with arbitrary probability measure, \emph{SIAM J. Sci. Comput.}, \textbf{26}(2)\string:395--410, 2004. }

\bibitem{Spalart92} \href{https://doi.org/10.2514/6.1992-439}{P. R. Spalart and S. R. Allmaras, A one-equation turbulence model for aerodynamic flows, In \emph{30th Aerospace Sciences Meeting and Exhibit, 6-9 January 1992, Reno NV}, AIAA paper \#1992-0439, 1992.}

\bibitem{Stein99} \href{https://doi.org/10.1007/978-1-4612-1494-6}{M. L. Stein, \emph{Interpolation of Spatial Data}, Springer-Verlag, New York NY, 1999.}

\bibitem{Stone74}\href{https://doi.org/10.1111/j.2517-6161.1974.tb00994.x}{M. Stone, Cross-validatory choice and assessment of statistical predictions, \emph{J. R. Stat. Soc. Series B Stat. Methodol.}, \textbf{36}(2):111--133, 1974.}

\bibitem{Sudret08} \href{https://doi.org/10.1016/j.ress.2007.04.002}{B. Sudret, Global sensitivity analysis using polynomial chaos expansions, \emph{Reliab. Eng. Syst. Saf.}, \textbf{93}(7)\string:964--979, 2008.}

\bibitem{Sun19} \href{https://doi.org/10.1177\%2F0954410019864485}{G. Sun and S. Wang, A review of the artificial neural network surrogate modeling in aerodynamic design, \emph{Proc. IMechE Part G: J. Aerospace Engineering}, \textbf{233}(16)\string:5863--5872, 2019.}

\bibitem{Sun20} \href{https://doi.org/10.1016/j.taml.2020.01.031}{L. Sun and J.-X. Wang, Physics-constrained Bayesian neural network for fluid flow reconstruction with sparse and noisy data, \emph{Theor. Appl. Mech. Lett.}, \textbf{10}(3)\string:161--169, 2020.}

\bibitem{Thirumalainambi03} \href{https://doi.org/10.1117/12.486343}{R. Thirumalainambi and J. Bardina, Training data requirement for a neural network to predict aerodynamic coefficients, In \emph{Proceedings Volume 5102, Independent Component Analyses, Wavelets, and Neural Networks}, 2003.}

\bibitem{Berg08} \href{https://doi.org/10.1137/080714488}{E. van den Berg and M. P. Friedlander, Probing the Pareto frontier for basis pursuit solutions, \emph{SIAM J. Sci. Comput}, \textbf{31}(2):890--912, 2008.}

\bibitem{Berg11} \href{https://doi.org/10.1137/100785028}{E. van den Berg and M. P. Friedlander, Sparse optimization with least-squares constraints, \emph{SIAM J. Optim.}, \textbf{21}(4)\string:1201--1229, 2011.}

\bibitem{Wallach06} \href{https://doi.org/10.2514/6.2006-658}{R. Wallach, B. Mattos, R Girardi, and M. Curvo, Aerodynamic coefficient prediction of transport aircraft using neural network, In \emph{44th AIAA Aerospace Sciences Meeting and Exhibit, 09-12 January 2006, Reno, NV}, AIAA paper \#2006-658, 2006.}

\bibitem{Wand95} \href{https://doi.org/10.1201/b14876}{M. P. Wand and M. D. Jones, \emph{Kernel Smoothing}, Chapman and Hall/CRC, Boca Raton FI, 1995.}

\bibitem{Ward09} \href{https://doi.org/10.1109/TIT.2009.2032712}{R. Ward, Compressed sensing with cross validation, \emph{IEEE Trans. Inf. Theory}, \textbf{55}(12)\string:5773--5782, 2009.}

\bibitem{Wiener38} \href{https://doi.org/10.2307/2371268}{N. Wiener, The homogeneous chaos, \emph{Amer. J. Math.}, \textbf{60}(4)\string:897--936, 1938.}

\bibitem{Wein2019} \href{https://doi.org/10.2514/1.J057527}{J. Weinmeister, X. Gao, and S. Roy, Analysis of a polynomial chaos-Kriging metamodel for uncertainty quantification in aerodynamics, \emph{AIAA J.}, \textbf{57}(6)\string:2280--2296, 2019.}

\bibitem{Williams96} \href{http://papers.nips.cc/paper/1048-gaussian-processes-for-regression.pdf}{C. K. I. Williams and C. E. Rasmussen, Gaussian processes for regression, In \emph{Advances in Neural Information Processing Systems 8 (D. S. Touretzky, M. C. Mozer, and M. E. Hasselmo, eds.)}, pp. 514--520, MIT Press, Cambridge MA, 1996.}

\bibitem{Xiu02} \href{https://doi.org/10.1137/S1064827501387826}{D. Xiu and G. E. Karniadakis, The Wiener--Askey polynomial chaos for stochastic differential equations, \emph{SIAM J. Sci. Comput.}, \textbf{24}(2)\string:619--644, 2002.}

\bibitem{Yan18} \href{https://doi.org/10.3390/e20030191}{L. Yan, X. Duan, B. Liu, and J. Xu, Gaussian processes and polynomial chaos expansion for regression problem\string: Linkage via the RKHS and comparison via the KL divergence, \emph{Entropy}, \textbf{20}(3)\string:191, 2018.}

\bibitem{Yoo21} \href{https://doi.org/10.1016/j.physd.2021.132952}{G. R. Yoo and H. Owhadi, Deep regularization and direct training of the inner layers of Neural Networks with Kernel Flows, \emph{Physica D}, \textbf{426}\string:132952, 2021.}

\bibitem{Zhang21} \href{https://doi.org/10.1016/j.cma.2020.113485}{X. Zhang, F. Xie, T. Ji, Z. Zhu, and Y. Zheng, Multi-fidelity deep neural network surrogate model for aerodynamic shape optimization, \emph{Comput. Methods Appl. Mech. Eng.}, \textbf{373}\string:113485, 2021.}

\end{thebibliography}

\end{document}